\documentclass[conference,twoside,table]{IEEEtran}
\usepackage{tabularx,booktabs}
\usepackage{amsmath}
\usepackage{multirow}
\usepackage{graphicx}
\usepackage{subcaption}
\usepackage[colorlinks=true, allcolors=blue]{hyperref}
\usepackage{amsmath}
\usepackage{epstopdf}
\usepackage{mathtools}
\usepackage{comment}
\usepackage{array}
\usepackage[table,xcdraw]{xcolor}
\usepackage{colortbl}
\usepackage{placeins}

\begin{document}
\title{A Tree-Structured Approach for Phishing Template and Attacker Attribution Analysis}

\makeatletter
\newcommand{\linebreakand}{%
  \end{@IEEEauthorhalign}
  \hfill\mbox{}\par
  \mbox{}\hfill\begin{@IEEEauthorhalign}
}
\makeatother

\author{
    \IEEEauthorblockN{Unai Agirre\IEEEauthorrefmark{1}, Imanol Jerico\IEEEauthorrefmark{1}, Felipe Castano\IEEEauthorrefmark{1}\IEEEauthorrefmark{2}, Andrea Venturi\IEEEauthorrefmark{1}, Francesco Zola\IEEEauthorrefmark{1}}
    \IEEEauthorblockA{\IEEEauthorrefmark{1}\textit{Digital Security Department}, \textit{Vicomtech (BRTA)}, Donostia / San Sebastian, Spain}
    \IEEEauthorblockA{\IEEEauthorrefmark{2}\textit{University of León}, León, Spain
    \\\{uagirre, ijerico, fcastano, aventuri, fzola\}@vicomtech.org}
}

\IEEEoverridecommandlockouts

\maketitle
\IEEEpubidadjcol

\begin{abstract}
Phishing remains a persistent and evolving cybersecurity threat, with attack volumes reaching record levels. This growth is driven by the industrialization of phishing through widely available phishing kits and reusable templates, which enable cybercriminals to rapidly generate and deploy large numbers of fraudulent webpages. Although surface-level attributes may differ across these websites, their underlying structures often exhibit significant similarities. However, most existing defenses rely on reactive blocklists or supervised classification models that focus on individual phishing instances, limiting their ability to identify structural reuse and detect coordinated phishing campaigns.
To address this limitation, this study investigates whether HTML structure can serve as a robust fingerprint for identifying phishing template reuse. We model webpages as Document Object Model (DOM) trees and extract structural features, optionally enriched with HTML tag-based content information. These representations are then clustered using unsupervised learning methods to group structurally similar webpages. Three clustering algorithms are evaluated and compared, while also analyzing how the depth of the extracted DOM-tree affects cluster formation and overall clustering performance. Finally, cluster quality is also evaluated both quantitatively and qualitatively, including a novel level-wise Jaccard Distance Score and manual inspection supported by visualization tools.
Results demonstrate that structural representations of webpages can effectively reveal hidden similarities across phishing sites, enabling the detection of emerging and zero-day templates and supporting the analysis of coordinated phishing threats.
\end{abstract}

\begin{IEEEkeywords}
Phishing Template, Structural Analysis, Attacker Attribution, Clustering, DOM analysis, Tree extraction
\end{IEEEkeywords}

\section{Introduction}
\label{sec:introduction}
Phishing remains a persistent and evolving threat within the cybersecurity landscape, maintaining its efficacy despite user awareness. According to the Anti-Phishing Working Group (APWG) \cite{APWG2025Q1}, attack volumes reached record levels in early 2025, with over 1 million unique phishing attacks detected in the first quarter alone, the highest volume observed since late 2023. This resurgence is characterized by a high degree of sector-specific targeting; the Software as a Service (SaaS) and Webmail category remains the most frequently attacked sector (17.6\%), while the financial and payment industries combined account for over 30.9\% of all observed incidents. These numbers underscore a broader trend: phishing is no longer a collection of isolated fraudulent events but a massive, industrialized operation that continues to scale across global infrastructures \cite{europol2026iocta}.

Traditional defense mechanisms primarily focus on identifying individual malicious instances through reactive blocklists or supervised machine learning models \cite{li2024machine}. While these approaches achieve high performance in binary classification \cite{Asiri2024PhishTransformer, Jiang2025DPhishNet}, they remain limited in their ability to provide adversarial insight, as they fail to capture the broader context of attack infrastructures or the structural signatures left by organized actors. This limitation is increasingly relevant, as the proliferation of phishing kits, reusable templates, and web components \cite{song2026phishing, castano2023phikita, bijmans2021catching} enables adversaries to deploy numerous fraudulent pages rapidly with minimal effort \cite{europol2026iocta}. In these cases, phishing instances may vary in their domain names or superficial visual elements \cite{jain2017phishing}, but they frequently retain a congruent internal architecture.

For this reason, the primary objective of this study is to evaluate whether the HTML structure of a webpage serves as a sufficiently robust fingerprint for identifying similarities among malicious sites, thereby providing a stable indicator of phishing structural logic and enabling the detection of systematic template reuse across malicious websites.

Specifically, to do this, the proposed methodology uses Document Object Model (DOM) information to build a parent–child tree. Then, tree properties (structural information) are used exclusively and combined with content-based data (HTML tags within the DOM) to feed unsupervised learning techniques that cluster similar samples without the need for prior labeling. Finally, once groups of similar trees are generated, the methodology proposes a quantitative and qualitative validation. The first is achieved by introducing a new metric based on level-wise Jaccard Distance Score, while the second is performed manually using dedicated graph/tree visualization tools.

By focusing on revealing structures rather than isolated events, this framework offers the adaptability necessary to identify emergent, zero-day phishing templates that evolve continuously to evade static or supervised detection rules. Once identified, these templates form the foundation for the comprehensive characterization and attribution of coordinated phishing campaigns.

Summarizing, the main contributions of this study are:
\begin{itemize}
    \item Validate if tree-based representation of phishing webpages derived from DOM can be used for detecting structural templates;
    
    \item Compare the performance of three different clustering methodologies when tree properties (structural) are used solely or combined with content-based information (HTML tags);

    \item Validate how the depth of the tree, i.e., the number of nested levels, affects clustering performance;
    
    \item Define and validate a new level-wise Jaccard Distance Score to evaluate clustering performance (quantitative validation).
    
\end{itemize}

The remainder of this article is structured as follows. Section~\ref{sec:background} provides the theoretical background and reviews the state-of-the-art relevant to the proposed approach. Section~\ref{sec:methodology} outlines the methodology, while Section~\ref{sec:expstudy} details the dataset, experiments, and model configurations. Section~\ref{sec:results} reports the results, while Section~\ref{sec:validation} presents qualitative insights, limitations, and ethical concerns. Finally, Section~\ref{sec:conclusion} concludes the study and suggests potential avenues for future research.

\section{Background}
\label{sec:background}

\subsection{Clustering}\label{subsec:clustering}
The goal of this work is to detect and group phishing templates that exhibit similar structural patterns. To achieve this, clustering algorithms are used as the main analytical tool, since they enable the identification of latent relationships between webpages without requiring prior labels. 

Among the different clustering paradigms, this study focuses on density-based and hierarchical approaches. Density-based methods are suitable for phishing analysis because they can identify groups with arbitrary shapes while also separating isolated samples that may correspond to structural outliers. Hierarchical methods provide a complementary perspective by organizing samples according to nested similarity relationships. Together, these two families of methods make it possible to capture diverse similarity patterns, which is particularly important in a phishing ecosystem characterized by template reuse, campaign variation, and heterogeneous construction strategies \cite{oest2018inside, merlo2022phishing}.

Among density-based methods, DBSCAN (Density-Based Spatial Clustering of Applications with Noise) \cite{ester1996density} forms clusters by connecting neighbouring samples that satisfy two conditions: a maximum neighbourhood distance $\epsilon$ and a minimum number of points \textit{min\_samples} to form a cluster. Its main advantages are that it does not require the number of clusters to be specified in advance and that it can identify noise or isolated samples. However, its performance is sensitive to the choice of $\epsilon$ and may decrease when clusters have different densities. OPTICS (Ordering Points To Identify the Clustering Structure) \cite{ankerst1999optics} addresses this limitation by producing an ordering of the samples that reflects their density-based connectivity. This makes it more suitable for data with varying density levels, although the resulting ordering must still be interpreted or converted into final clusters.

Agglomerative Hierarchical Clustering (AHC) \cite{mullner2011modern} follows a bottom-up strategy, where each sample initially forms its own cluster and the most similar clusters are progressively merged according to a linkage criterion. This process produces a hierarchy of nested groups, allowing the data to be examined at different levels of granularity. AHC does not necessarily require a predefined number of clusters, depending on the chosen stopping criterion. However, once two clusters are merged, the decision cannot be reversed, and the final structure can be sensitive to the selected distance measure and linkage criterion.

\subsection{Related Work}
Phishing Analysis is divided into two complementary groups \cite{sahingoz2019machine}. On the one hand, the first kind of approaches treats phishing detection as a binary classification problem, prioritizing the immediate identification and mitigation of malicious instances to protect the end-user. On the other hand, the second approach focuses on post-detection analysis, using confirmed phishing samples to extract structural patterns, common infrastructures, and forensic markers. This second perspective shifts the objective from simple filtering to the characterization of phishing templates and the potential attribution of adversarial groups. The following sections detail the technical evolution of both methodologies.

\subsubsection{Phishing Detection as Binary Classification}
This research group focuses on the development of computationally efficient and robust architectures to distinguish between legitimate and malicious entities. One of the highlight approaches is that proposed by Otieno et al. \cite{Otieno2023Detecting}. In this approach, the authors use semantic analysis to eliminate manual feature engineering, demonstrating that bidirectional Transformers (BERT) capture contextual representations of URL strings, achieving 96\% accuracy. To address the trade-off between efficiency and robustness, hybrid ensemble systems have been proposed to integrate BERT-derived vectors with stacked classifiers and technical parameters like SSL validity, achieving accuracies up to 98.6\% \cite{Mandapati2023Hybrid, Kumar2024Proactive}.

Beyond URL analysis, Asiri et al. \cite{Asiri2024PhishTransformer} introduced PhishTransformer, an approach that examines the HTML web page source as complementary information to identify sophisticated threats. The authors use CNNs and Transformers to analyze embedded URL links within the HTML content, addressing the limitations of traditional systems that rely solely on the primary URL. Experimental results on a balanced dataset of 50,000 samples demonstrated that PhishTransformer achieves an average accuracy of 98.9\% and an F1-score of 99\%. Later, Jiang et al. \cite{Jiang2025DPhishNet} developed D-PhishNet, employing Graph Attention Networks (GATs) to fuse URL and HTML features dynamically. The authors introduced the Dual-Branch Alignment Mixer (DBA-CMixer), which uses gated fusion and channel attention to integrate structural URL data with latent semantic features extracted via a pre-trained BERT module to address the ``semantic gap" inherent in multi-feature fusion. The authors report an accuracy of up to 99.30\%.

While binary classification models achieve high performance in isolating malicious samples, they remain limited in their ability to provide adversarial insight. By focusing strictly on the categorization of isolated events, these methods fail to capture the broader context of the attack infrastructure or the underlying relationships between samples. Given that our research focus is the analysis of patterns within attacks to detect authorship and phishing templates, traditional binary models are insufficient for identifying the systemic signatures left by organized actors.

\subsubsection{Phishing Template Analysis}
Another group focuses on analyzing confirmed phishing samples to identify templates and shared patterns. Nguyen et al. \cite{Nguyen2014Detecting} proposed identifying structural clones of legitimate pages by calculating similarity between DOM trees using genetic algorithms, achieving a 0\% false-negative rate at specific thresholds. While achieving high accuracy at a threshold of $\delta = 0.6$, the method remains sensitive to obfuscation and exhibits a significant trade-off between false positives and negatives depending on threshold calibration. Unlike this approach, which seeks to identify a phishing site by its proximity to a legitimate target, our research shifts the focus from victim-mimicry detection toward uncovering latent structural links and shared indicators across diverse phishing attacks to characterize broader patterns. 

To bridge the gap between active attacks and their generative sources, Castaño et al. \cite{castano2023phikita} introduced PhiKitA, a dataset specifically designed to link phishing kits directly to their deployed website counterparts. Their analysis employed graph representation of the HTML DOM, MD5 hashing, and structural fingerprints to identify familiarity among samples, successfully uncovering a coordinated campaign involving 172 distinct attacks against a single financial institution. The graph-based approach achieved high performance in binary detection tasks (92.5\% accuracy).  While the findings demonstrated that structural HTML data is critical for tracing the lineage of phishing operations, the reliance on known kit repositories highlights a dependency that limits detection to documented sources. In contrast, the methodology proposed in this work uses unsupervised clustering of web pages, which does not require a prior sample of detected phishing templates. By operating independently of kit-specific metadata, this approach enables the identification of zero-day samples and emergent templates that remain undocumented in existing repositories.

As mentioned earlier, the nature of templates detection involves extracting patterns across heterogeneous file types and different attack vectors, which interferes with the use of supervised learning methods due to the lack of annotated datasets. Consequently, unsupervised methodologies are required to discover latent relationships without prior labeling. Althobaiti et al. \cite{Althobaiti2023Using} addressed this by evaluating Mean Shift and DBSCAN algorithms to group corporate email threats, demonstrating that Mean Shift achieves 82\% precision in assigning samples to representative clusters based on URL and origin metadata. Although this approach uses email corpora rather than raw HTML, the underlying methodology aligns with current research objectives; the unsupervised nature of the framework provides critical adaptability to emergent templates that constantly evolve to evade static detection rules.

\section{Methodology}
\label{sec:methodology}

The main goal of this methodology is to analyse and validate structural similarity among phishing webpages. In this sense, the methodology leverages DOM information to build a tree-based representation and then extracts structural and content-based features to assess similarities among phishing webpages. In this way, the process enables the identification of pages that are likely (i) deployed by the same attacker or (ii) created using the same template or phishing kit.

The proposed methodology comprises four main steps: \textit{data preprocessing}, \textit{tree construction}, \textit{feature extraction \& clustering}, and \textit{validation}.

\subsection{Phase 1: Data Preprocessing}

The experimental evaluation of the proposed methodology uses the dataset developed by Aljofey et al. \cite{aljofey2025comprehensive}, specifically the subset identified as DS-2. This dataset was selected for its inclusion of full HTML source code, which is an essential requirement for the tree-based representation described in Phase 2. Collected during 2023 and published in 2025, the corpus captures current evasion tactics and modern web design patterns, thereby serving as a benchmark for structural analysis. Phishing samples were collected from PhishStats and validated between July 15 and August 9, 2023. Legitimate instances were sourced from Alexa top-ranked websites as of November 2023. From the complete dataset, this study focuses exclusively on phishing instances, discarding legitimate samples as noise for the specific scope of this research. During preprocessing, the remaining samples are thoroughly cleaned by removing duplicate records and incomplete data.


\begin{figure}[htbp]
    \centering
        \includegraphics[width=\linewidth]{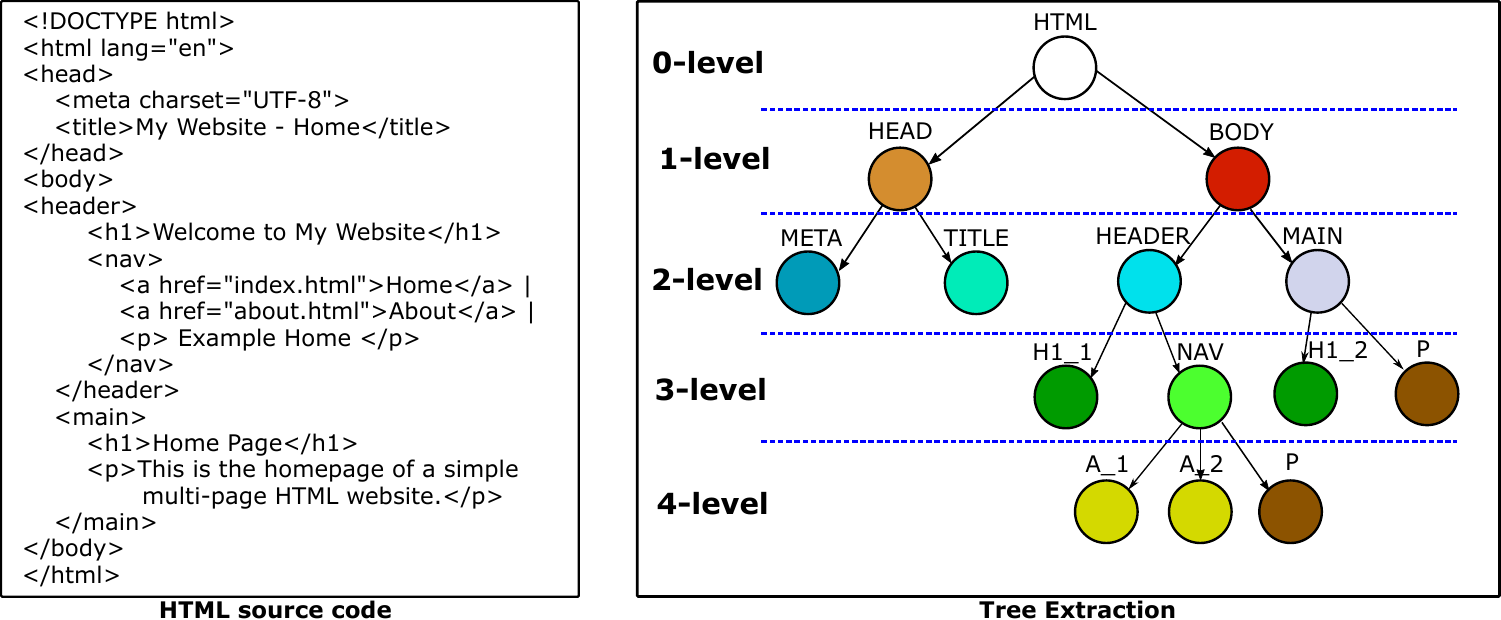}
        \caption{Example of a tree extracted from an HTML DOM.}
        \label{fig:tree}
\end{figure}

\subsection{Phase 2: Tree Construction}
Starting from the HTML content of phishing pages, the first phase of the proposed methodology consists of constructing a tree-based representation of their DOM. The objective is to preserve the hierarchical structure of the original document by representing each HTML element (or tag) as a node in the tree. In this representation, nodes correspond to individual HTML tag, while edges capture the parent–child relationships between elements. More specifically, an edge connects a parent node to a child node when the corresponding HTML tag is nested within another tag in the original document, as shown in Figure \ref{fig:tree}. This representation allows the construction of a tree with varying depths, according to the nesting structure of the DOM. Each depth in the tree defines a level. Thus, webpages share a common top-level DOM structure, where the root node (0-level) corresponds to the \texttt{<html>} element and its immediate children (1-level) correspond to the \texttt{<head>} and \texttt{<body>} elements.

To prevent node collision when identical tags appear multiple times at the same hierarchical level, each node receives a unique identifier comprising the tag name, depth level, local child index, and a cumulative traversal counter.

\subsection{Phase 3: Feature Extraction \& Clustering}\label{subsec:phase2}
Once the trees have been generated, this phase focuses on extracting structural and content-based parameters to characterize each individual tree.  More specifically, the study considers two complementary groups of features. The first group relies on 5 structural properties of the tree, reported in Table \ref{tab:treeproperties}. These features capture the overall shape and complexity of the tree. On the other hand, the second group extends this structural description by incorporating content-based information, that are the occurrence of the HTML tags observed at each level of the DOM tree (their distribution across the hierarchy). Thus, the following 22 HTML tags are used: \texttt{form}, \texttt{input}, \texttt{button}, \texttt{a}, \texttt{iframe}, \texttt{script}, \texttt{img}, \texttt{link}, \texttt{meta}, \texttt{div}, \texttt{span}, \texttt{label}, \texttt{style}, \texttt{title}, \texttt{svg}, \texttt{section}, \texttt{picture}, \texttt{ul}, \texttt{li}, \texttt{nav}, \texttt{g}, \texttt{p}. These elements are selected for their relevance in phishing threat since their are commonly associated with credential collection interfaces, embedded external resources, page layout, and client-side behavior. By integrating this tag-level information, the resulting representation preserves not only the structural arrangement of nodes but also the semantic role of the elements composing the tree.

\begin{table*}[h]
    \centering
    \begin{tabularx}{\linewidth}{ll X}
        \hline
        \textbf{} &\textbf{Features} & \textbf{Description} \\ \hline
        1 & Number of nodes & It represents the total count of HTML elements within the pruned DOM tree, indicating the overall structural complexity of the web page. \\ 
        2 & Number of edges & It quantifies the total hierarchical and relational links between elements. A high edge-to-node ratio often suggests a more complex layout. \\ 
        3 & Maximum centrality & It identifies the most interconnected node in the graph. In phishing, this often corresponds to a central container or a critical form element \cite{white1994betweenness}. \\ 
        4 & Minimum centrality & It measures the lowest degree of connectivity, helping to identify peripheral elements or isolated scripts often used for obfuscation \cite{white1994betweenness}. \\ 
        5 & Centrality on average & This feature provides a global measure of structural importance across all elements, capturing the general density of the page's architecture \cite{Freeman1978}. \\ \hline
    \end{tabularx}
    \caption{List of structural properties extracted from the tree.}
    \label{tab:treeproperties}
\end{table*}

Thanks to this feature extraction process, trees are converted into numerical vectors that provide a compact yet informative representation of webpages, enabling objective mathematical comparisons and aggregation. Thus, clustering algorithms are applied to group webpages exhibiting similar structural and content-based characteristics, allowing for the identification of latent patterns and templates that may not be immediately apparent through manual inspection.

\subsection{Phase 4: Validation}\label{subsec:validation}

In this fourth phase, the groups generated by the clustering algorithms are evaluated both quantitatively and qualitatively. Specifically, in the first case, a dedicated metric based on Jaccard distance is used, while in the second case, representative samples are visually inspected for the final evaluation.

\textbf{Quantitative Validation.} In this step, to compare the performance of the clustering algorithms, a metric based on the level-wise Jaccard comparison of HTML tag-frequency distributions is used. We refer to this metric as the Level-wise Jaccard Distance Score ($\mathcal{LJD}$). The proposed metric evaluates cluster quality by measuring whether samples within the same cluster exhibit similar distributions of HTML tags across DOM levels, while samples assigned to different clusters remain sufficiently dissimilar.

First, for each pair of samples $(x,y)$ and for each DOM level $l$, both samples are represented by tag-frequency counters. Specifically, $c_{x,l}(t)$ and $c_{y,l}(t)$ denote the number of occurrences of tag $t$ at level $l$ in samples $x$ and $y$, respectively. The level-wise Jaccard similarity is computed as:

\begin{equation}
\mathcal{J}_{l}(x,y) =
    \frac{
        \sum_{t \in \mathcal{T}_{l}} \min(c_{x,l}(t), c_{y,l}(t))
        }{
        \sum_{t \in \mathcal{T}_{l}} \max(c_{x,l}(t), c_{y,l}(t))
    }
\label{eq:ljd_level_jaccard}
\end{equation}

where the numerator represents the shared tag occurrences between the two samples at level $l$, while the denominator represents the total tag occurrences after combining both samples, computed through the maximum tag counts.

The level-wise similarities are then combined through a weighted average:

\begin{equation}
\mathcal{LJ}(x,y) =
\sum_{l \in \mathcal{L}} w_l \cdot \mathcal{J}_{l}(x,y),
\quad
\text{with}
\quad
w_l =
\frac{\frac{1}{l}}
{\sum_{k \in \mathcal{L}} \frac{1}{k}}
\label{eq:ljd_weighted_similarity}
\end{equation}

where $\mathcal{L}$ denotes the set of DOM levels considered in the comparison. In this work, the first two DOM levels are not considered, since they usually correspond to the common top-level structure of HTML documents. The remaining levels are weighted according to their depth, assigning larger weights to upper levels and lower weights to deeper levels. This weighting scheme reflects the intuition that differences in higher DOM levels are more relevant for identifying changes in the global structure of a webpage, whereas differences in deeper levels are more likely to correspond to local variations, nested layout details, or minor implementation differences.

Since $\mathcal{LJ}(x,y)$ defines a similarity score, it is converted into a distance as follows:

\begin{equation}
\mathcal{LJD}(x,y) = 1 - \mathcal{LJ}(x,y)
\label{eq:ljd_distance}
\end{equation}

This transformation preserves the usual interpretation of distance-based cluster validation: small distances indicate similar samples, whereas large distances indicate dissimilar samples.

Let $\mathcal{P}^{A} = \{C_1, C_2, \ldots, C_K\}$ be the partition generated by a clustering algorithm $A$, where $K$ is the number of clusters produced by the algorithm. For each cluster $C_i$, it is possible to compute its internal compactness (the intra metric) using Equation~\ref{eq:ljd_intra}, i.e., the average distance between unordered pairs of distinct samples within $C_i$. Likewise, it is possible to compute the degree of separation between a cluster and all the others (the inter metric), i.e., the average distance between samples belonging to different clusters, as shown in Equation~\ref{eq:ljd_inter}. This latter metric yields an array of distances for each cluster. Therefore, for each $C_i$ they can be aggregated into a single value, denoted by $\overline{\mathcal{LJD}}_{\text{inter}}$, as shown in Equation~\ref{eq:ljd_inter_mean}.


\begin{equation}
\mathcal{LJD}_{\text{intra}}(C_i) =
\frac{1}{\binom{|C_i|}{2}}
\sum_{\substack{x,y \in C_i \\ x < y}}
\mathcal{LJD}(x,y)
\label{eq:ljd_intra}
\end{equation}

\begin{equation}
\mathcal{LJD}_{\text{inter}}(C_i,C_j) =
\frac{1}{|C_i|\,|C_j|}
\sum_{x \in C_i}
\sum_{y \in C_j}
\mathcal{LJD}(x,y),
\quad i \neq j
\label{eq:ljd_inter}
\end{equation}

\begin{equation}
\overline{\mathcal{LJD}}_{\text{inter}}(C_i) =
\frac{1}{K-1}
\sum_{\substack{j=1 \\ j \neq i}}^{K}
\mathcal{LJD}_{\text{inter}}(C_i,C_j)
\label{eq:ljd_inter_mean}
\end{equation}

Using equation \ref{eq:ljd_cluster_score}, intra and inter metrics can be finally combined to generate an unique value for each cluster ($\mathcal{LJD}(C_i)$), where small values indicate that cluster $C_i$ is internally compact and well separated from the remaining clusters. Conversely, high values indicate that the internal dispersion of the cluster is large relative to its separation from the rest of the partition.

\begin{equation}
\mathcal{LJD}(C_i) =
\frac{
\mathcal{LJD}_{\text{intra}}(C_i)
}{
\overline{\mathcal{LJD}}_{\text{inter}}(C_i)
}
\label{eq:ljd_cluster_score}
\end{equation}

The $\mathcal{LJD}^{A}$ (Level-wise Jaccard Distance Score associated with algorithm $A$) is defined as a vector containing each $\mathcal{LJD}(C_i)$ for all the $C_i \in \mathcal{P}^{A}$, as detailed in Equation~\ref{eq:ljd_algorithm_vector}. 

\begin{equation}
\mathcal{LJD}^{A} =
\left[
\mathcal{LJD}(C_1),
\mathcal{LJD}(C_2),
\ldots,
\mathcal{LJD}(C_K)
\right]
\label{eq:ljd_algorithm_vector}
\end{equation}

Finally, to summarize the behaviour of each clustering algorithm, three descriptive statistics are computed over the elements of the vector $\mathcal{LJD}^{A}$: the mean ($\mathcal{LJD}^A_{mean}$), the standard deviation ($\mathcal{LJD}^A_{std}$), and the maximum value ($\mathcal{LJD}^A_{max}$). Yet, $\mathcal{LJD}^A_{mean}$ provides an overall estimate of cluster quality, the $\mathcal{LJD}^A_{std}$ measures the variability of the cluster-level scores, and the $\mathcal{LJD}^A_{max}$ identifies the most problematic cluster generated by the algorithm. As $\mathcal{LJD}_{intra}(C_i)$ is not defined for singletons (i.e., clusters with just one element), we do not consider them for the $\mathcal{LJD}^A_{mean}$ and $\mathcal{LJD}^A_{std}$ computations.

It should be noted that $\mathcal{LJD}^{A}$ differs from standard feature-space clustering metrics. Rather than evaluating the clustering partition directly on the normalized feature vectors used by the clustering algorithms, it is computed from the extracted DOM-tree representations by comparing the level-wise distributions of HTML tags. Therefore, its interpretation plays a key role in the cluster evaluation, providing a DOM-level validation criterion complementary to standard feature-space clustering metrics.

\textbf{Qualitative Validation.} In this case, due to the high number of groups generated by the clustering algorithms, only $N$ representative clusters are considered, as shown in step 2 in Figure \ref{fig:cluster}. Specifically, the $N$ most compact are selected, i.e., those with the lowest $\mathcal{LJD_{\text{intra}}}$ values. Then, they are visually inspected through their cluster \textit{medoids}.
 
\textit{Medoid Evaluation.} For each of the selected $N$ compact cluster, the sample with the minimum average $\mathcal{LJD}$ to the other elements of the same cluster is identified as the cluster medoid (step 3 in Figure \ref{fig:cluster}). These medoids are then visually compared across the selected clusters to assess whether they provide representative and distinguishable structural patterns. Specifically, the graphs are analysed and visualised using the Gephi tool \cite{bastian2009gephi}. Then, to obtain a more comprehensive view of the selected groups, $C$ additional samples within each considered cluster are selected and compared with their medoid to assess consistency within each cluster (step 4 in Figure \ref{fig:cluster}).

\begin{figure}[htbp]
    \centering
        \includegraphics[width=\linewidth]{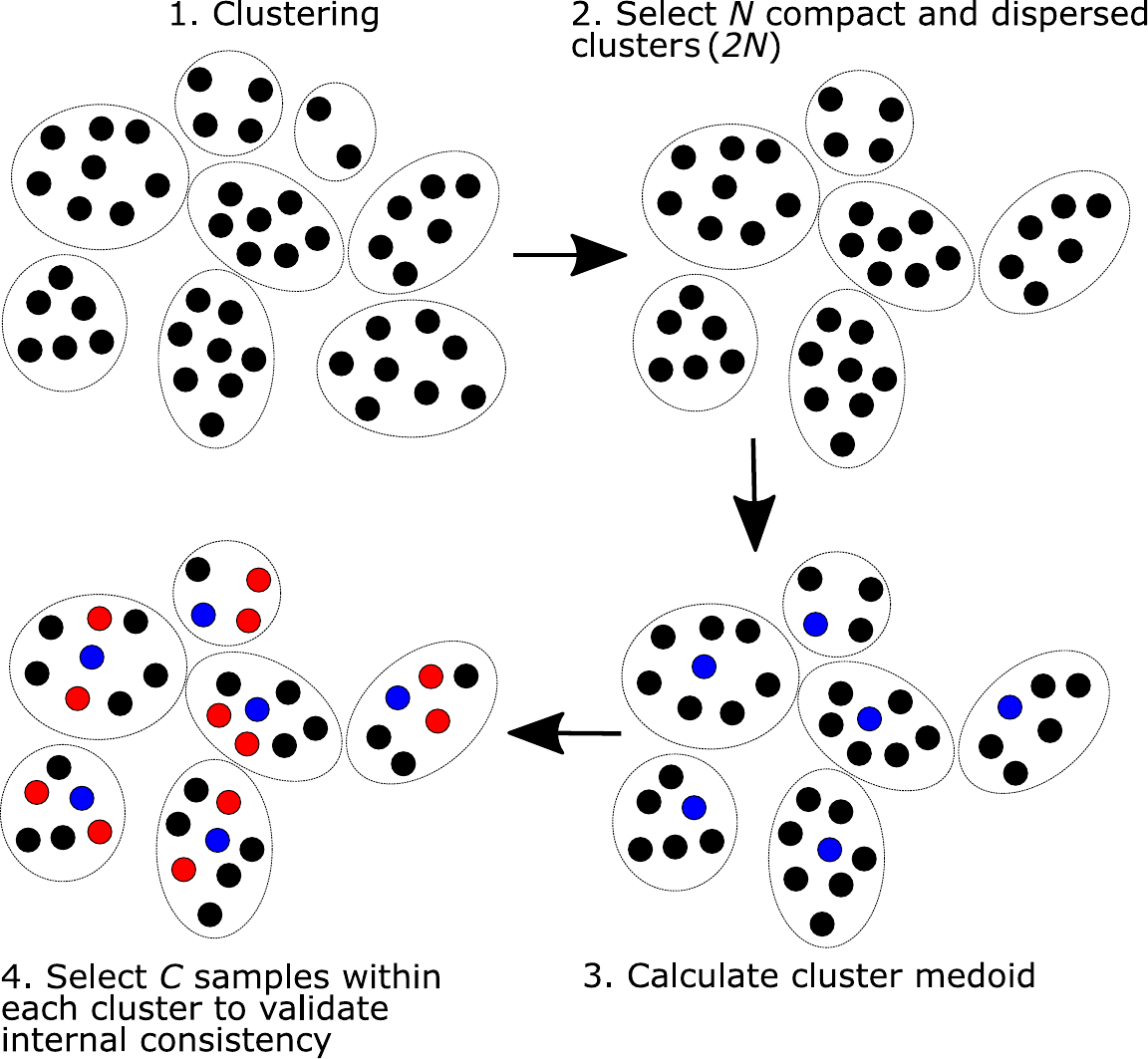}
        \caption{Schema of the qualitative validation steps followed during the analysis.}
        \label{fig:cluster}
\end{figure}

\section{Experimental Study} \label{sec:expstudy}


\subsection{Dataset}\label{subsec:dataset}
The experimental evaluation of the proposed methodology uses the dataset developed by Aljofey et al. \cite{aljofey2025comprehensive}, specifically the subset identified as DS-2. This dataset was selected for its inclusion of full HTML source code, which is an essential requirement for the tree-based representation described in Phase 1. Collected during 2023 and published in 2025, the corpus captures current evasion tactics and modern web design patterns, thereby serving as a benchmark for structural analysis. The dataset contains 23,366 samples, including 8,366 phishing instances and 15,000 legitimate instances. Phishing samples were collected from PhishStats\footnote{https://www.phishstats.info/} and validated between July 15 and August 9, 2023. Only the sub-corpus of 8,366 phishing pages is used in this study, as the primary objective is the structural analysis and clustering of phishing threats to identify recurrent attack patterns.

\subsection{Experiment setup}\label{subsec:experiment}
As mentioned earlier, each webpage can generate a tree with different depth, depending on the number of nested elements. In order to provide a more comprehensive view of similarity, this study investigates how the number of considered levels can affect the clustering task. In particular, varying the depth of the extracted DOM representation may influence both the structural detail captured and the resulting similarity between webpages. For this reason, eight different depths are considered and compared, ranging from 8 to 15 levels. This range is chosen to balance representational completeness and computational efficiency, ensuring that both shallow and deep structural information are evaluated. Specifically, for each maximum level selected, if a webpage has a deeper tree, it is pruned to the maximum fixed depth. If a sample has fewer levels, the missing-level (tags) information is set to 0.

For the qualitative analysis, the parameters $N$ and $C$ must be fixed. In this study, the number of representative compact clusters is set to $N$ = 3, while the number of additional samples per cluster is set to $C$ = 2. Furthermore, to ensure a meaningful and reliable validation, only clusters containing more than 4 elements are considered in the qualitative analysis. This filtering step excludes clusters that, although potentially relevant, do not contain a sufficient number of samples to support a representative visual inspection and robust qualitative assessment of compact consistency. As a result, the analysis focuses on clusters with adequate cardinality, reducing the risk of drawing conclusions from sparsely populated or statistically unstable groups.

\subsection{Model configurations}\label{subsec:modelsconfig}
To obtain a broader view of the clustering performance, the results of three different algorithms are compared. Specifically, two density-based methods (DBSCAN and OPTICS) and one hierarchical method (Agglomerative-AHC) are used. Nevertheless, as introduced in Section \ref{subsec:clustering}, these algorithms require the tuning of specific parameters. Thus, for both density-based methods, a grid-search mechanisms is implemented varying the $\epsilon$ (or $xi$) and \texttt{min\_samples} parameters, while different distance thresholds are used for hierarchically-based method, as reported in Table \ref{tab:clustering_params}. 

To evaluate the performance of the clustering algorithms and perform the grid search analysis, state-of-the-art metrics are considered \cite{ashari2023analysis}, such as the Silhouette Coefficient ($\mathcal{S}_{score}$), the Calinski-Harabasz Index ($\mathcal{CH}_{index}$) and the Davies-Bouldin Index ($\mathcal{DB}_{index}$). These metrics are mainly used during grid search to identify the best configuration for each algorithm at each considered level of the analysis. Yet, the search approach is initially applied using only the structural properties of the tree, and then combining also content-based features, as detailed in Section \ref{subsec:phase2}.

Notably, during the search task, the introduced metric ($\mathcal{LJD}$) is not used, as the analysis prioritizes established and well-validated metrics. Thus, once best configurations are detected, they are further validated using quantitative and qualitative approaches.

\begin{table}[h]
\resizebox{\linewidth}{!}{%
\centering
\begin{tabular}{c c c}
\hline
\textbf{Algorithms} &\textbf{Parameter}  & \textbf{Values} \\
\hline
\multirow{2}{*}{DBSCAN}  &$\epsilon$  &\{0.1, 0.2,..., 1.0\}, step=0.1\\
&\texttt{min\_samples} & 2, 3, 5, 8, 10 \\
\hline
\multirow{2}{*}{OPTICS}  &$xi$  &\{0.1, 0.2,..., 1.0\}, step=0.1\\
&\texttt{min\_samples} & 2, 3, 5, 8, 10 \\\hline
AHC  &$t$  &5, 25, 50, 75, 100\\
\hline
\end{tabular}}
\caption{Grid search parameters for clustering algorithms.}
\label{tab:clustering_params}
\end{table}

\section{Results}\label{sec:results}

\subsection{Grid-search results}\label{subsec:gridsearch}

The results of the clustering algorithms during the search phase are reported in Figures \ref{fig:dbscan_gridsearch}, \ref{fig:optics_gridsearch}, and \ref{fig:ahcdhc_gridsearch}. Specifically, for simplicity, the analysis reports only the $\mathcal{S}_{score}$, used as the performance-driven metric to detect the best configuration. Then, for the best configuration obtained at each level, the corresponding $\mathcal{CH}_{index}$ and $\mathcal{DB}_{index}$ values are further detailed in the next section. However, this section shows only the results obtained when structural features are used. The same grid-search analysis is also repeated using both structural and content-based features, and the results of this second case are reported in Appendix \ref{sub:app1}.
\begin{figure*}[!htbp]
    \centering
    \begin{subfigure}[b]{0.24\textwidth}
        \includegraphics[width=\textwidth]{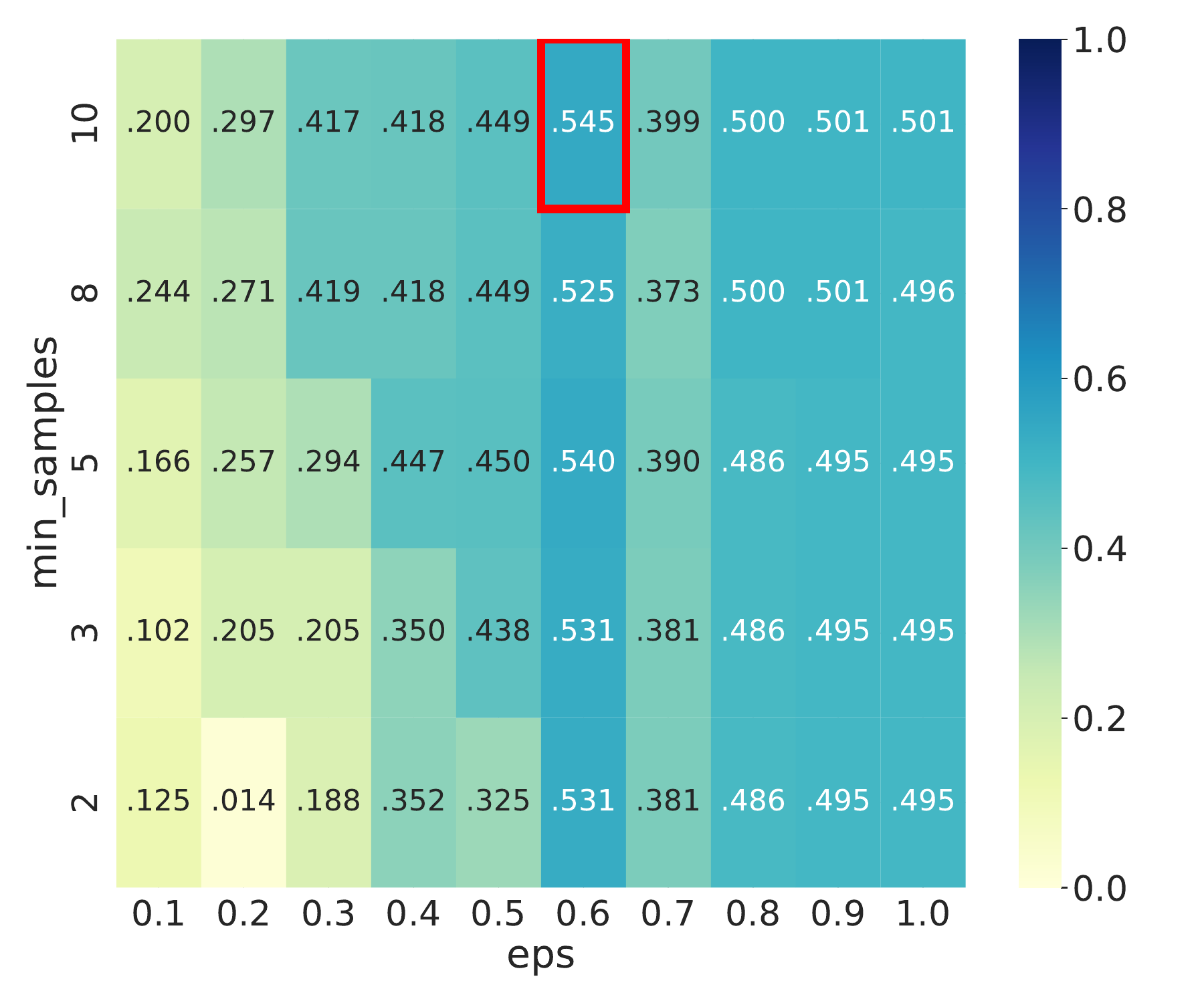}
        \caption{Level 8.}
        \label{fig:d_lvl8}
    \end{subfigure}
    \begin{subfigure}[b]{0.24\textwidth}
        \centering
        \includegraphics[width=\textwidth]{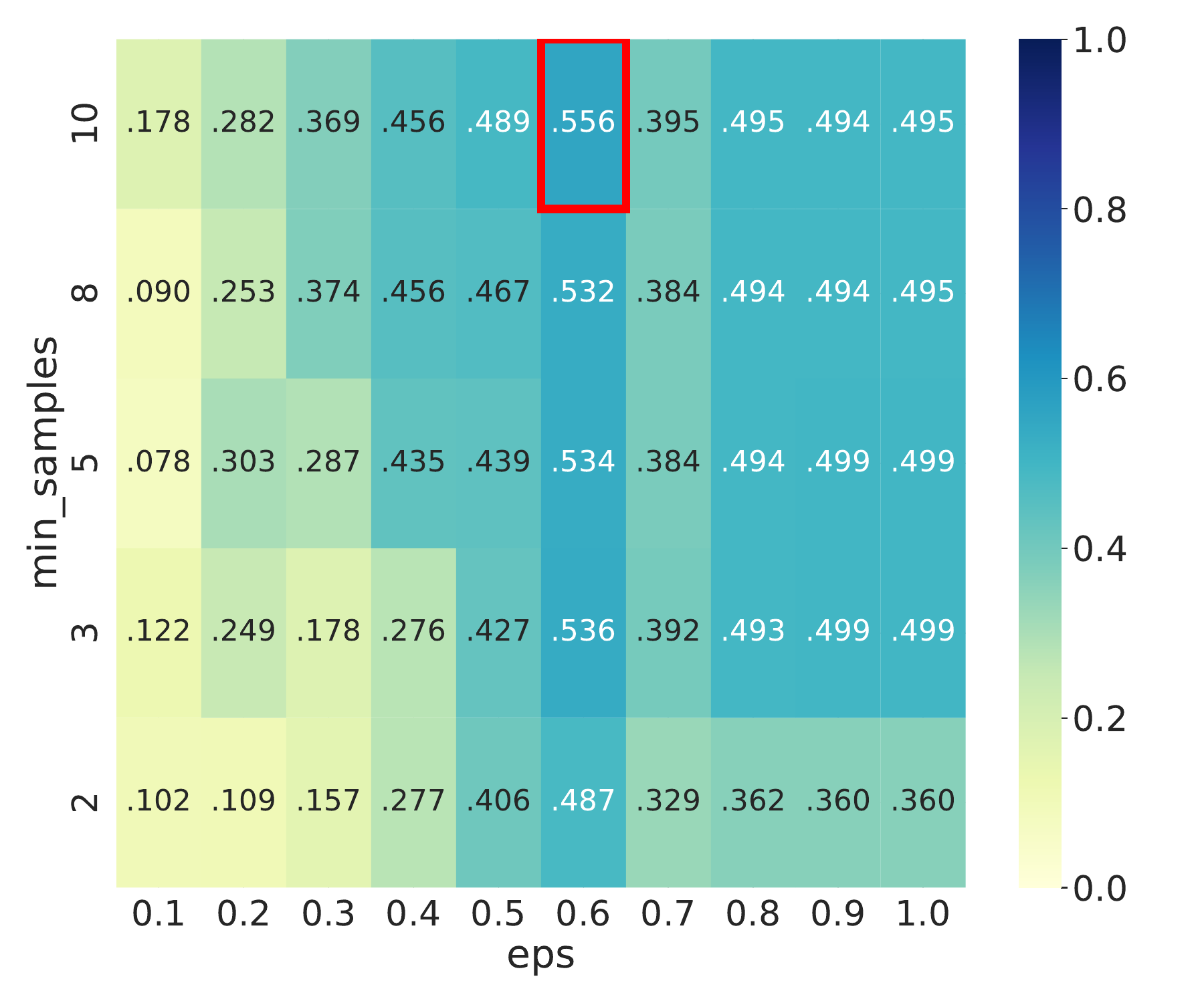}
        \caption{Level 9.}
        \label{fig:d_lvl9}
    \end{subfigure}
    \begin{subfigure}[b]{0.24\textwidth}
        \centering
        \includegraphics[width=\textwidth]{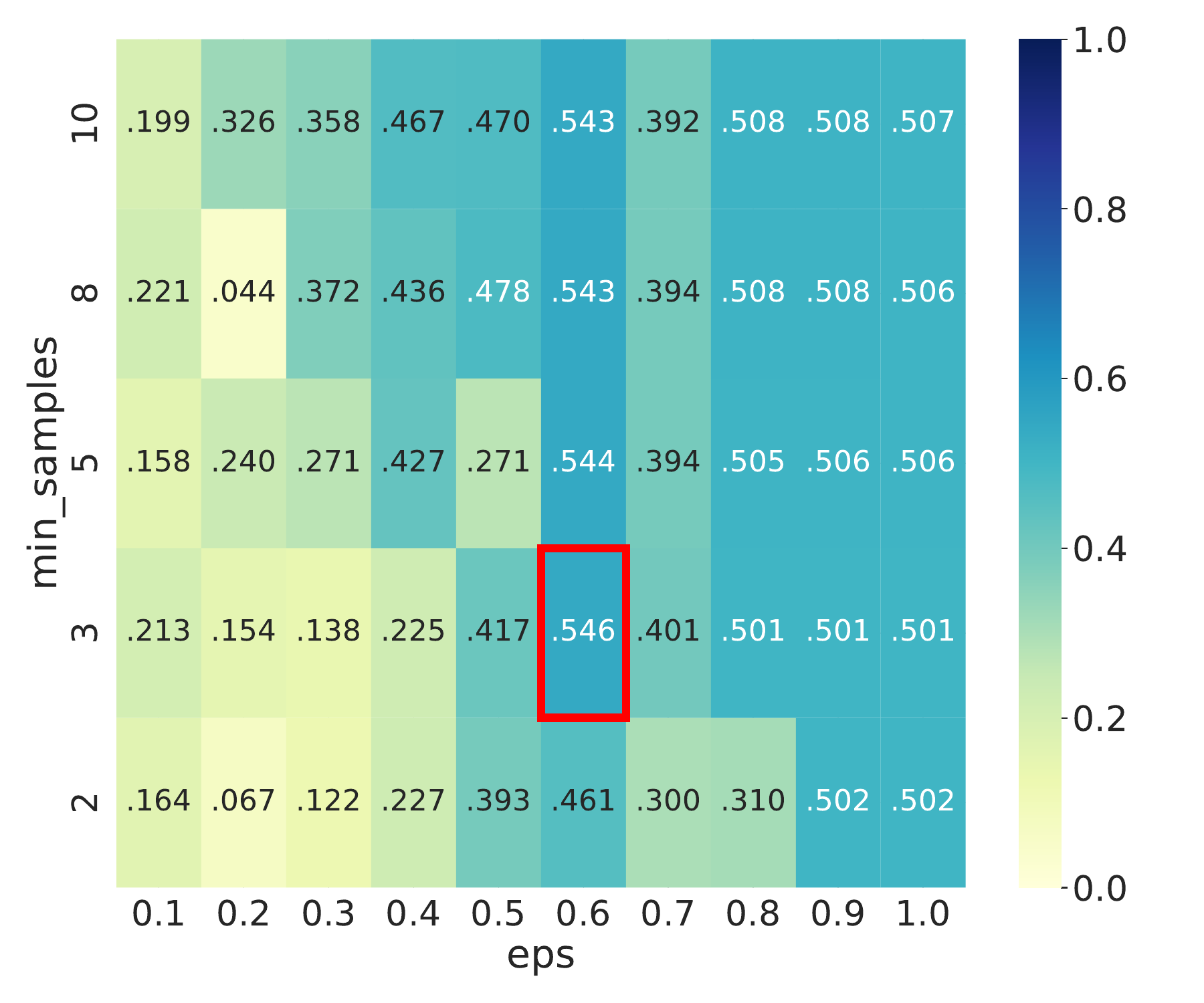}
        \caption{Level 10.}
        \label{fig:d_lvl10}
    \end{subfigure}
    \begin{subfigure}[b]{0.24\textwidth}
        \centering
        \includegraphics[width=\textwidth]{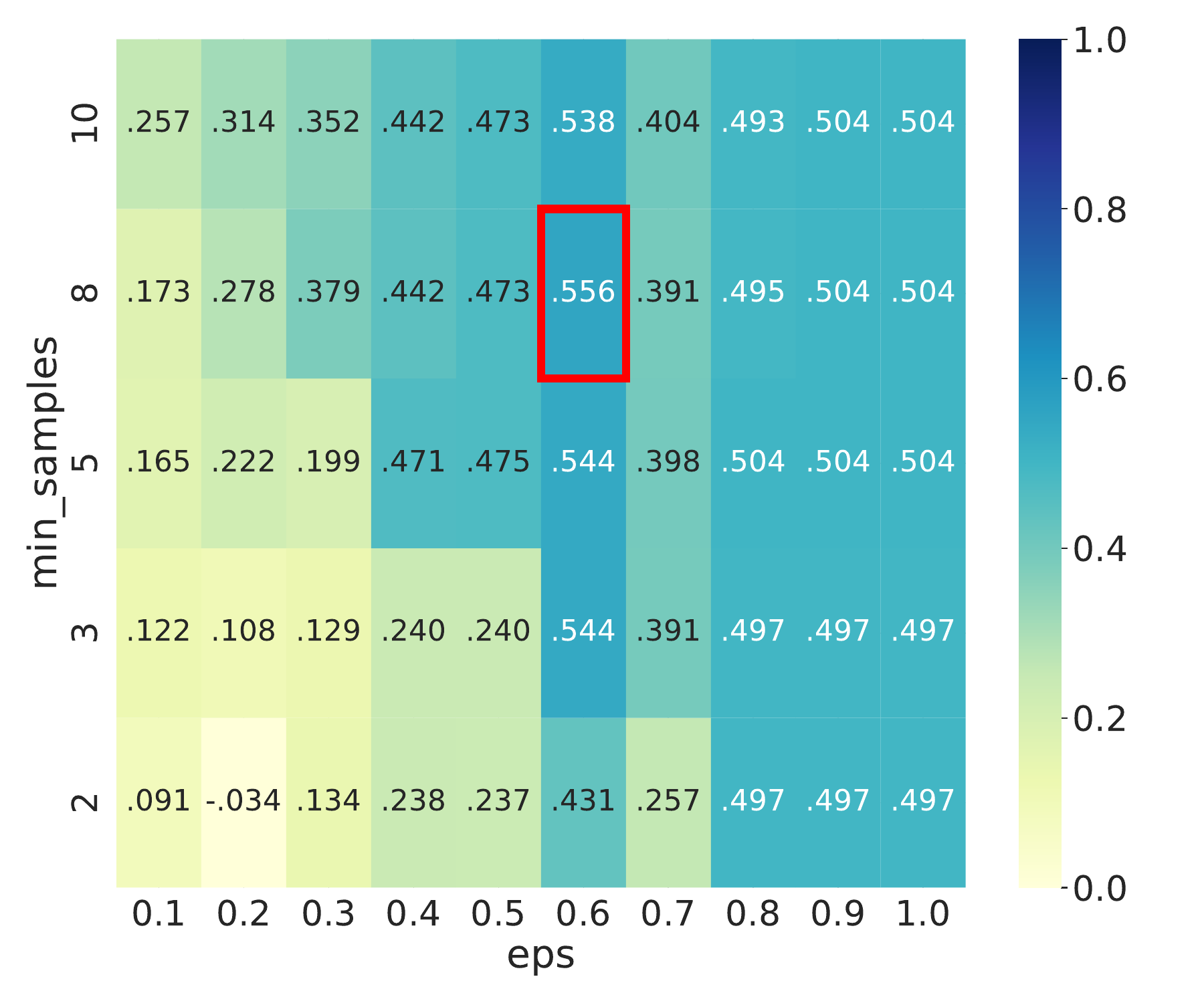}
        \caption{Level 11.}
        \label{fig:d_lvl11}
    \end{subfigure}
    \begin{subfigure}[b]{0.24\textwidth}
        \includegraphics[width=\textwidth]{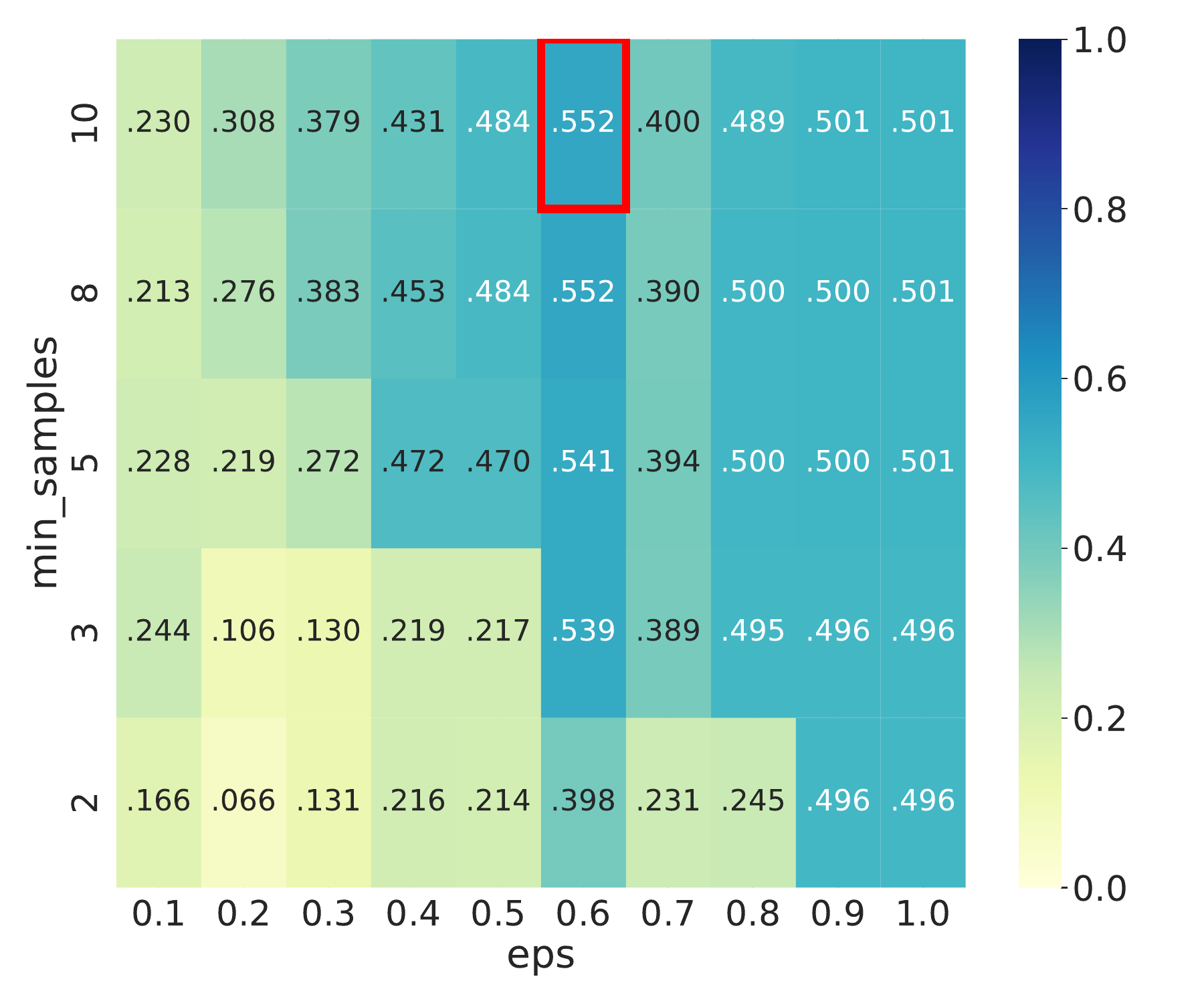}
        \caption{Level 12.}
        \label{fig:d_lvl12}
    \end{subfigure}
    \begin{subfigure}[b]{0.24\textwidth}
        \centering
        \includegraphics[width=\textwidth]{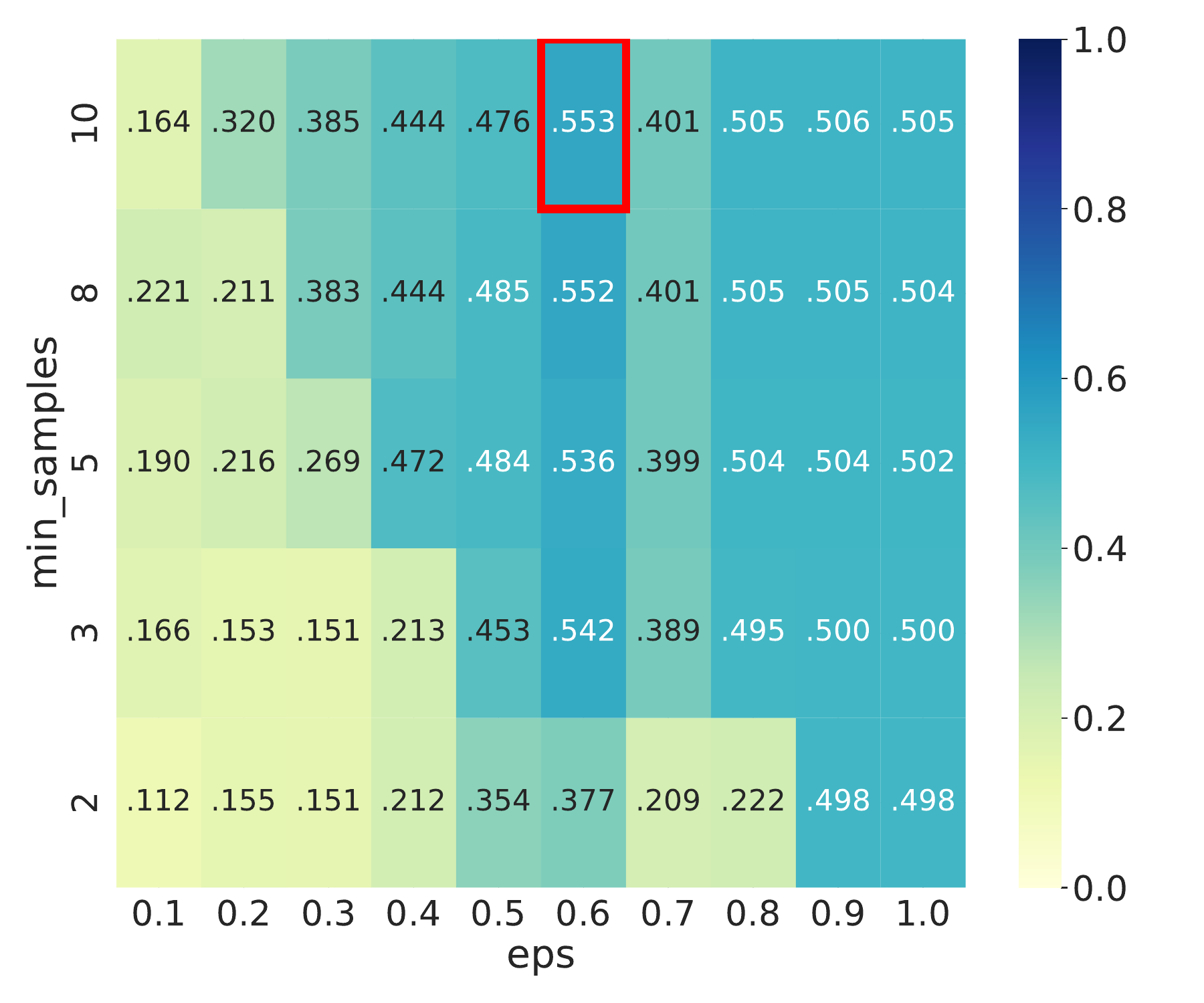}
        \caption{Level 13.}
        \label{fig:d_lvl13}
    \end{subfigure}
    \begin{subfigure}[b]{0.24\textwidth}
        \centering
        \includegraphics[width=\textwidth]{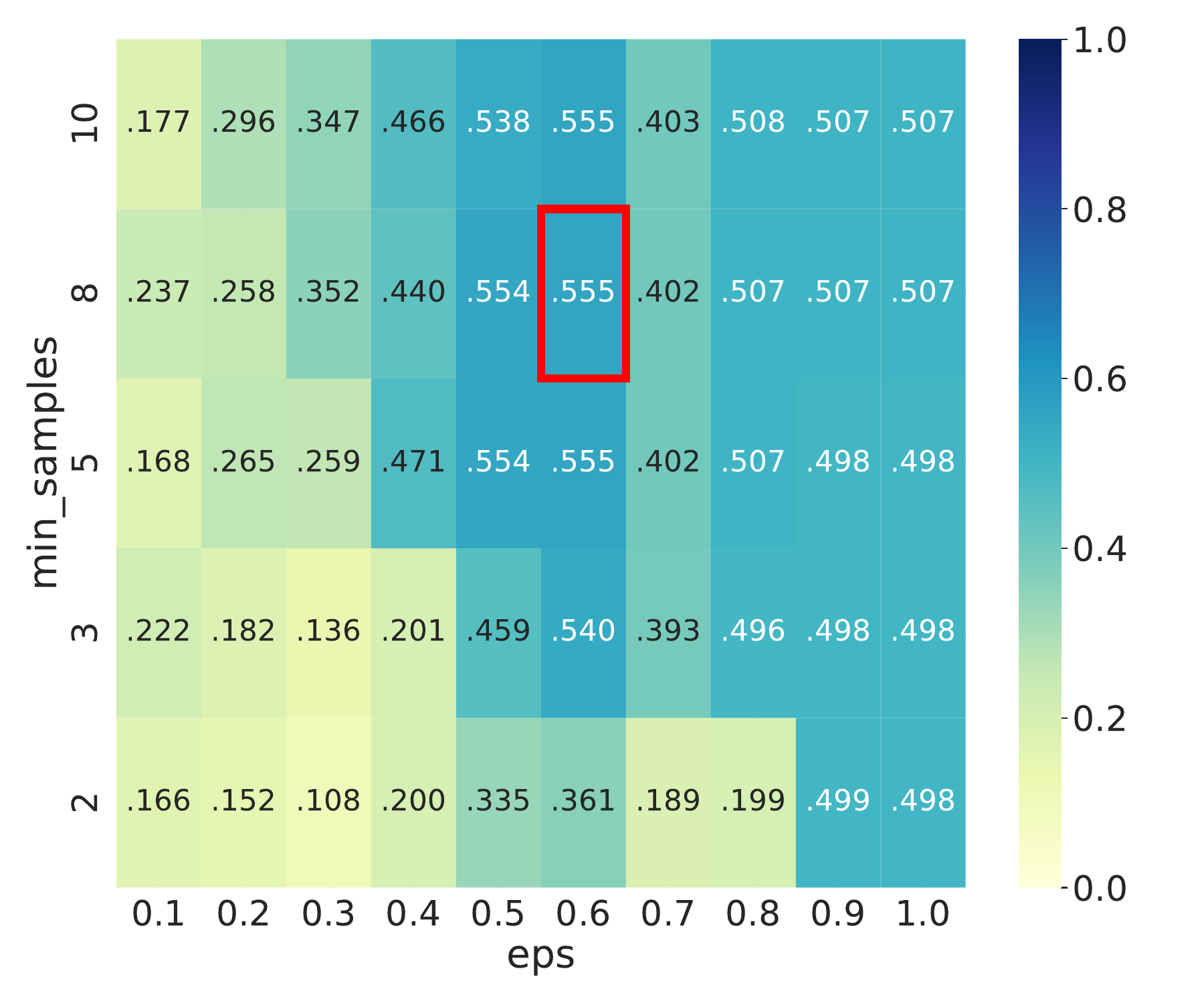}
        \caption{Level 14.}
        \label{fig:d_lvl14}
    \end{subfigure}
    \begin{subfigure}[b]{0.24\textwidth}
        \centering
        \includegraphics[width=\textwidth]{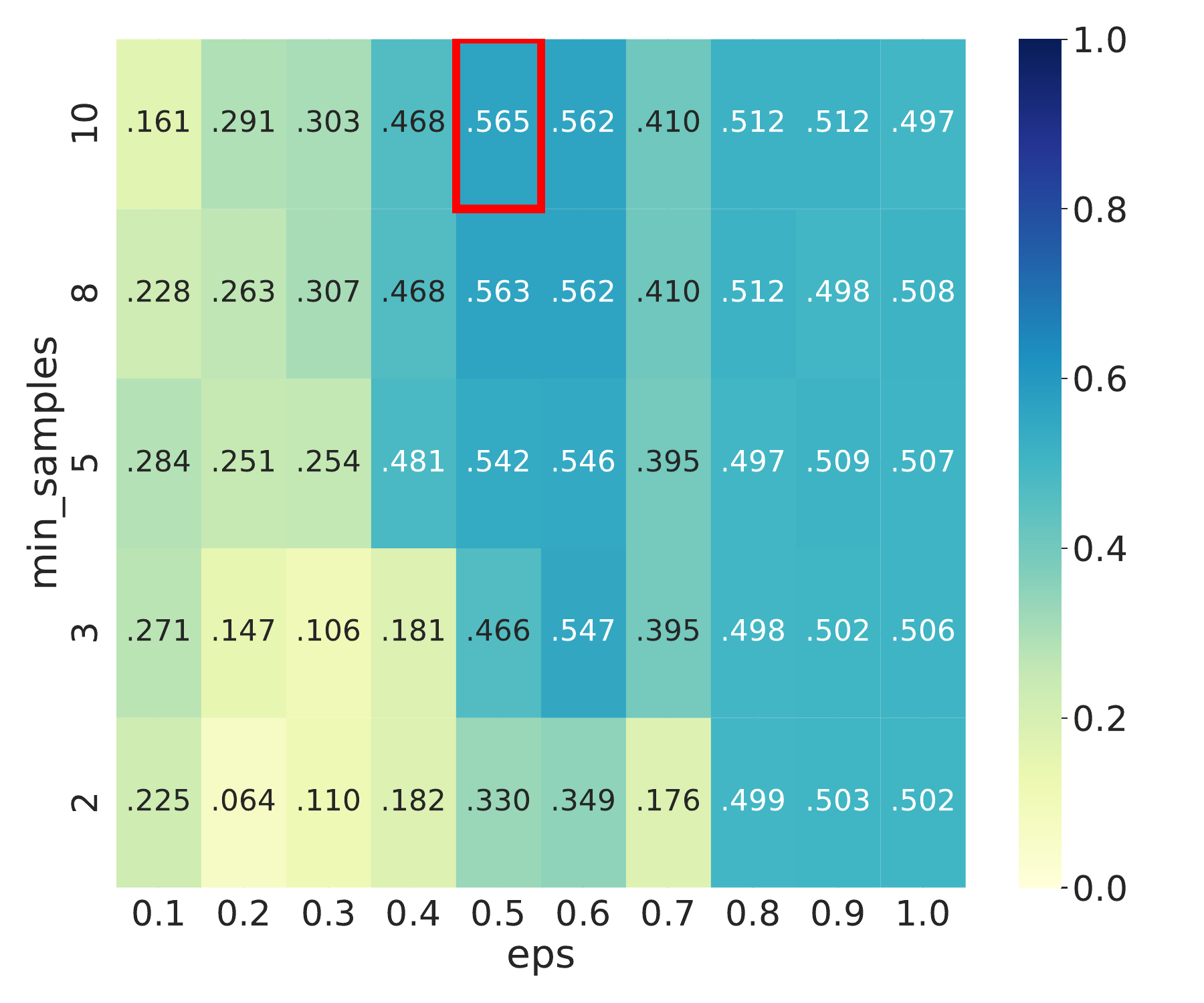}
        \caption{Level 15.}
        \label{fig:d_lvl15}
    \end{subfigure}
    \centering
    \caption{DBSCAN $\mathcal{S}_{score}$ during the grid-search using only structural features.}
    \label{fig:dbscan_gridsearch}
\end{figure*}

\begin{figure*}[!htbp]
    \centering
    \begin{subfigure}[b]{0.24\textwidth}
        \includegraphics[width=\textwidth]{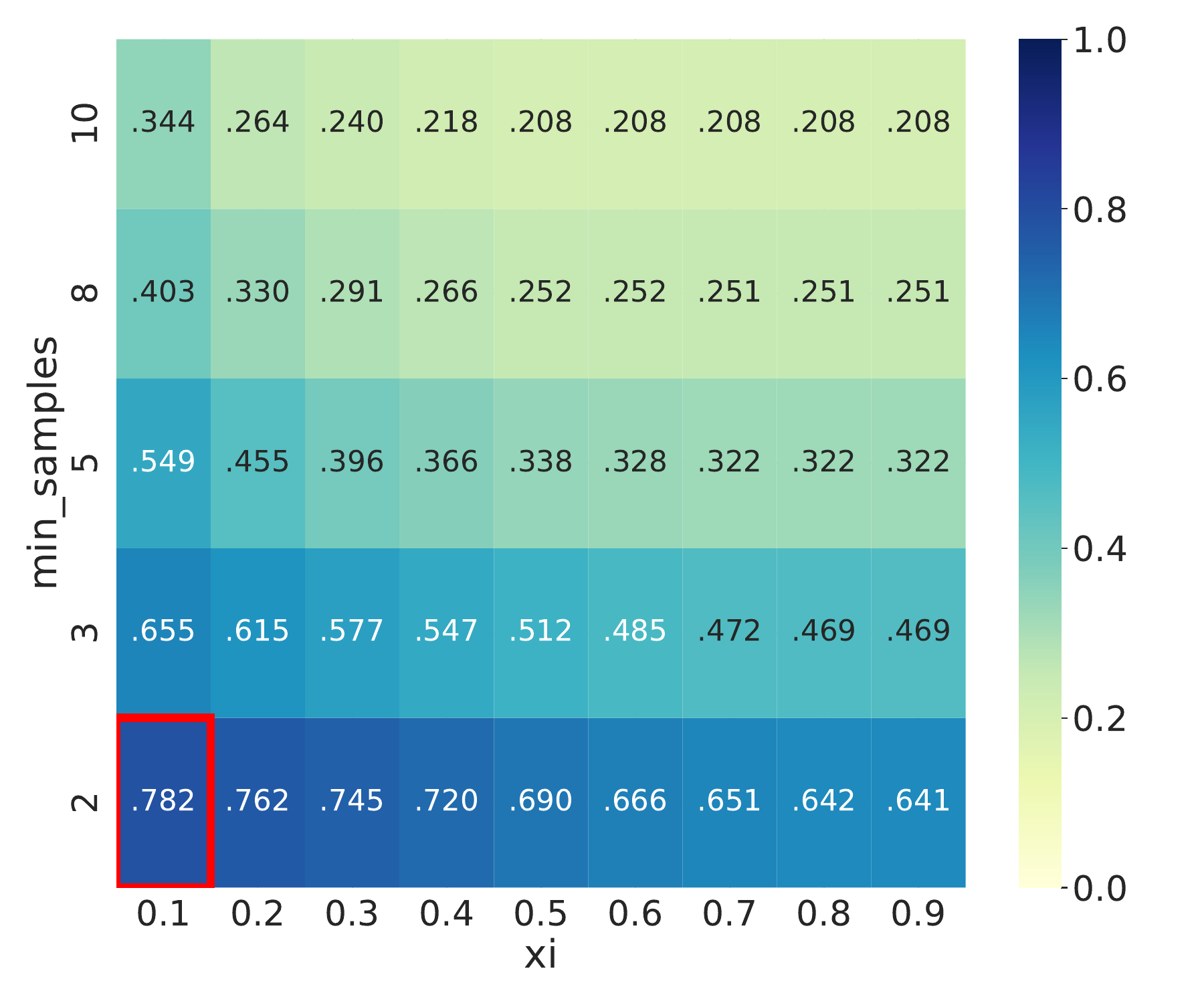}
        \caption{Level 8.}
        \label{fig:o_lvl8}
    \end{subfigure}
    \begin{subfigure}[b]{0.24\textwidth}
        \centering
        \includegraphics[width=\textwidth]{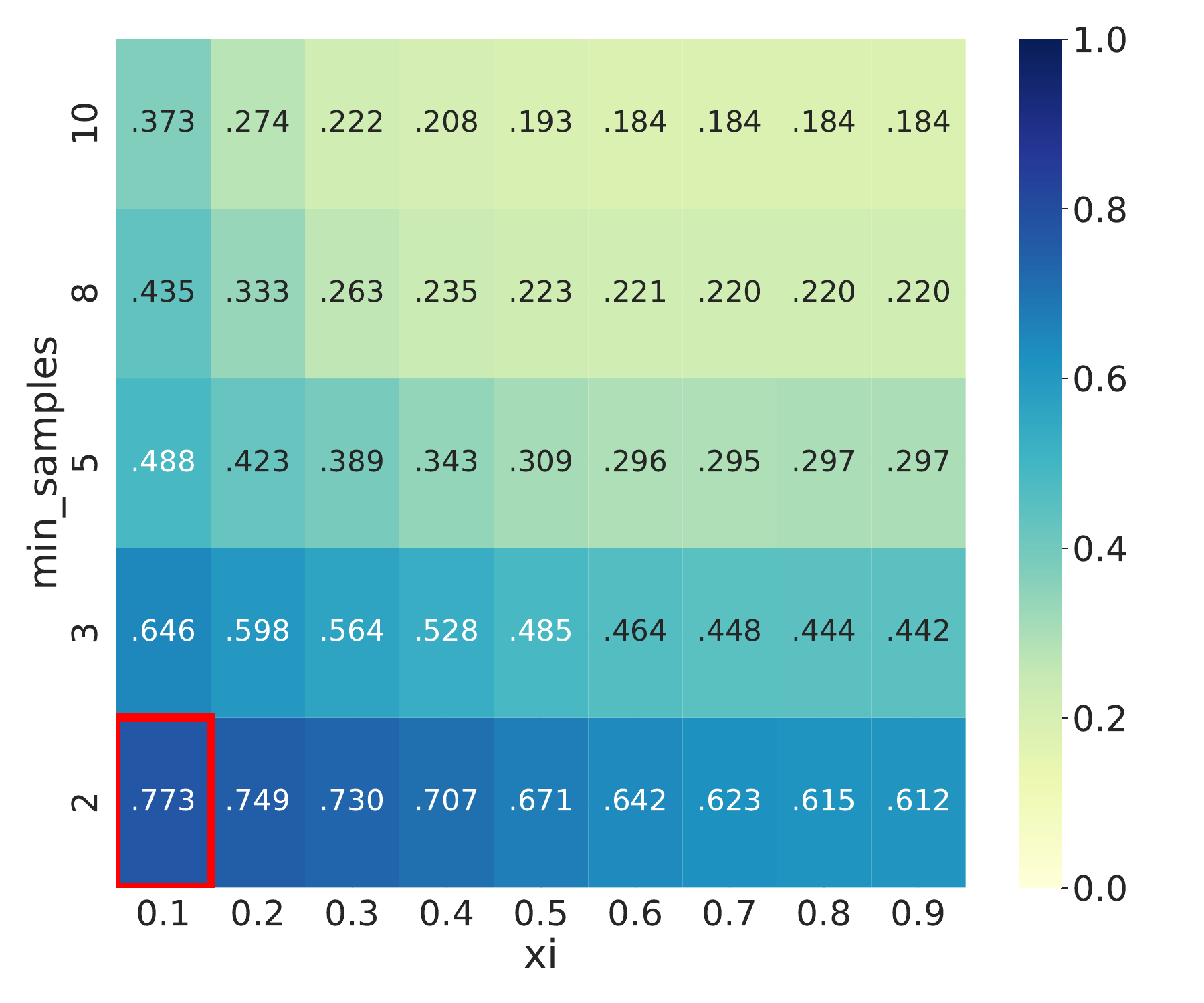}
        \caption{Level 9.}
        \label{fig:o_lvl9}
    \end{subfigure}
    \begin{subfigure}[b]{0.24\textwidth}
        \centering
        \includegraphics[width=\textwidth]{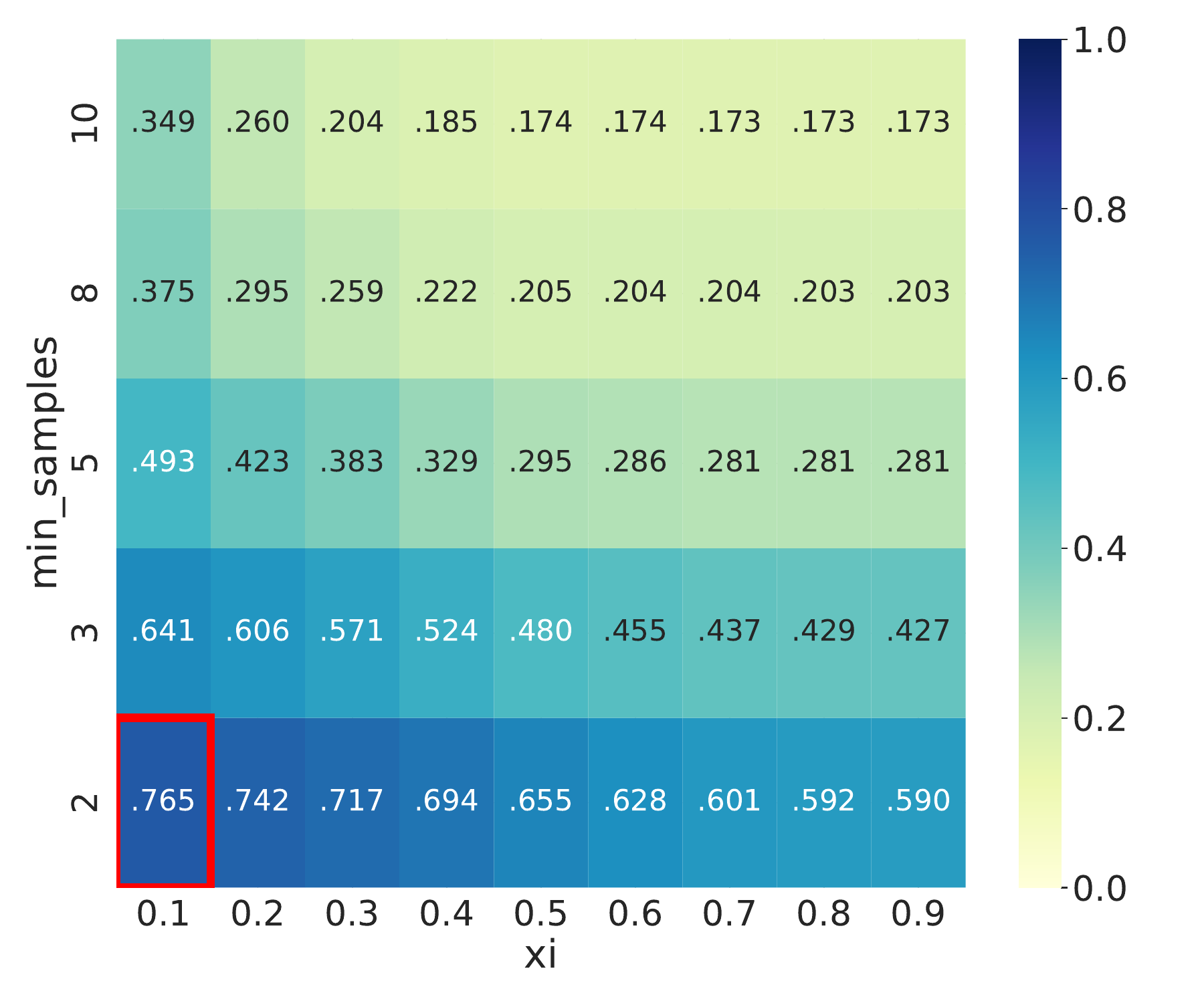}
        \caption{Level 10.}
        \label{fig:o_lvl10}
    \end{subfigure}
    \begin{subfigure}[b]{0.24\textwidth}
        \centering
        \includegraphics[width=\textwidth]{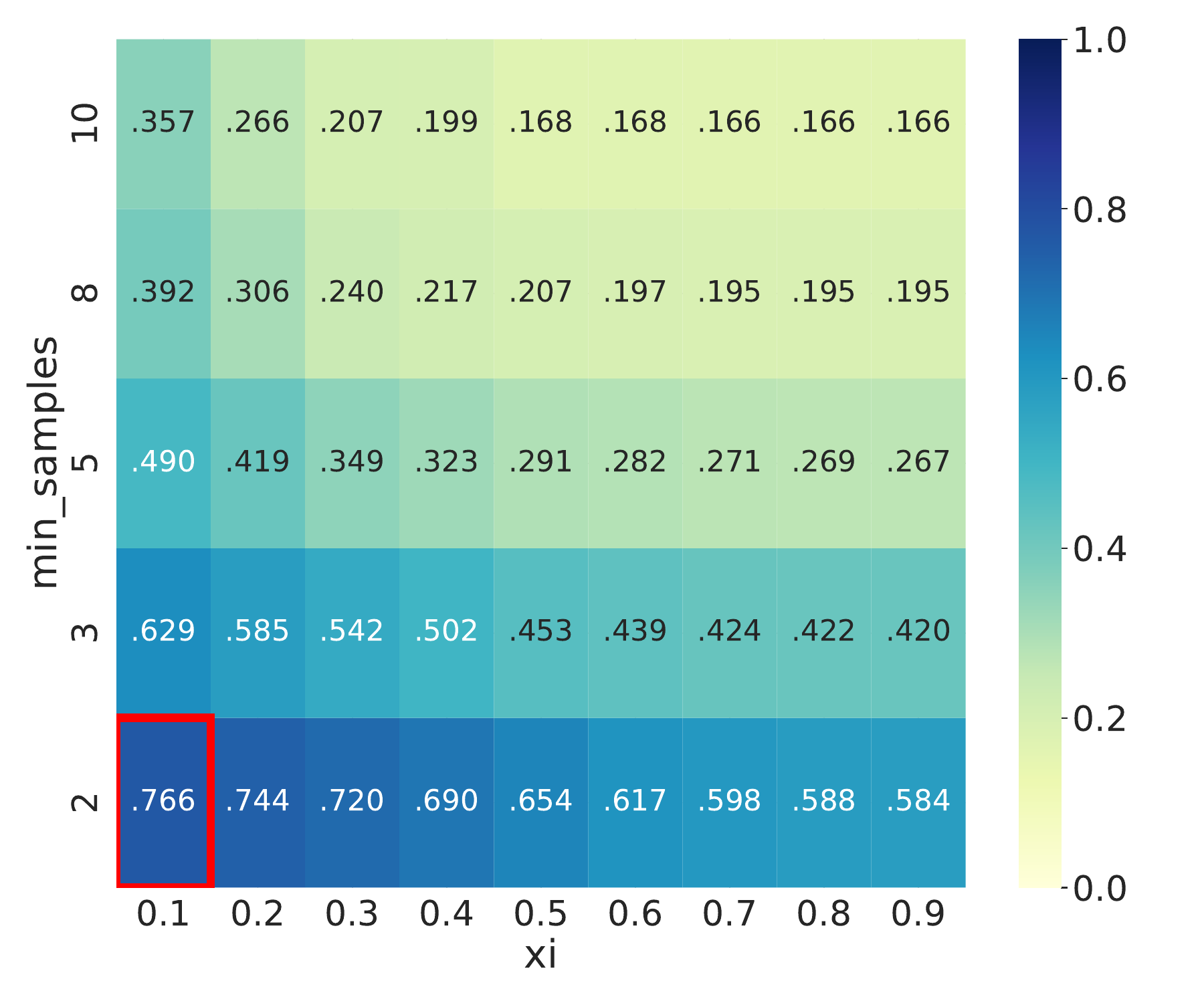}
        \caption{Level 11.}
        \label{fig:o_lvl11}
    \end{subfigure}
    \begin{subfigure}[b]{0.24\textwidth}
        \includegraphics[width=\textwidth]{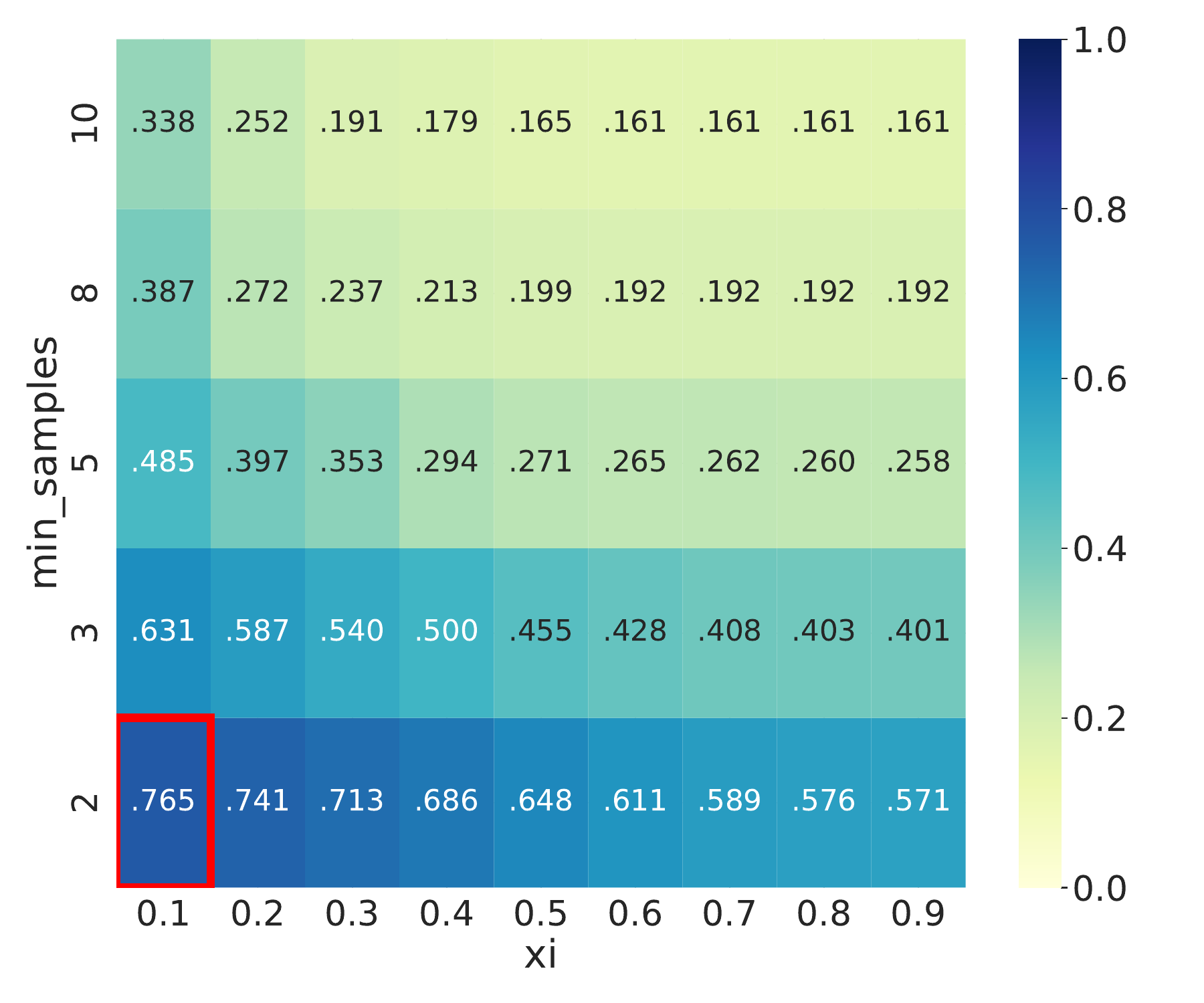}
        \caption{Level 12.}
        \label{fig:o_lvl12}
    \end{subfigure}
    \begin{subfigure}[b]{0.24\textwidth}
        \centering
        \includegraphics[width=\textwidth]{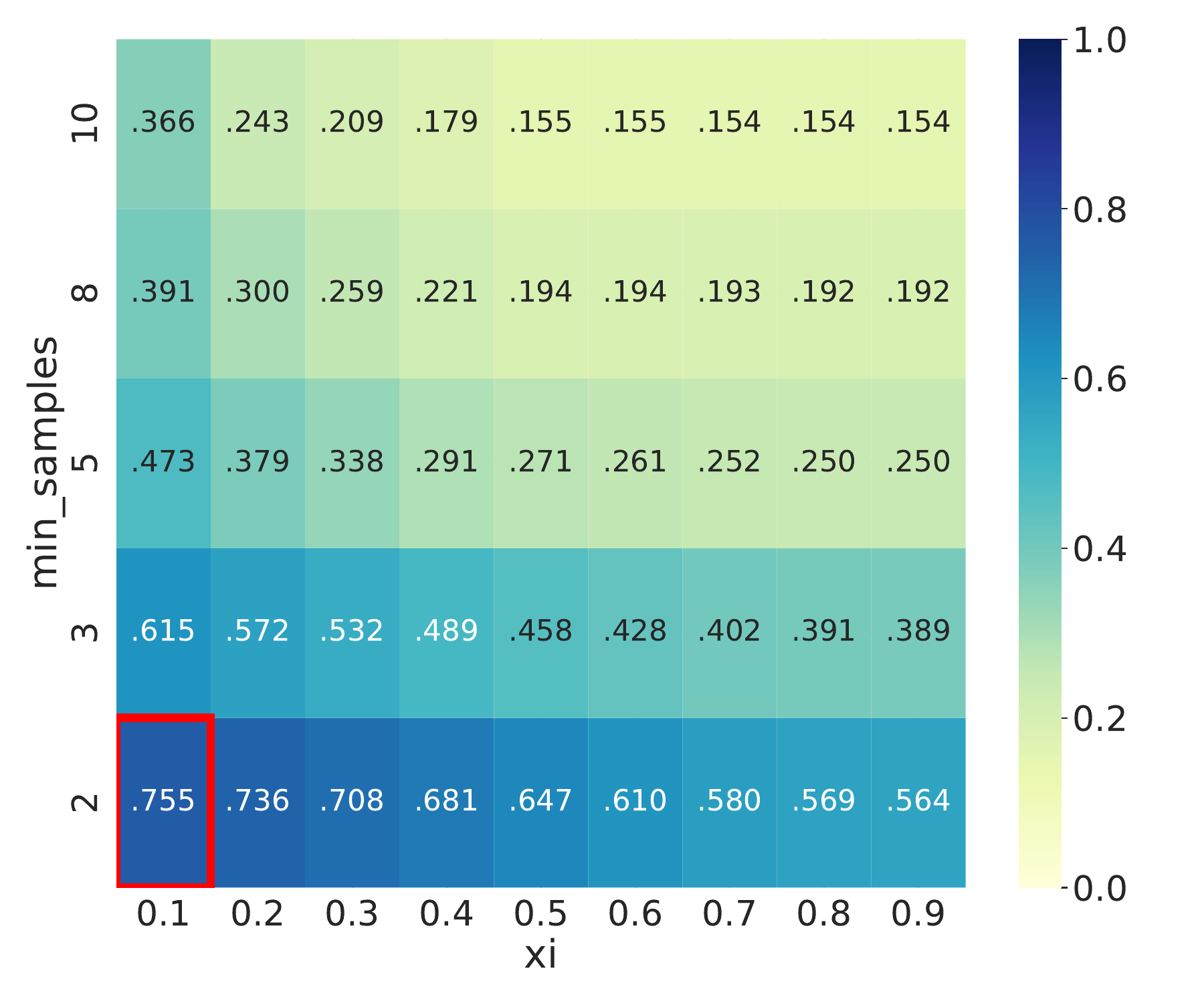}
        \caption{Level 13.}
        \label{fig:o_lvl13}
    \end{subfigure}
    \begin{subfigure}[b]{0.24\textwidth}
        \centering
        \includegraphics[width=\textwidth]{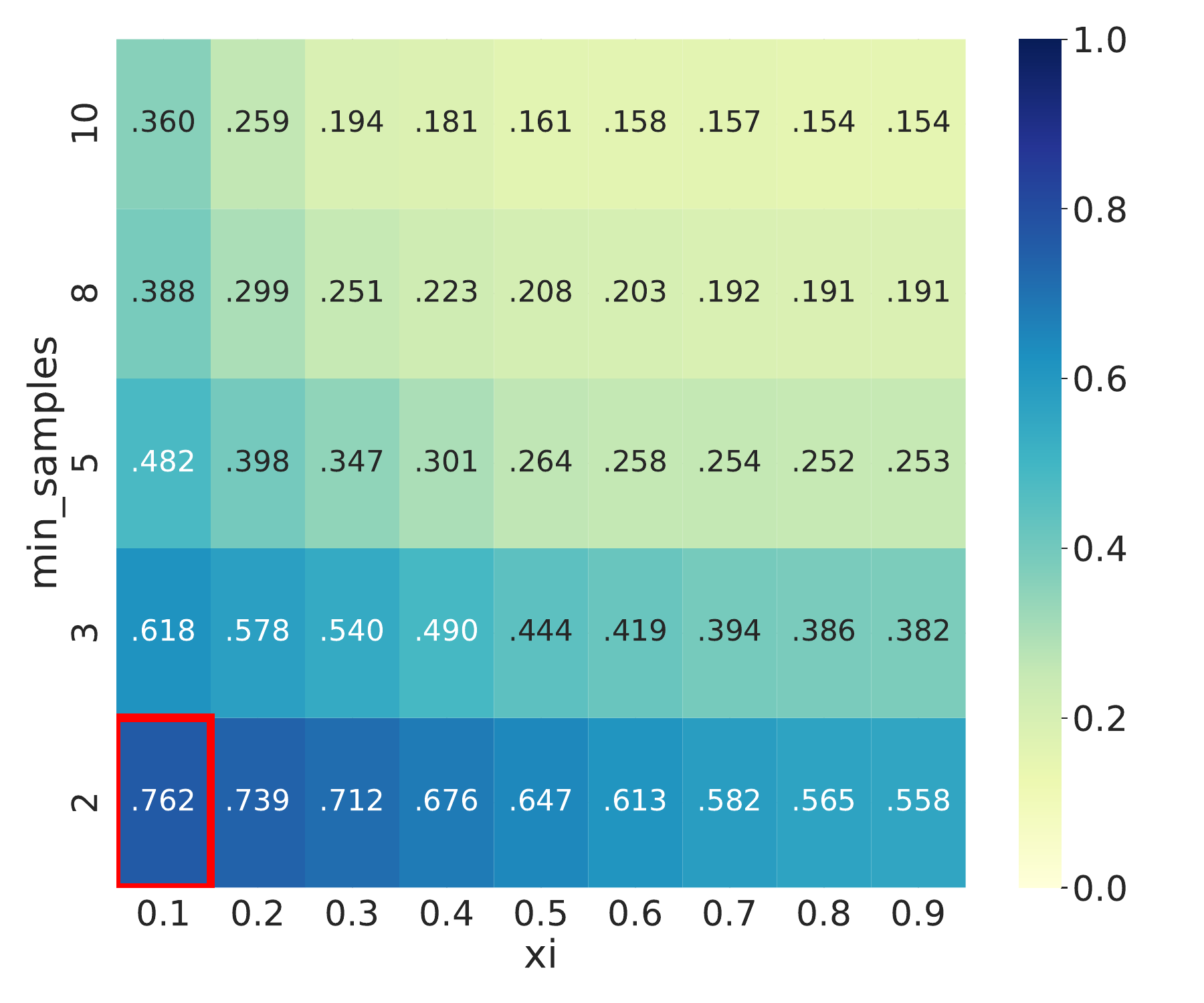}
        \caption{Level 14.}
        \label{fig:o_lvl14}
    \end{subfigure}
    \begin{subfigure}[b]{0.24\textwidth}
        \centering
        \includegraphics[width=\textwidth]{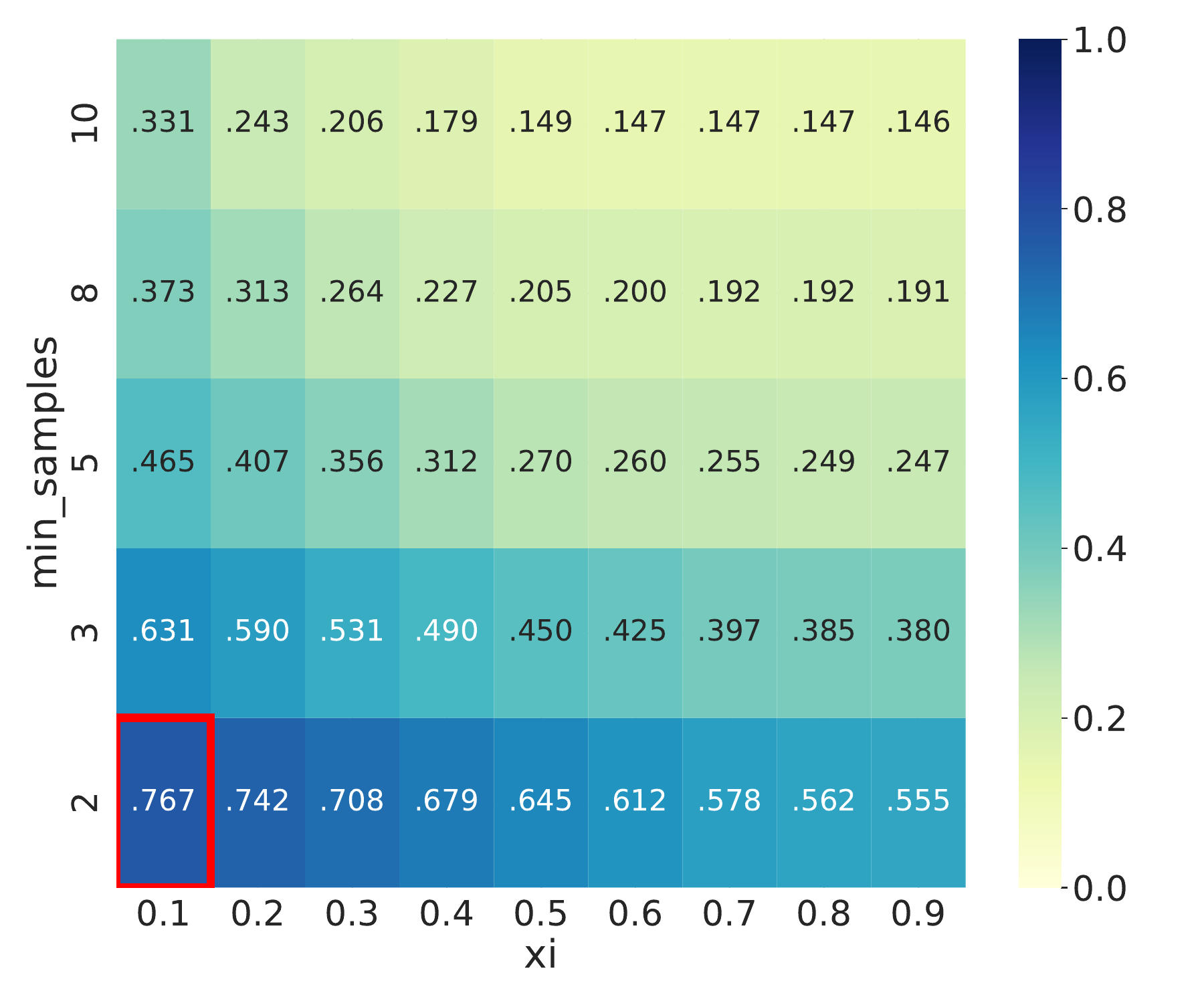}
        \caption{Level 15.}
        \label{fig:o_lvl15}
    \end{subfigure}
    \centering
    \caption{OPTICS $\mathcal{S}_{score}$ during the grid-search using only structural features.}
    \label{fig:optics_gridsearch}
\end{figure*}

\begin{figure}[!htbp]
    \centering
        \includegraphics[width=0.9\linewidth]{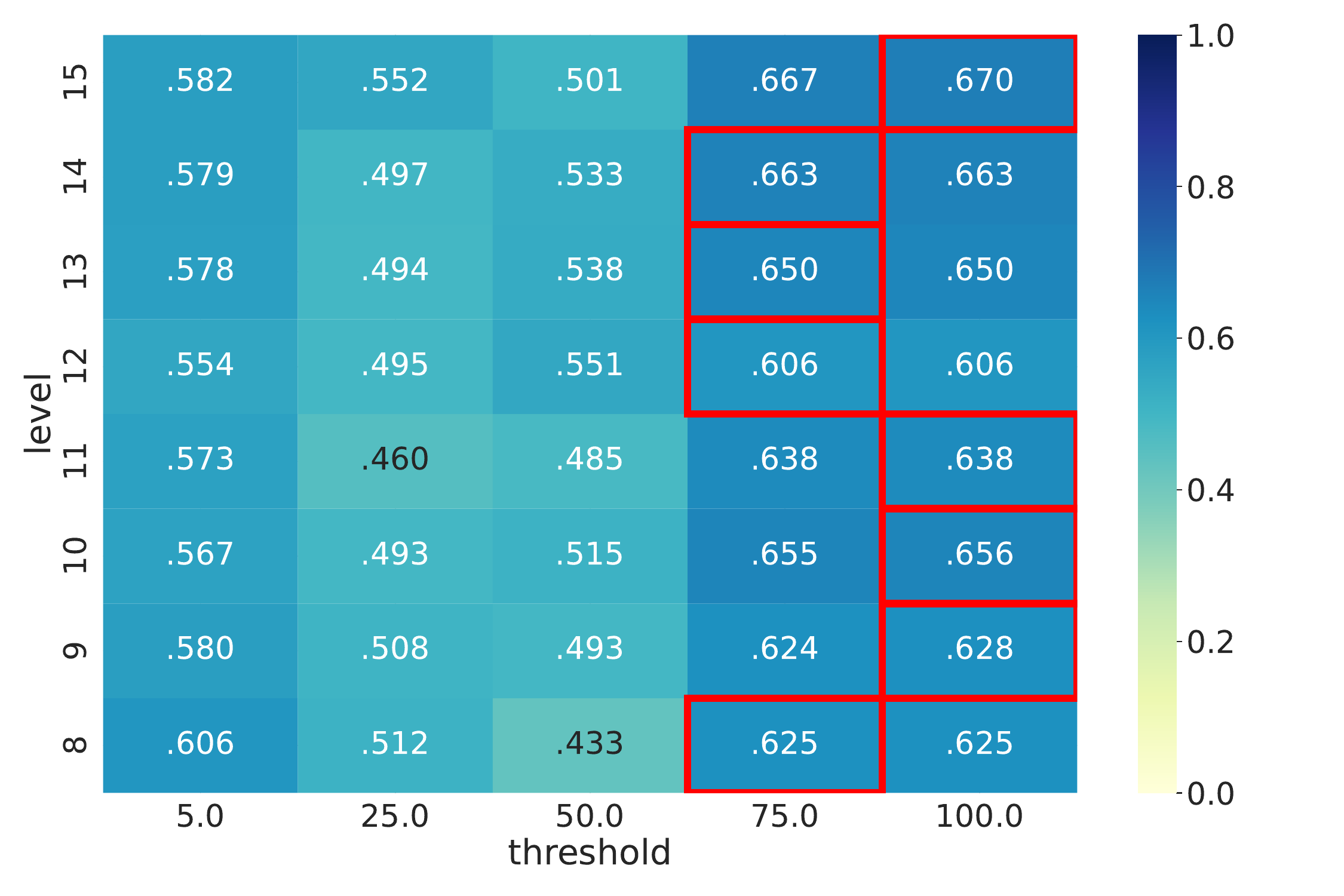}
    \caption{AHC $\mathcal{S}_{score}$ during the grid-search using only structural features.}
    \label{fig:ahcdhc_gridsearch}
\end{figure}

Figure \ref{fig:dbscan_gridsearch} reports the DBSCAN performance, demonstrating a consistent behavior of the algorithm across most tree depths. In fact, the highest $\mathcal{S}_{score}$ is obtained for a fixed value of $\epsilon = 0.6$ at almost all levels (8, 9, 11, 12, and 13). Deviations from this pattern occur only at levels 10 and 14, where the peak shifts toward lower-density configurations despite maintaining $\epsilon = 0.6$. At level 15, however, the maximum $\mathcal{S}_{score}$ is achieved when the neighborhood radius decreases to $\epsilon = 0.5$. Considering the \texttt{min\_samples} parameter, the highest scores are generally obtained with larger values (\texttt{min\_samples} = 8 or 10) across all levels except level 10, where the optimal \texttt{min\_samples} decreases to 3.

Figure \ref{fig:optics_gridsearch} shows the OPTICS performance, highlighting that, across all pruning stages, the highest value is invariably achieved with the lowest density value (\texttt{min\_samples}= 2). Conversely, looking at the $xi$ parameter, it concentrates at $0.1$ in all stages excluding level 10, whereas the xi value increases to $0.2$.

AHC grid-search results are reported in Figure \ref{fig:ahcdhc_gridsearch}. Specifically, the figure shows consistent behavior in the algorithm performance, indicating that across all levels, the highest $\mathcal{S}_{score}$ is achieved with the same threshold $t=75$.

\subsection{Structural and content-based features contribution analysis}\label{subsec:compare}

Once the best configurations are identified through the grid-search analysis conducted in Section \ref{subsec:gridsearch} (and Appendix \ref{sub:app1}), the clustering metrics presented in Section \ref{subsec:modelsconfig} for each algorithm across all structural depths are summarized in Table \ref{tab:exp}. Specifically, the first table reports the results when only structural properties are used, while the latter reports the results when content-based information is also included.

\begin{table*}[htbp]
\centering
\begin{subtable}{\linewidth}
\resizebox{\linewidth}{!}{%
\begin{tabular}{@{}cccccccccccccccc@{}}
\toprule
 & \multicolumn{5}{c}{\textbf{DBSCAN}} & \multicolumn{5}{c}{\textbf{OPTICS}} & \multicolumn{5}{c}{\textbf{AHC}} \\
\cmidrule(lr){2-6} \cmidrule(lr){7-11} \cmidrule(lr){12-16}
\textbf{Level} & $N_{c}$ & $N_{o\%}$ & $\mathcal{S}_{score}$ & $\mathcal{DB}_{index}$ & $\mathcal{CH}_{index}$ &
$N_{c}$ & $N_{o\%}$ &  $\mathcal{S}_{score}$ & $\mathcal{DB}_{index}$ & $\mathcal{CH}_{index}$ &
$N_{c}$ & $S_{i\%}$ &  $\mathcal{S}_{score}$ & $\mathcal{DB}_{index}$ & $\mathcal{CH}_{index}$ \\
\midrule

8  & 7 & 1.04\% & 0.545 & 0.484 & 732 & 551 & 7.34\% & \textbf{0.782} & 1.696 & 29 & 3 & 0.0\% & 0.625 & 0.575 & 3,487 \\

9  & 7 & 0.91\% & 0.556 & 0.464 & 778 & 566 & 7.34\% & 0.773 & 1.605 & 45 & 3 & 0.0\% & 0.628 & 0.569 & 3,376 \\

10 & 10 & 0.20\% & 0.546 & 0.924 & 650 & 567 & 7.11\% & 0.765 & \textbf{1.448} & 181 & 3 & 0.0\% & 0.656 & 0.566 & 3,435 \\

11 & 7 & 0.68\% & 0.556 & 0.492 & 787 &\cellcolor{yellow}574 &\cellcolor{yellow}6.89\% &\cellcolor{yellow}0.766 &\cellcolor{yellow}1.502 &\cellcolor{yellow}\textbf{191} & 3 & 0.0\% & 0.638 & 0.568 & 3,558 \\

12 & 7 & 0.68\% & 0.552 & 0.461 & 786 & 583 & 6.86\% & 0.765 & 1.545 & 172 & 3 & 0.0\% & 0.606 & 0.622 & 3,457 \\

13 & 7 & 0.63\% & 0.553 & \textbf{0.444 }& 792 & 578 & 7.50\% & 0.755 & 1.525 & 114 & 3 & 0.0\% & 0.650 & 0.600 & 3,828 \\

14 & \cellcolor{yellow}7 & \cellcolor{yellow}0.52\% & \cellcolor{yellow}0.555 & \cellcolor{yellow}0.475 & \cellcolor{yellow}\textbf{794} & 591 & 7.07\% & 0.762 & 1.576 & 96 & \cellcolor{yellow}4 & \cellcolor{yellow}0.0\% & \cellcolor{yellow}0.663 & \cellcolor{yellow}\textbf{0.432} & \cellcolor{yellow}\textbf{4,450} \\

15 & 7 & 1.04\% & \textbf{0.565} & 0.559 & 753 & 593 & 6.66\% & 0.767 & 1.547 & 132 & 4 & 0.0\% & \textbf{0.670} & 0.521 & 3,696 \\

\bottomrule
\end{tabular}}
\caption{Only structural properties results.}
\label{tab:sensibilidad_clustering}
\end{subtable}

\vspace{0.3cm}

\begin{subtable}{\linewidth}
\centering
\resizebox{\linewidth}{!}{%
\begin{tabular}{@{}cccccccccccccccc@{}}
\toprule
 & \multicolumn{5}{c}{\textbf{DBSCAN}} & \multicolumn{5}{c}{\textbf{OPTICS}} & \multicolumn{5}{c}{\textbf{AHC}} \\
\cmidrule(lr){2-6} \cmidrule(lr){7-11} \cmidrule(lr){12-16}
\textbf{Level} & $N_{c}$ & $N_{o\%}$ & $\mathcal{S}_{score}$ & $\mathcal{DB}_{index}$ & $\mathcal{CH}_{index}$ &
$N_{c}$ & $N_{o\%}$ & $\mathcal{S}_{score}$ & $\mathcal{DB}_{index}$ & $\mathcal{CH}_{index}$ &
$N_{c}$ & $S_{i\%}$ & $\mathcal{S}_{score}$ & $\mathcal{DB}_{index}$ & $\mathcal{CH}_{index}$ \\
\midrule

8  & \cellcolor{yellow}287 & \cellcolor{yellow}20.64\% & \cellcolor{yellow}\textbf{0.591} & \cellcolor{yellow}\textbf{1.908} & \cellcolor{yellow}\textbf{8.63} & \cellcolor{yellow}434 & \cellcolor{yellow}20.53\% & \cellcolor{yellow}\textbf{0.639} & \cellcolor{yellow}1.981 & \cellcolor{yellow}\textbf{7.21} & 900 & 65.22\% & \textbf{0.715} & 0.316 & 1,455 \\

9  & 293 & 21.82\% & 0.583 & 1.944 & 7.77 & 463 & 23.77\% & 0.557 & \textbf{1.820} & 6.23 & 944 & 66.84\% & 0.711 & 0.292 & 1,661 \\

10 & 295 & 22.00\% & 0.581 & 1.996 & 7.22 & 468 & 24.15\% & 0.550 & 2.015 & 5.80 & 966 & 68.12\% & 0.707 & 0.285 & 1,879 \\

11 & 299 & 22.32\% & 0.581 & 2.056 & 6.67 & 486 & 23.77\% & 0.544 & 2.066 & 5.58 & 992 & 69.25\% & 0.707 & 0.265 & 2,173 \\

12 & 298 & 22.72\% & 0.572 & 2.056 & 6.70 & 423 & 29.63\% & 0.488 & 2.000 & 5.17 & 1,009 & 69.87\% & 0.703 & 0.255 & 2,387 \\

13 & 302 & 23.24\% & 0.568 & 2.074 & 6.21 & 389 & 26.26\% & 0.571 & 2.074 & 5.32 & 1,028 & 69.46\% & 0.707 & 0.247 & 2,532 \\

14 & 309 & 23.58\% & 0.561 & 2.096 & 6.10 & 424 & 23.95\% & 0.598 & 2.158 & 5.26 & 1,047 & 69.34\% & 0.707 & 0.247 & 2,708 \\

15 & 307 & 23.81\% & 0.559 & 2.138 & 6.03 & 489 & 27.57\% & 0.488 & 2.090 & 4.80 & \cellcolor{yellow}1,062 & \cellcolor{yellow}69.77\% & \cellcolor{yellow}0.706 & \cellcolor{yellow}\textbf{0.243} & \cellcolor{yellow}\textbf{2,956} \\
\end{tabular}}
\caption{Structural and content-based properties results.}
\label{tab:sensibilidad_escenario2}
\end{subtable}
\caption{Metrics of the best configuration of each algorithms for each level. $N_{c}$ indicates the number of generated clusters; $N_{o\%}$ the number of samples in percentage clustered as noisy points; $S_{i\%}$ the number of cluster in percentage that have only one sample within. For each metric and each algorithm, the best metric value is shown in bold, while the model selected as the best performer is highlighted in yellow.}
\label{tab:exp}
\end{table*}

As shown in Table \ref{tab:sensibilidad_clustering}, OPTICS achieves the highest performance according to $\mathcal{S}_{score}$, obtaining values greater than 0.75 across all eight evaluated levels. However, it performs worst with respect to $\mathcal{DB}_{index}$ and $\mathcal{CH}_{index}$. In contrast, DBSCAN and AHC achieve lower $\mathcal{DB}_{index}$ values, while AHC also attains higher $\mathcal{CH}_{index}$ values.
Yet, OPTICS yields a relatively large number of clusters and a higher proportion of outliers ($N_{o\%} \in$ [6.66\%, 7.34\%]). On the other hand, DBSCAN and AHC produce only a few clusters: the DBSCAN with very low proportions of outliers ($\leq$ 1.04\%); and the AHC with no singletons (clusters with just one element).

\begin{table*}[!htbp]
\centering
\begin{tabular}{c|ccc|ccc|c}
\hline
\textbf{Algorithms} & \textbf{Features} & \textbf{Parameters} & \textbf{Level} &\textbf{$\mathcal{LJD}_{mean}$} & \textbf{$\mathcal{LJD}_{std}$} & \textbf{$\mathcal{LJD}_{max}$} & \textbf{\begin{tabular}[c]{@{}c@{}}\# Cluster with\\$\mathcal{LJD(C)}=0$\end{tabular}} \\ \hline
\multirow{2}{*}{\textit{DBSCAN}}    & structural & $\epsilon=0.6$,  \texttt{min\_samples = 10}  & 14  & 0.903 & 0.052 & 0.963 & 0\\
 & structural + content & $\epsilon=0.9$,  \texttt{min\_samples = 2}  & 8  & \textbf{0.158} & 0.254 & 0.979 & 133\\
 \hline
 \multirow{2}{*}{\textit{OPTICS}}    & structural & $xi=0.1$,  \texttt{min\_samples = 2}  & 11  & \textbf{0.592} & 0.368 & 1.093 & 87 \\
 & structural + content & $xi=0.3$,  \texttt{min\_samples = 2}  & 8  & 0.242 & 0.305 & 1.088 & 170\\
 \hline
 \multirow{2}{*}{\textit{AHC}}    & structural & $t=75.0$  & 14  & 0.901 & 0.047 & 0.959 & 0\\
 & structural + content & $t=5.0$  & 15 & 0.351 & 0.335 & 1.036 & 55\\
 \hline
\end{tabular}
\caption{Best configuration for each algorithm using structural and content-based information, and the corresponding results obtained during quantitative validation. In bold the best values in each case.}
\label{tab:bestconfig}
\end{table*}

When content-based features are incorporated into the clustering process, two distinct behaviors are observed (Table \ref{tab:sensibilidad_escenario2}). On the one hand, OPTICS exhibits a marked deterioration in clustering performance, reflected by higher $\mathcal{DB}_{index}$ values and lower $\mathcal{S}_{score}$ and $\mathcal{CH}_{index}$ values. In addition, the proportion of outliers increases considerably, ranging from approximately 20\% to 32\%. Similarly, DBSCAN exhibits poorer performance in terms of $\mathcal{DB}_{index}$ and $\mathcal{CH}_{index}$, while its $\mathcal{S}_{score}$ values slightly increased compared to those observed in the previous case. (Table \ref{tab:sensibilidad_clustering}). Furthermore, DBSCAN is now able to generate a considerable number of clusters ($N_{c} > 286$), albeit with a higher proportion of outliers (approximately 20-24\%).
On the other hand, AHC shows improvements in in $\mathcal{S}_{score}$ and $\mathcal{DB}_{index}$, with an increase of 0.046 and a decrease of 0.189 in their best values, respectively. Yet, AHC now produces approximately 65–70\% of clusters containing a single element.

A comprehensive analysis of Table \ref{tab:exp} reveals that the algorithms do not consistently achieve their best performance at a single configuration level. In fact, only in four cases do they achieve at least two best metrics at the same level, namely structural AHC at level 14, content-based DBSCAN at level 8, content-based OPTICS at level 8, and content-based AHC at level 15. In the remaining cases, the final level selection was based on a holistic assessment of all three metrics, prioritizing the overall trade-off among them rather than the best performance in any single metric.

Looking at these insights and the extracted level information, the results show that, when considering only structural properties, all the models perform better with a higher number of levels within the trees (14, 11, and 14). However, when content-based information is incorporated, DBSCAN and OPTICS tend to perform better with fewer levels (just 8). These results are somewhat expected since the content-based information acts as noisy information during the clustering process, increasing uncertainty in terms of clustering metrics.


\section{Cluster Validation}\label{sec:validation}

\subsection{Quantitative Validation}

Table \ref{tab:bestconfig} reports the level-wise Jaccard Distance Score computed for the best configuration of each algorithm, as introduced in Section \ref{subsec:validation}. The table shows that all three algorithms produce the lowest $\mathcal{LJD}_{mean}$ values when content features are used, with DBSCAN achieving the overall lowest value of 0.158. These lower values indicate that, on average, the ratio between cluster compactness and dispersion decreases when this additional information is incorporated. Thus, this suggests improved cluster separation among samples. However, considering $\mathcal{LJD}_{std}$, high variability is generally observed across all algorithms, except for DBSCAN and AHC using only structural features. In fact, in both cases, $\mathcal{LJD}_{mean}$ is similar to $\mathcal{LJD}_{max}$. Looking at these latter values ($\mathcal{LJD}_{max}$), the algorithms remain consistent, showing that in the worst case at least one cluster exhibits a ratio close to 1, meaning that samples within the cluster are approximately as distant from each other as they are from samples in different clusters (or even more so). Finally, the last column of Table \ref{tab:bestconfig} reports the number of clusters for which $\mathcal{LJD}(C)=0$, i.e., fully compact clusters where $\mathcal{LJD}_{\text{intra}}=0$, meaning that all samples within the cluster are structurally identical. Surprisingly, even when only structural features are considered during clustering, OPTICS is already able to identify 87 fully compact clusters. This suggests that structural information alone can be sufficient to group together subsets of webpages with identical level-wise DOM-tag distributions. When content-based features are incorporated, the number of fully compact clusters further increases, reaching 133 clusters for DBSCAN, 170 for OPTICS, and 55 for AHC. This result indicates that adding tag-level information strengthens the formation of highly homogeneous groups, although it also contributes to a more fragmented clustering structure.

\subsection{Qualitative Validation}

Figures \ref{fig:structural-features} and \ref{fig:structural-content_features} report the tree structures belonging to the three most compact clusters, i.e., the clusters with the lowest $\mathcal{LJD}(C)$ values, when considering structural features and structural-content features, respectively. To avoid presenting trivial cases composed of identical samples and to provide a more informative visual comparison, clusters with $\mathcal{LJD}(C)=0$ were excluded from the selection.

Figure \ref{fig:structural-features} shows the medoids and samples of the clusters obtained using only structural features, labeled as S1, S2, and S3. By inspecting the medoids, it can be observed that they share a similar structure in their \texttt{<head>} branch. Specifically, S2 and S3 exhibit the same tag structure. Moreover, S2 and S3 also share a similar structure within the \texttt{<body>} branch. However, they differ significantly in their last two levels, where S3 contains more nodes than S2.
Yet, looking at the samples extracted from each cluster, it possible to see that samples from S1 have the same structure of the medoid, due to the very low $\mathcal{LJD}$ (0.0022). On the other hand, for higher values, such as 0.0102 in S3, the samples show small difference with the medoids.

Figure \ref{fig:structural-content_features} reports the qualitative results obtained when content-based features are also included, labeled as SC1, SC2, and SC3. In this case, considering only the medoids, their structures differ substantially from one another, with SC2 exhibiting a particularly complex configuration. Nevertheless, in SC1, the medoid and the two selected samples are visually almost identical, differing only by a single change in the \texttt{<head>} branch. In contrast, SC2 is characterized by much larger and deeper trees. Despite their size, the medoid and the selected samples share a highly similar overall structure, mainly consisting of an extended horizontal organization with several repeated small branches. Finally, the samples belonging to SC3 differ in the number of nodes present in their deepest levels, as shown in Figure \ref{fig:structural-content_features}.

\setlength{\tabcolsep}{4pt}
\renewcommand{\arraystretch}{1.0}

\begin{figure*}[!ht]
    \centering

    {\small\bfseries S1 ($\mathcal{LJD}(C) = 0.0022$)\par}
    \vspace{0.15cm}

    {\small\bfseries Medoid\par}
    \vspace{0.15cm}

    \includegraphics[width=0.90\textwidth,keepaspectratio]
    {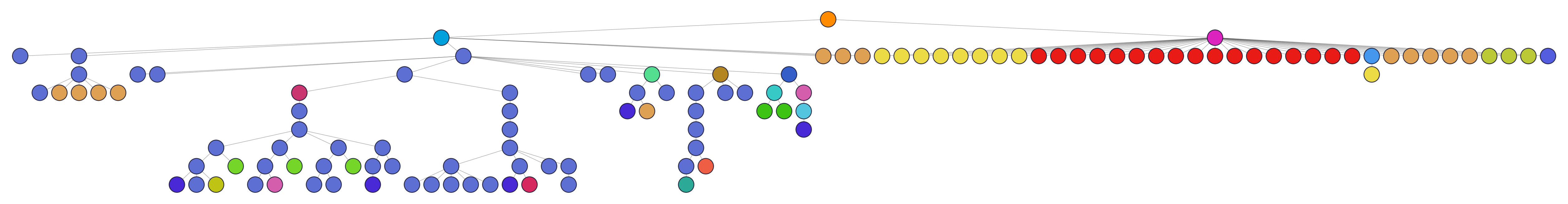}

    \par\vspace{0.25cm}

    {\small\bfseries Samples\par}
    \vspace{0.15cm}

    \begin{minipage}[t]{0.48\textwidth}
        \centering
        \includegraphics[width=\linewidth,keepaspectratio]
        {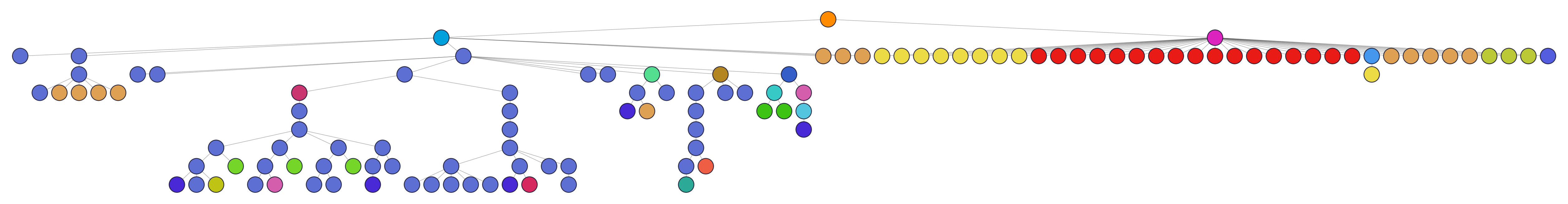}
    \end{minipage}
    \hfill
    \begin{minipage}[t]{0.48\textwidth}
        \centering
        \includegraphics[width=\linewidth,keepaspectratio]
        {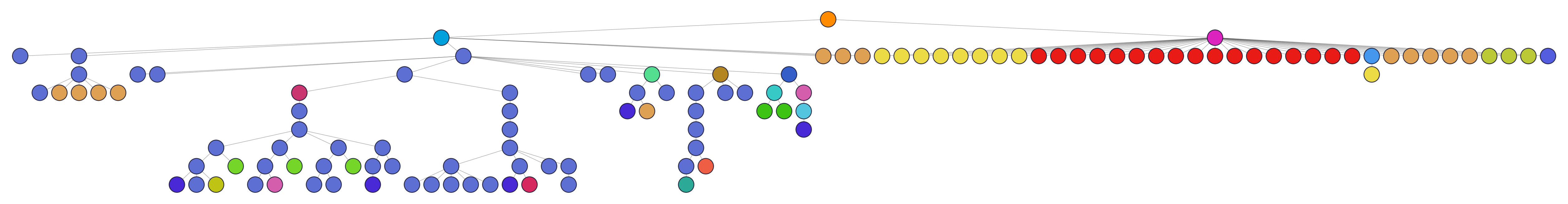}
    \end{minipage}

    \par\vspace{0.30cm}
    \hrule
    \vspace{0.30cm}

    {\small\bfseries S2 ($\mathcal{LJD}(C) = 0.0081$)\par}
    \vspace{0.15cm}

    {\small\bfseries Medoid\par}
    \vspace{0.15cm}

    \includegraphics[width=0.90\textwidth,keepaspectratio]
    {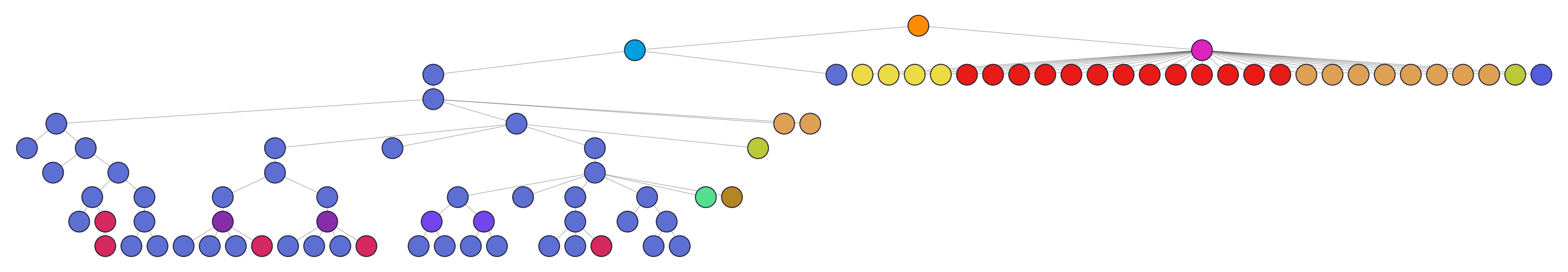}

    \par\vspace{0.25cm}

    {\small\bfseries Samples\par}
    \vspace{0.15cm}

    \begin{minipage}[t]{0.48\textwidth}
        \centering
        \includegraphics[width=\linewidth,keepaspectratio]
        {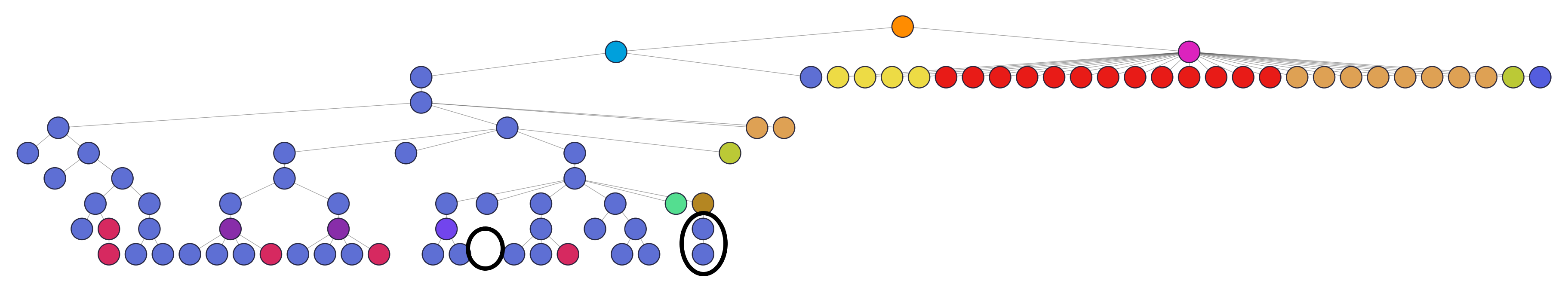}
    \end{minipage}
    \hfill
    \begin{minipage}[t]{0.48\textwidth}
        \centering
        \includegraphics[width=\linewidth,keepaspectratio]
        {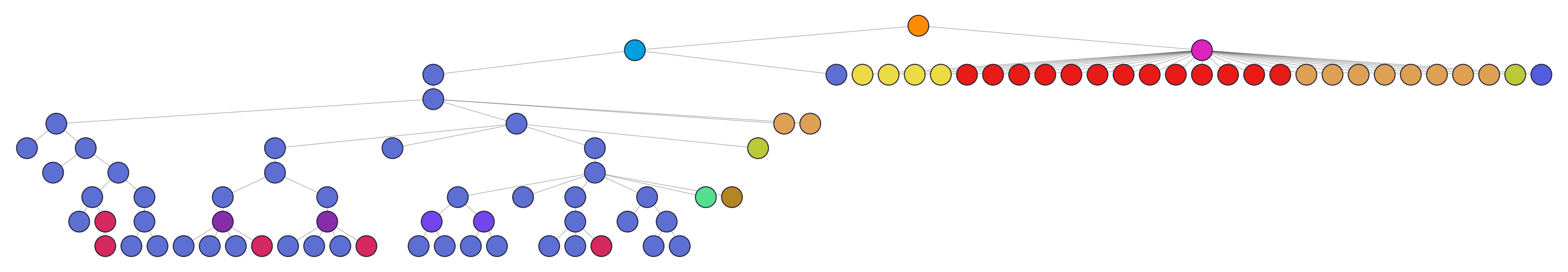}
    \end{minipage}

    \par\vspace{0.30cm}
    \hrule
    \vspace{0.30cm}

    {\small\bfseries S3 ($\mathcal{LJD}(C) = 0.0102$)\par}
    \vspace{0.15cm}

    {\small\bfseries Medoid\par}
    \vspace{0.15cm}

    \includegraphics[width=0.90\textwidth,keepaspectratio]
    {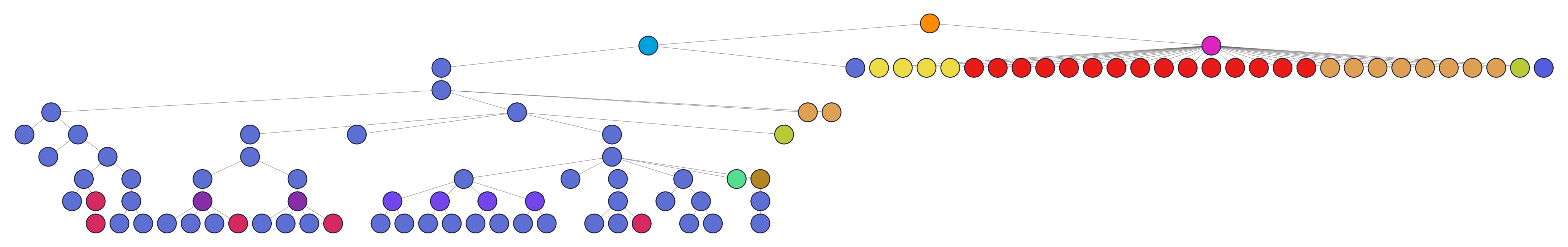}

    \par\vspace{0.25cm}

    {\small\bfseries Samples\par}
    \vspace{0.15cm}

    \begin{minipage}[t]{0.48\textwidth}
        \centering
        \includegraphics[width=\linewidth,keepaspectratio]
        {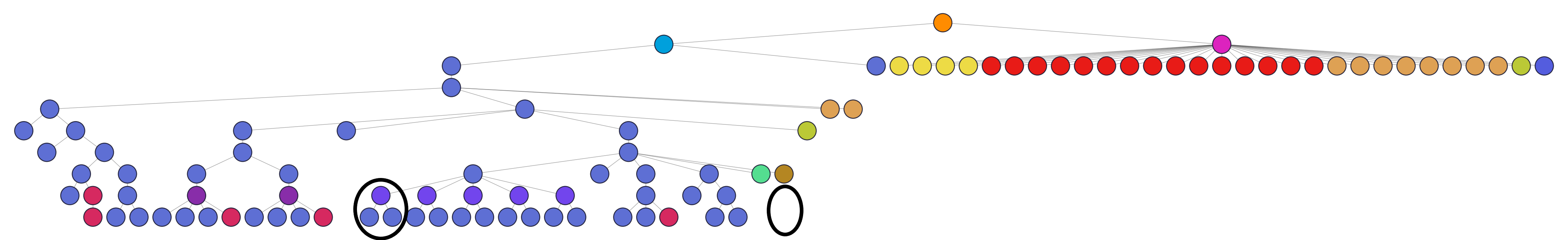}
    \end{minipage}
    \hfill
    \begin{minipage}[t]{0.48\textwidth}
        \centering
        \includegraphics[width=\linewidth,keepaspectratio]
        {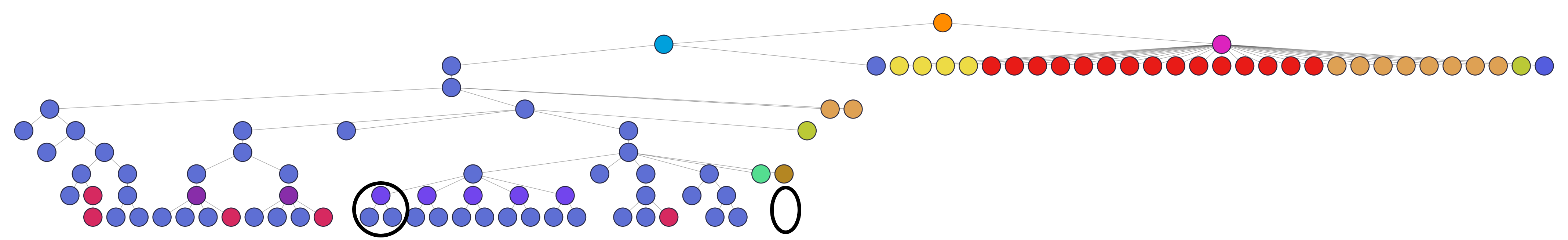}
    \end{minipage}

    \par\vspace{0.35cm}
    \hrule
    \includegraphics[width=0.9\linewidth]{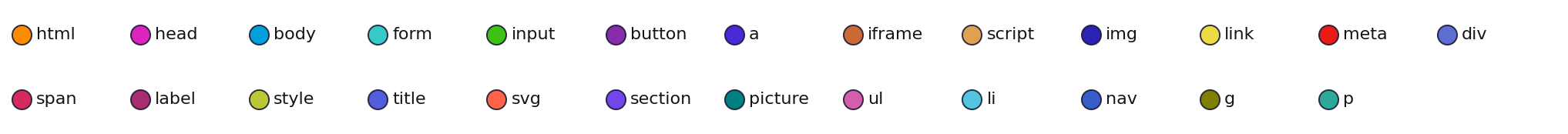}
    \caption{Qualitative analysis: medoids of N compacted clusters and two random samples from each cluster, using structural based features. Changes are marked with a black circle.}
    \label{fig:structural-features}
\end{figure*}

\begin{figure*}[!ht]
    \centering

    {\small\bfseries SC1 ($\mathcal{LJD}(C) = 0.0003$)\par}
    \vspace{0.20cm}

    \begin{minipage}[t]{0.25\textwidth}
        \centering
        {\small\bfseries Medoid\par}
        \vspace{0.15cm}
        \includegraphics[
            width=\linewidth,
            height=0.18\textheight,
            keepaspectratio
        ]{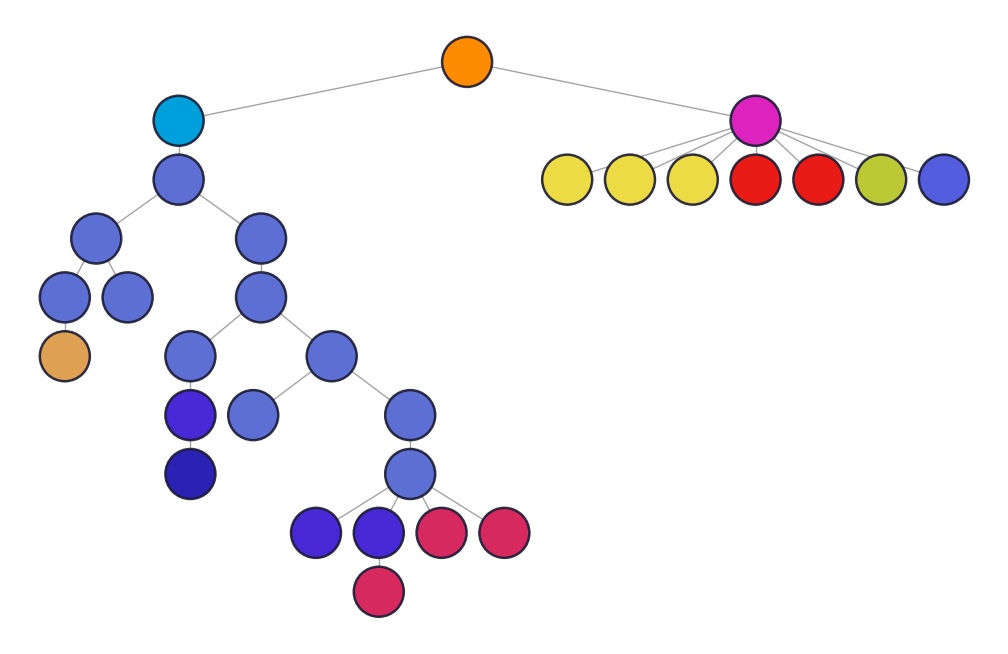}
    \end{minipage}
    \hfill
    \begin{minipage}[t]{0.25\textwidth}
        \centering
        {\small\bfseries Sample 1\par}
        \vspace{0.15cm}
        \includegraphics[
            width=\linewidth,
            height=0.18\textheight,
            keepaspectratio
        ]{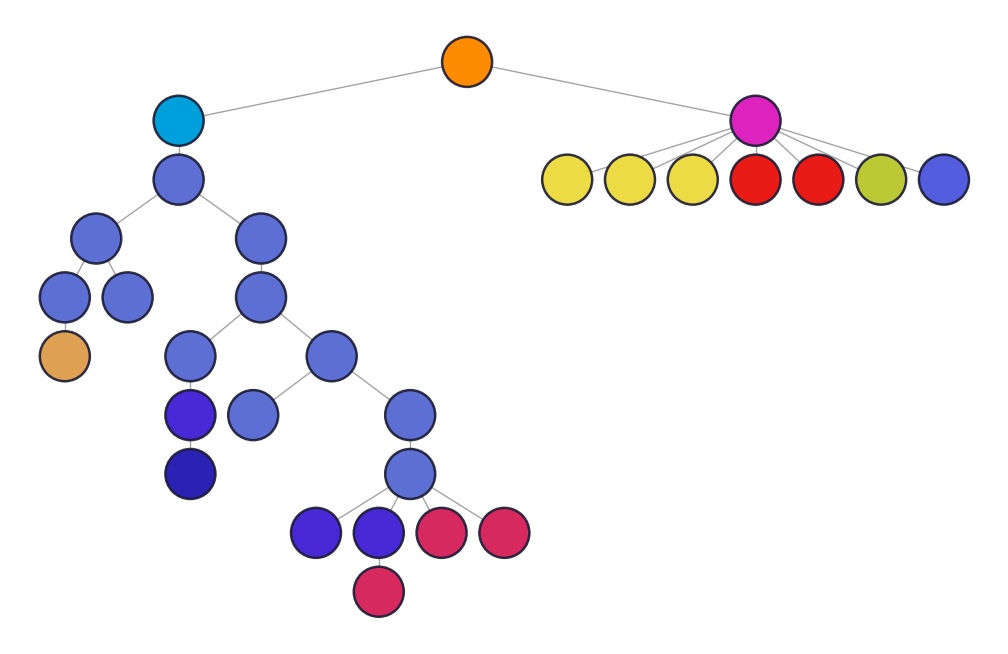}
    \end{minipage}
    \hfill
    \begin{minipage}[t]{0.25\textwidth}
        \centering
        {\small\bfseries Sample 2\par}
        \vspace{0.15cm}
        \includegraphics[
            width=\linewidth,
            height=0.18\textheight,
            keepaspectratio
        ]{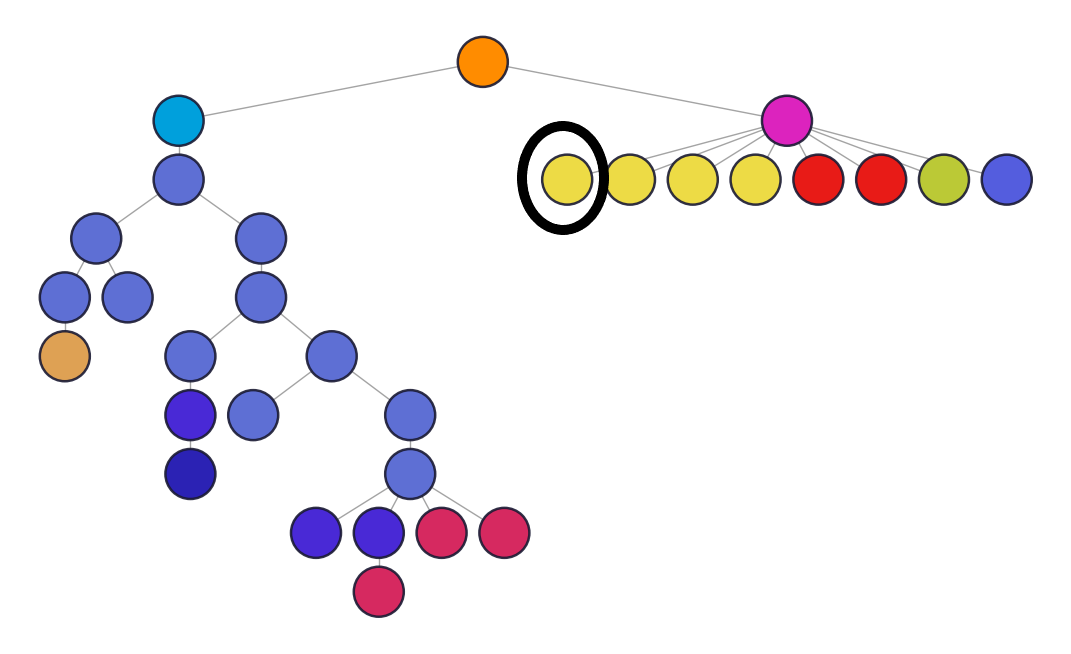}
    \end{minipage}

    \par\vspace{0.30cm}
    \hrule
    \vspace{0.30cm}

    {\small\bfseries SC2 ($\mathcal{LJD}(C) = 0.0053$)\par}
    \vspace{0.15cm}

    {\small\bfseries Medoid\par}
    \vspace{0.15cm}

    \includegraphics[width=\textwidth,keepaspectratio]
    {figs/output_png/cluster_DBSCAN_compacto_101_lvl8_full_14232u_medoid.png}

    \par\vspace{0.25cm}

    {\small\bfseries Samples\par}
    \vspace{0.15cm}

    \begin{minipage}[t]{\textwidth}
        \centering
        \includegraphics[width=\linewidth,keepaspectratio]
        {figs/output_png/cluster_DBSCAN_compacto_101_lvl8_full_19588u.png}
    \end{minipage}
    \hfill
    \begin{minipage}[t]{\textwidth}
        \centering
        \includegraphics[width=\linewidth,keepaspectratio]
        {figs/output_png/cluster_DBSCAN_compacto_101_lvl8_full_28611u.png}
    \end{minipage}

    \par\vspace{0.30cm}
    \hrule
    \vspace{0.30cm}

    {\small\bfseries SC3 ($\mathcal{LJD}(C) = 0.0058$)\par}
    \vspace{0.15cm}

    {\small\bfseries Medoid\par}
    \vspace{0.15cm}

    \includegraphics[width=0.90\textwidth,keepaspectratio]
    {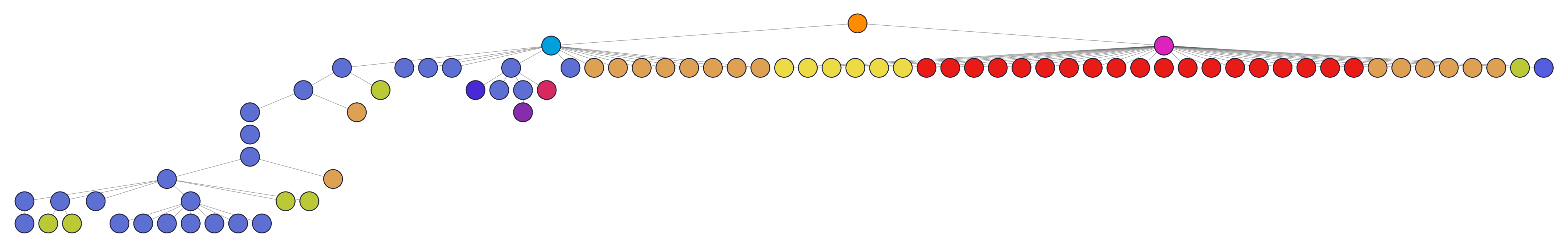}

    \par\vspace{0.25cm}

    {\small\bfseries Samples\par}
    \vspace{0.15cm}

    \begin{minipage}[t]{0.48\textwidth}
        \centering
        \includegraphics[width=\linewidth,keepaspectratio]
        {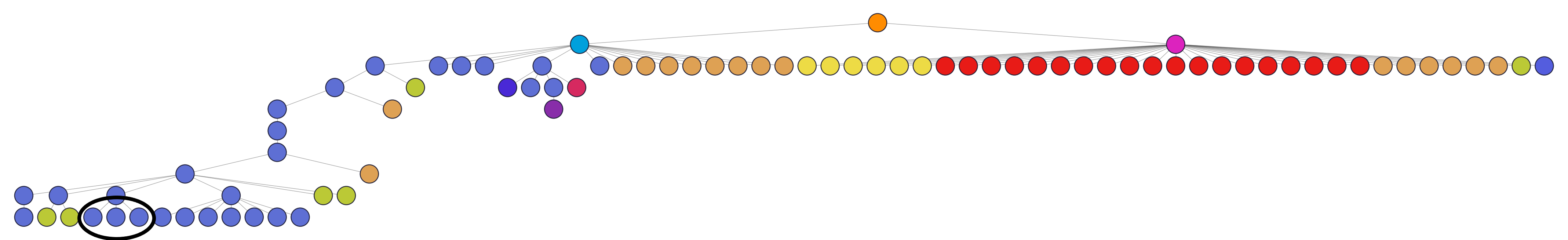}
    \end{minipage}
    \hfill
    \begin{minipage}[t]{0.48\textwidth}
        \centering
        \includegraphics[width=\linewidth,keepaspectratio]
        {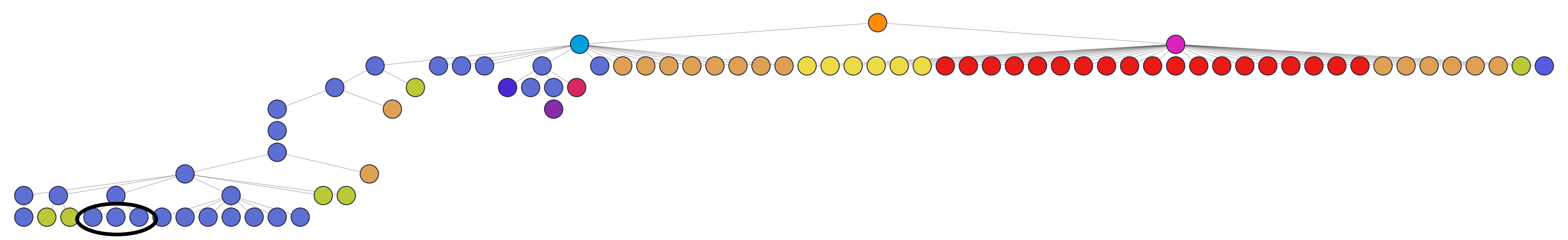}
    \end{minipage}

    \par\vspace{0.35cm}
 \hrule
    \includegraphics[width=0.9\linewidth]{legend.png}
    \caption{Qualitative analysis: medoids of N compacted clusters and two random samples from each cluster, using structural and content-based features. Changes are marked with a black circle.}
    \label{fig:structural-content_features}
\end{figure*}

Finally, to provide a more comprehensive view of the generated clusters, a qualitative analysis of the medoids extracted from the most dispersed clusters was conducted and is reported in Appendix \ref{app:qualitative}.

\subsection{Discussion}

The results presented in this work show that DOM-tree representations provide a useful basis for identifying structural similarities among phishing webpages. Across the evaluated clustering algorithms, the grid-search analysis indicates that meaningful groupings can be obtained from the extracted tree features, although each algorithm reacts differently to the type of information used. When only structural properties are considered, the best configurations tend to appear at deeper DOM levels, suggesting that global tree complexity and nested organization provide relevant information for distinguishing phishing templates. In contrast, when content-based features are incorporated, the best configurations for density-based methods are generally obtained at shallower levels. This indicates that tag-level information introduces a more specific but also more variable description of the webpages, making deeper levels more sensitive to local implementation differences.

The comparison between structural-only and structural-content representations highlights a central trade-off. Structural features provide a general characterization of the DOM, capturing aspects such as size, connectivity, and global topology. As a result, they are useful for detecting broad structural reuse patterns, but they may group together webpages that share a similar shape while differing in the specific HTML elements used. Conversely, adding content-based features substantially reduces $\mathcal{LJD}_{mean}$ for all algorithms, indicating more compact clusters according to the level-wise distribution of HTML tags. This suggests that content information refines the notion of similarity and helps identify more specific template variants. However, this refinement also increases fragmentation: density-based methods produce a higher proportion of noisy samples, while AHC generates many singleton clusters. Therefore, content-based features improve specificity but reduce coverage, whereas structural-only features provide broader but less fine-grained groupings.

The qualitative validation further supports this interpretation. The selected compact clusters correspond to visually coherent groups of DOM trees, which is consistent with the idea that phishing kits, reusable templates, and shared web components can generate multiple webpages with highly similar internal structures. In these cases, even if pages differ in domains, visual details, or deployment context, their underlying DOM organization may preserve evidence of template reuse or common construction logic. Conversely, the dispersed clusters contain heterogeneous structures, including simplified or almost flattened DOMs. These cases suggest that not all phishing webpages expose the same level of structural regularity: some may rely on minimal HTML, redirection mechanisms, dynamically generated content, or incomplete/obfuscated structures, making them more difficult to compare through tree-based representations.

\subsection{Limitation \& Future Work} Despite the promising results achieved by the proposed approach, several (potential) issues need to be highlighted. First, although the use of content-based features yielded the best overall performance according to the introduced metric, clustering algorithms based on these features were characterized by a high number of unclustered elements (DBSCAN, OPTICS) and singleton groups (AHC). In this regard, strategies for incorporating these instances into the analysis, such as nearest-cluster assignment or two-stage clustering, should be further investigated; otherwise, a significant portion of the dataset remains unanalyzed. Furthermore, although the proposed metric proved useful for assessing cluster quality, it is important to note that it relies exclusively on tag-based and level information. As these same features are also employed during clustering, the evaluation may be partially biased toward clustering solutions generated from similar information. Consequently, approaches based on content features may naturally yield more compact clusters according to this metric. To mitigate this potential bias, alternative and more feature-agnostic distance measures, such as graph edit distance, should be explored.

\subsection{Ethics Concerns}
From an ethical research perspective, three key considerations must be considered. Firstly, although the dataset primarily consists of publicly accessible phishing webpages collected for security research purposes, it may still contain sensitive information, including personally identifiable information (PII) such as email addresses, names, telephone numbers, and other identifiers inadvertently exposed within page content. 

Secondly, the proposed methodology raises dual-use concerns. While the approach is intended to support cybersecurity objectives by characterising phishing infrastructure and improving detection and defensive mechanisms, the inferred structural and behavioural insights could potentially be exploited by malicious actors to adapt their tactics, techniques, and procedures (TTPs), thereby increasing the sophistication and evasiveness of phishing campaigns. 

Thirdly, although clustering and structural similarity analysis may suggest commonalities between phishing webpages, such inferences should not be interpreted as definitive evidence of shared authorship or attacker identity. Any attribution-related conclusions must be treated as probabilistic and used exclusively for defensive and threat intelligence purposes.

\section{Conclusion}
\label{sec:conclusion}

This work investigated whether the HTML structure of phishing webpages can be used as a reliable fingerprint for identifying structural similarities and potential template reuse. By representing webpages as DOM trees and clustering them using structural and content-based features, the results show that recurrent patterns can be detected without relying on prior labels. Structural features provide a broader view of common DOM organization, while the inclusion of tag-level content information enables a more fine-grained identification of highly homogeneous template variants. The proposed $\mathcal{LJD}$ metric and the qualitative validation further confirm that compact clusters correspond to visually coherent tree structures. Overall, the findings support the use of DOM-based clustering as a promising approach for phishing template analysis, while also highlighting the need to better handle noisy, flattened, or highly fragmented samples in future work.


\section*{Acknowledgments}
This work has been partially supported by the European Union through the Horizon Europe Programme under the projects SAFEHORIZON (Grant Agreement No. 101168562), ENSEMBLE (Grant Agreement No. 101168360), and by Vicomtech under the project ATHENA-CV (Grant Number E22503/2024). The content of this article does not reflect the official opinion of the European Union. Responsibility for the information and views expressed therein lies entirely with the authors.


\bibliographystyle{IEEEtran}
\bibliography{references}

\appendices

\section{Grid-search results (part II)}\label{sub:app1}

As mentioned, while Section \ref{subsec:gridsearch} reports grid-search results obtained using clustering algorithms based solely on structural information, this section presents the results when structural information is combined with content-based features. In these figures (\ref{fig:dbscan_gridsearch_v2},\ref{fig:optics_gridsearch_v2},\ref{fig:ahcdhc_gridsearch_v2}), the best-performing configurations are highlighted with a red box. These configurations are then used in Section \ref{subsec:compare}.

\begin{figure}[!htbp]
    \centering
        \includegraphics[width=0.9\linewidth]{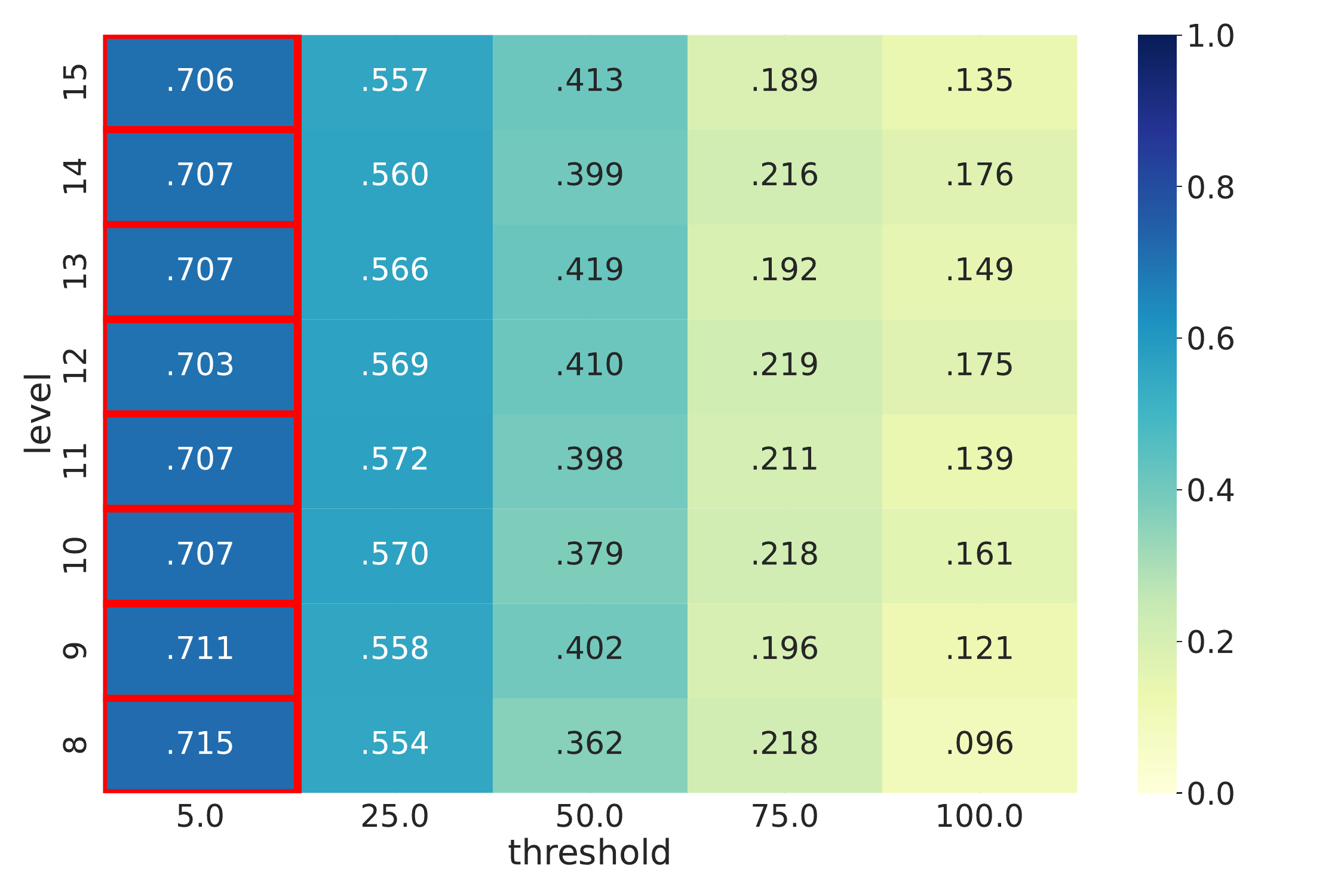}
    \caption{AHC $\mathcal{S}_{score}$ during the grid-search using structural and content-based features.}
    \label{fig:ahcdhc_gridsearch_v2}
\end{figure}

\begin{figure*}[!htbp]
    \centering
    \begin{subfigure}[b]{0.24\textwidth}
        \includegraphics[width=\textwidth]{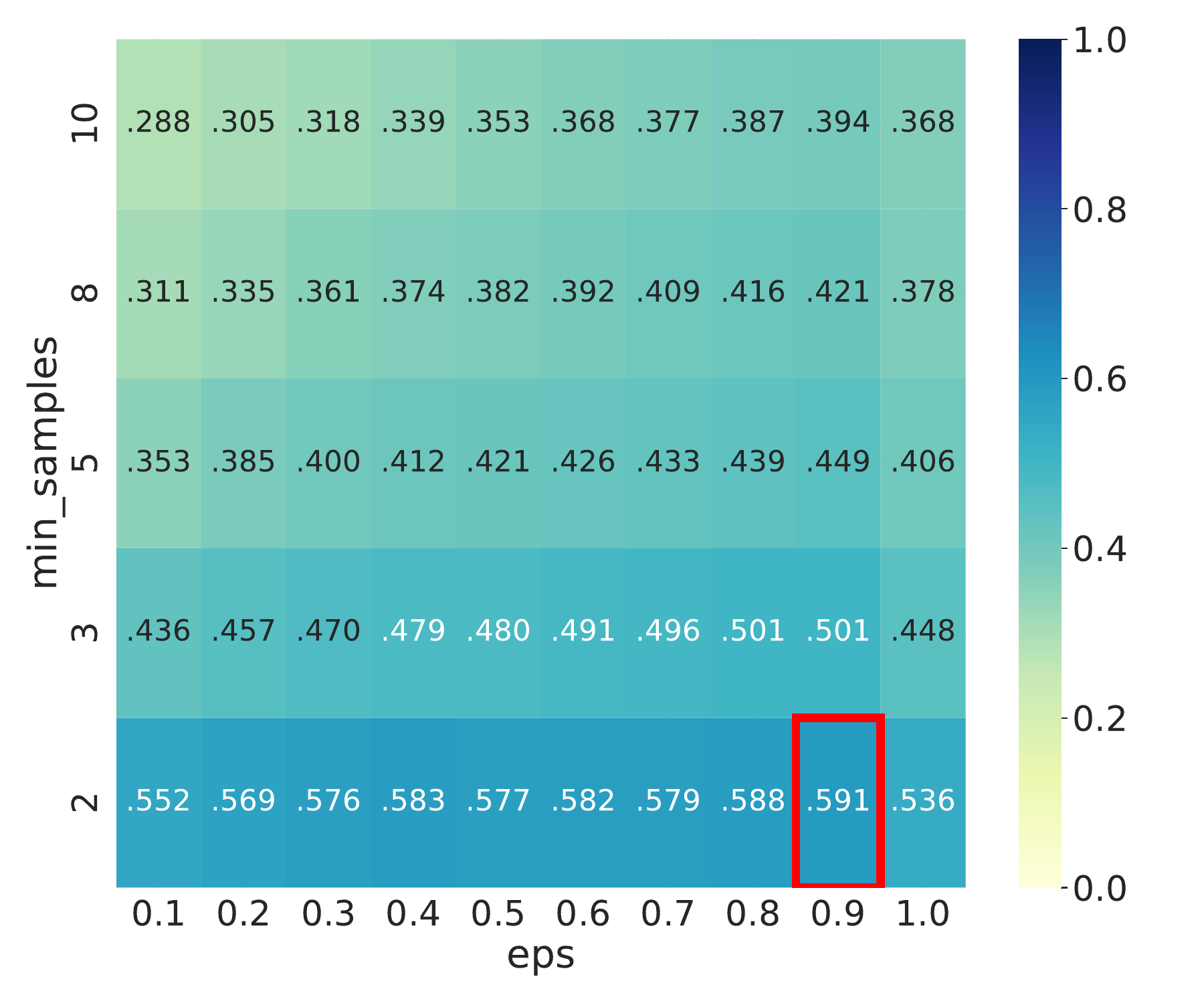}
        \caption{Level 8.}
        \label{fig:d_lvl8_bis}
    \end{subfigure}
    \begin{subfigure}[b]{0.24\textwidth}
        \centering
        \includegraphics[width=\textwidth]{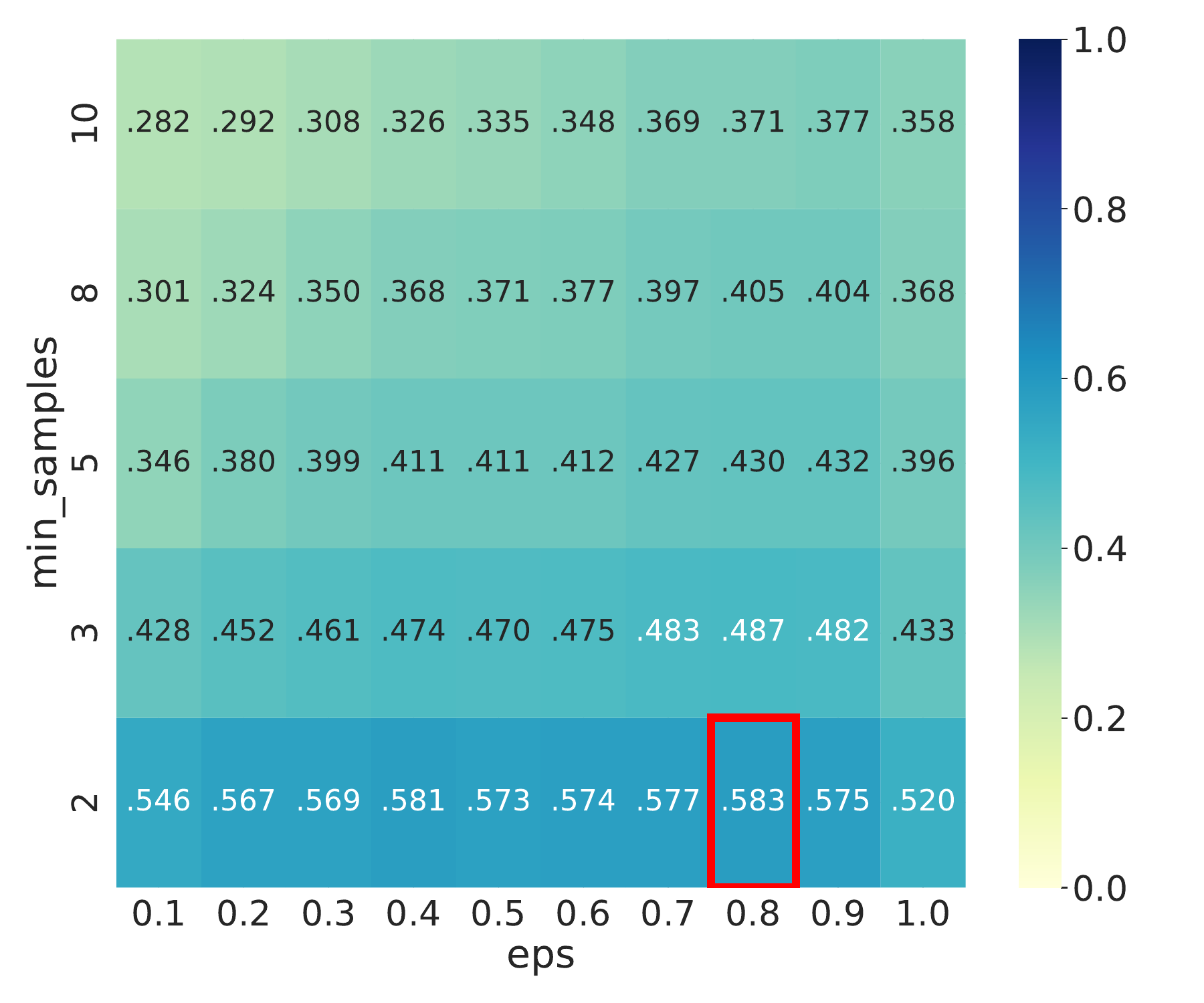}
        \caption{Level 9.}
        \label{fig:d_lvl9_bis}
    \end{subfigure}
    \begin{subfigure}[b]{0.24\textwidth}
        \centering
        \includegraphics[width=\textwidth]{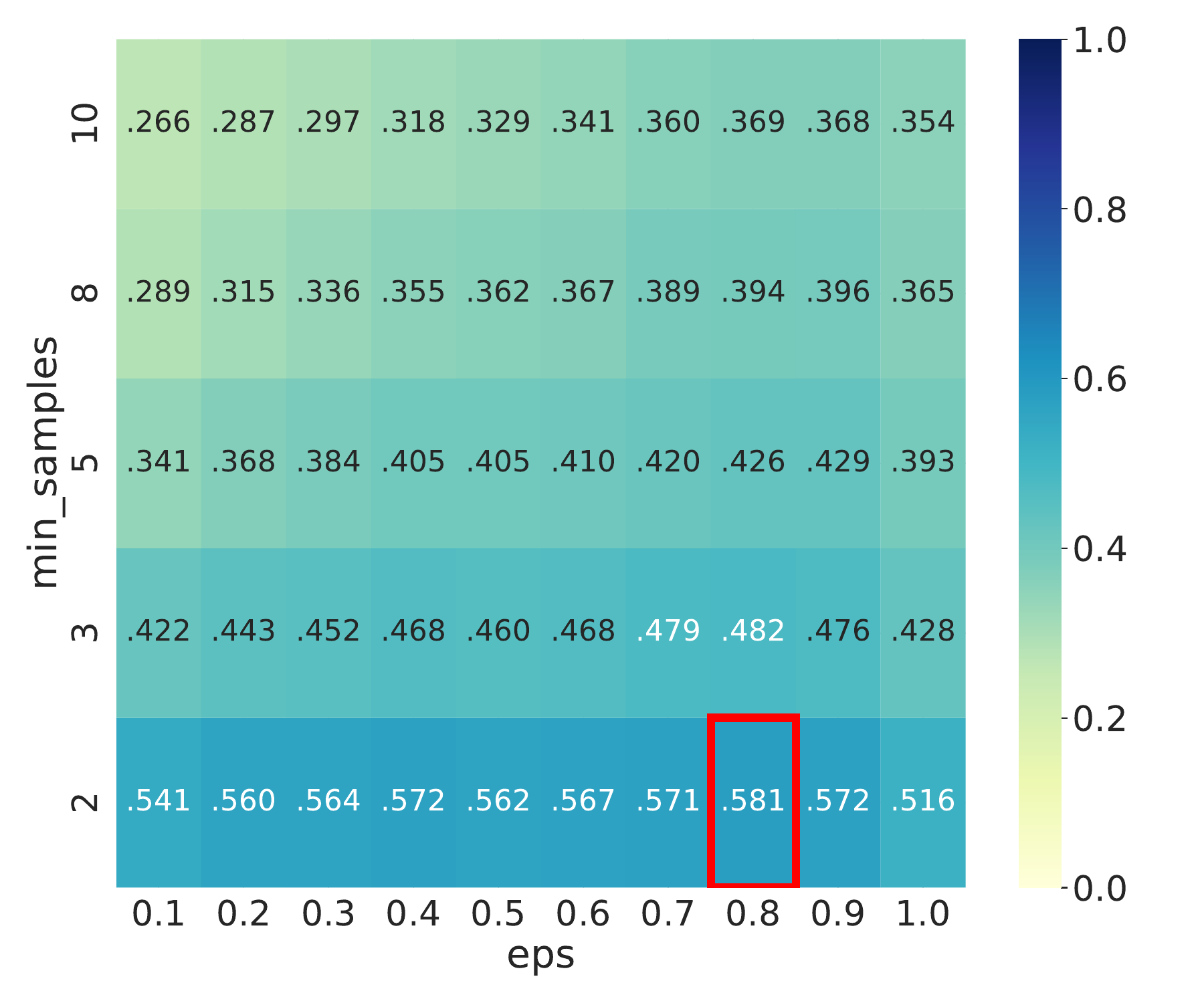}
        \caption{Level 10.}
        \label{fig:d_lvl10_bis}
    \end{subfigure}
    \begin{subfigure}[b]{0.24\textwidth}
        \centering
        \includegraphics[width=\textwidth]{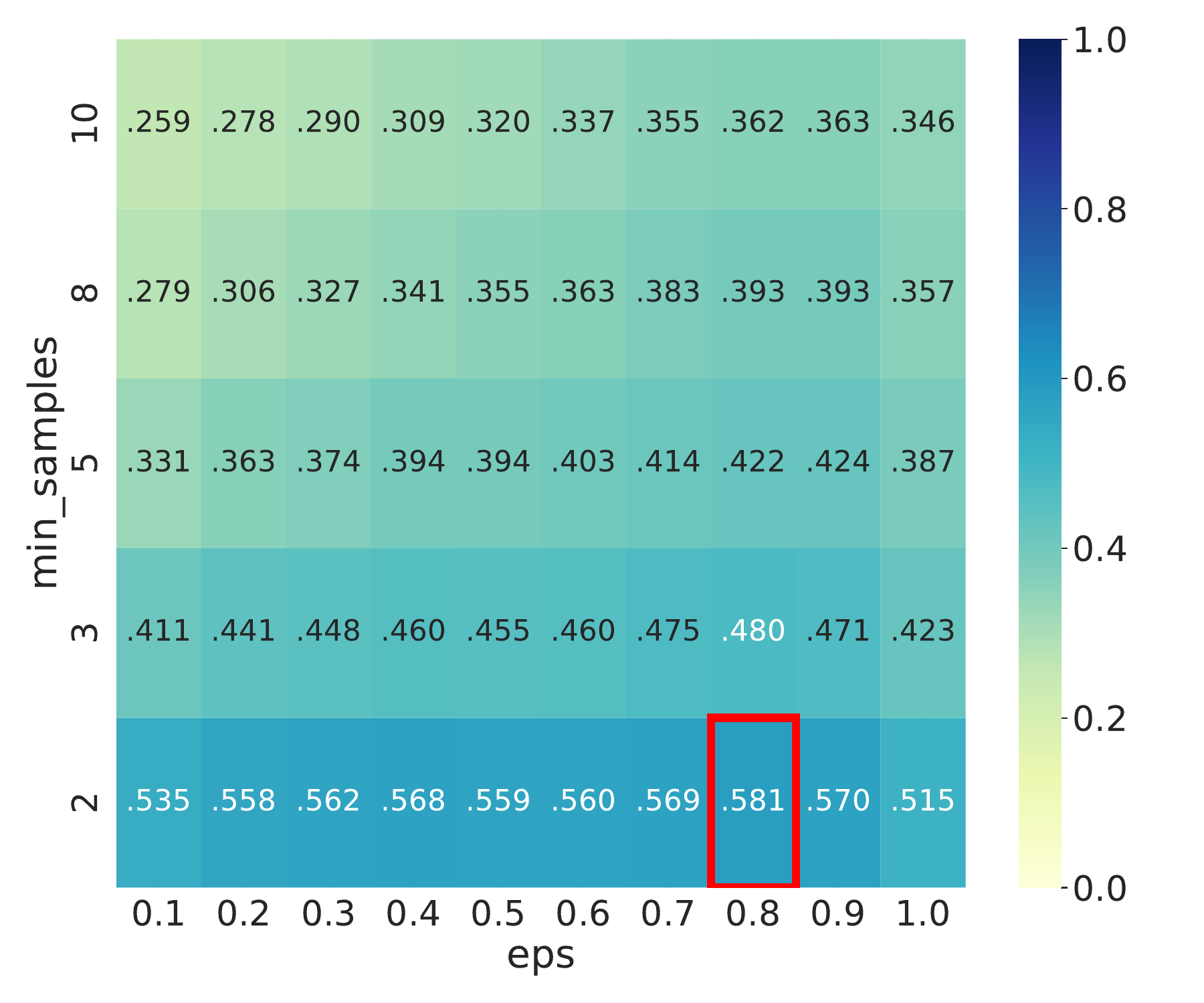}
        \caption{Level 11.}
        \label{fig:d_lvl11_bis}
    \end{subfigure}
    \begin{subfigure}[b]{0.24\textwidth}
        \includegraphics[width=\textwidth]{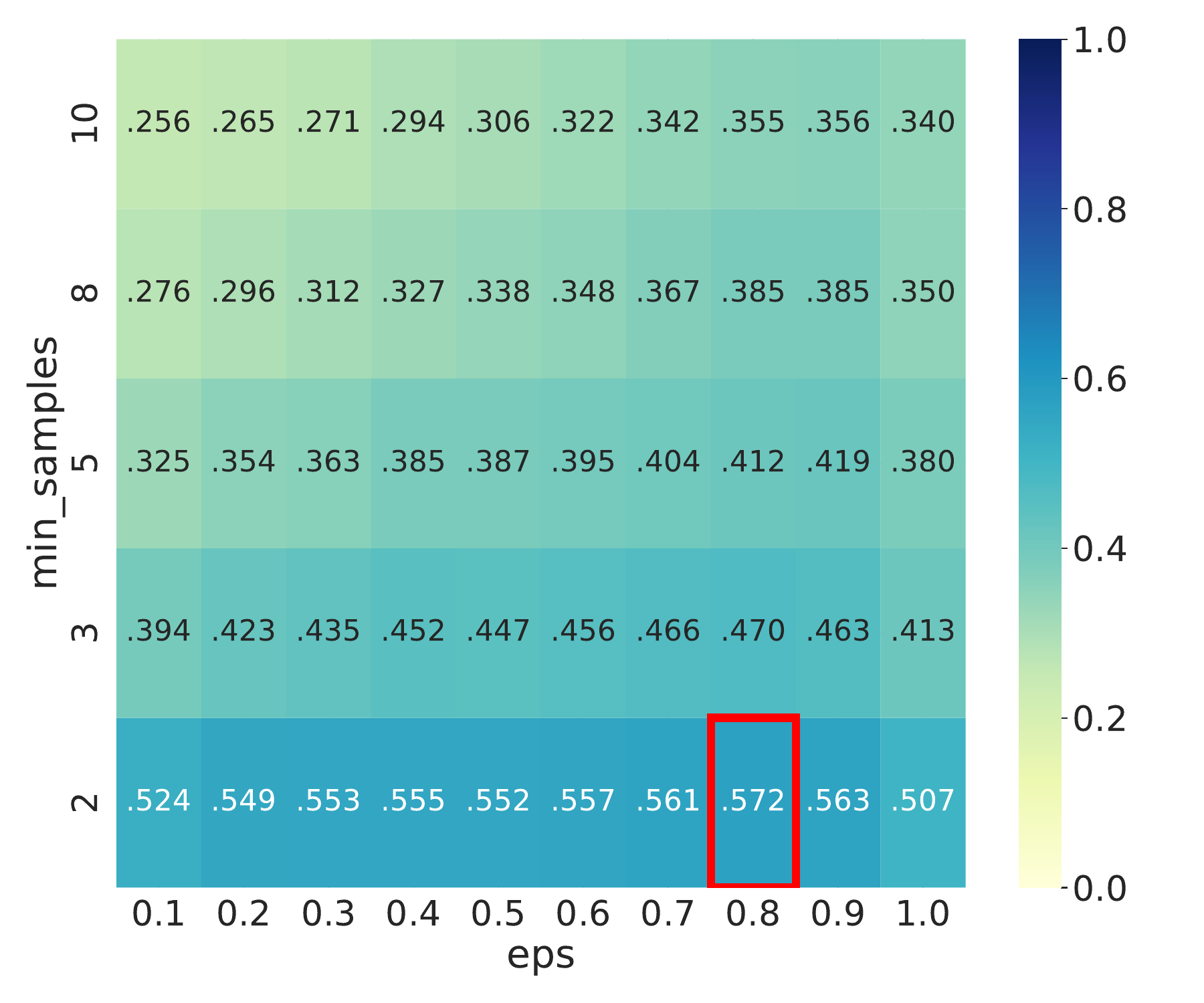}
        \caption{Level 12.}
        \label{fig:d_lvl12_bis}
    \end{subfigure}
    \begin{subfigure}[b]{0.24\textwidth}
        \centering
        \includegraphics[width=\textwidth]{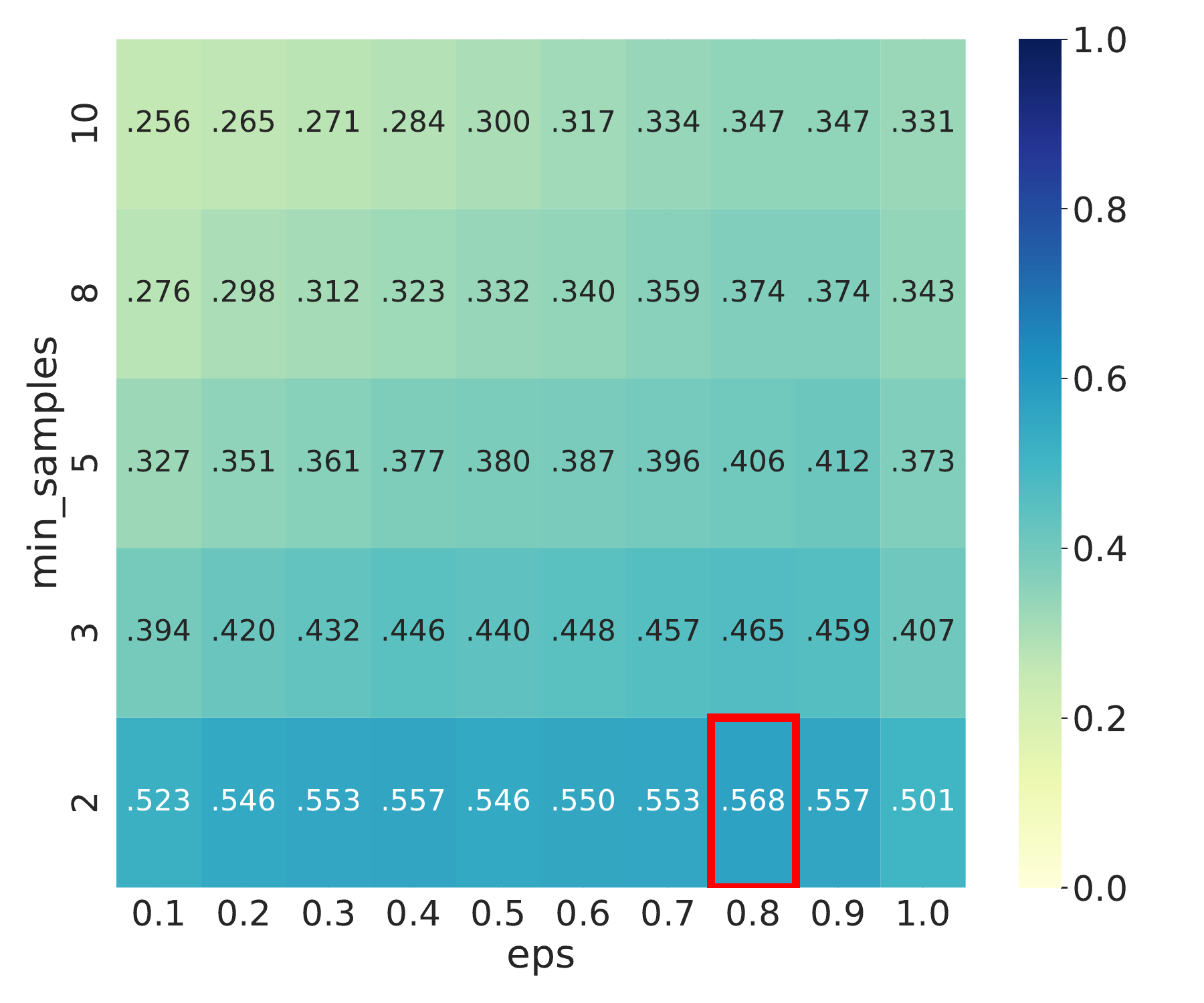}
        \caption{Level 13.}
        \label{fig:d_lvl13_bis}
    \end{subfigure}
    \begin{subfigure}[b]{0.24\textwidth}
        \centering
        \includegraphics[width=\textwidth]{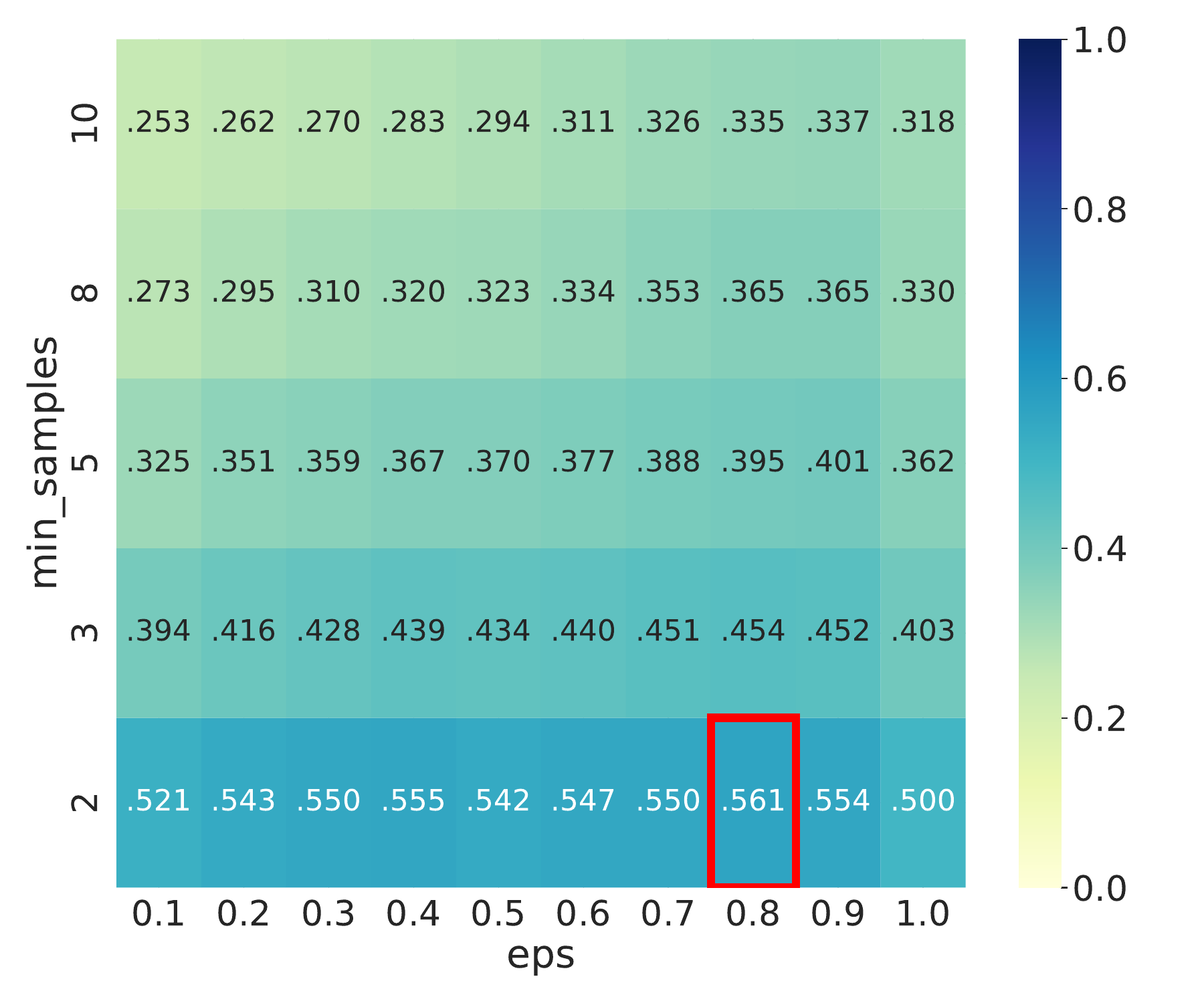}
        \caption{Level 14.}
        \label{fig:d_lvl14_bis}
    \end{subfigure}
    \begin{subfigure}[b]{0.24\textwidth}
        \centering
        \includegraphics[width=\textwidth]{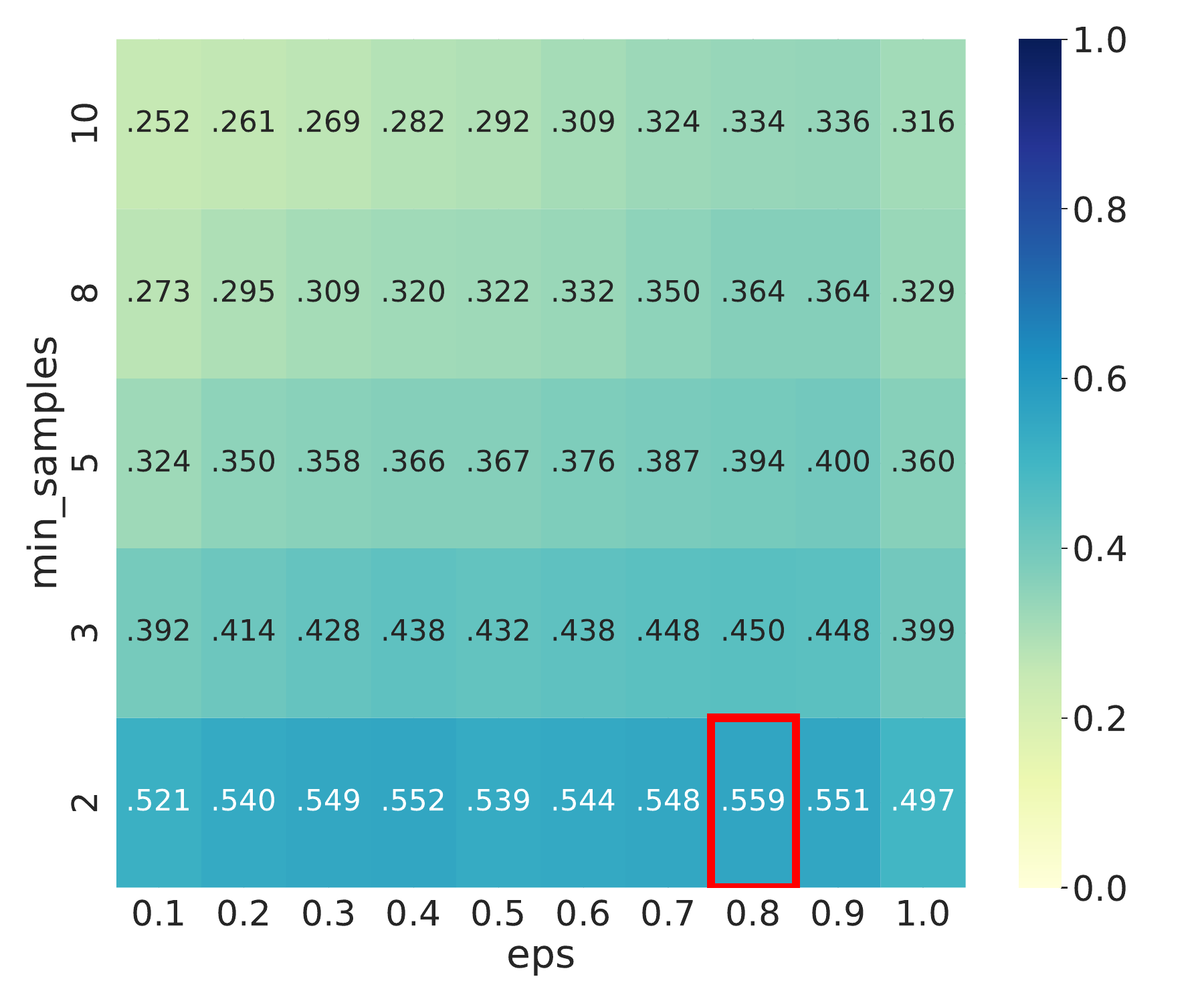}
        \caption{Level 15.}
        \label{fig:d_lvl15_bis}
    \end{subfigure}
    \centering
    \caption{DBSCAN $\mathcal{S}_{score}$ during the grid-search using structural and content-based features.}
    \label{fig:dbscan_gridsearch_v2}
\end{figure*}

\begin{figure*}[!htbp]
    \centering
    \begin{subfigure}[b]{0.24\textwidth}
        \includegraphics[width=\textwidth]{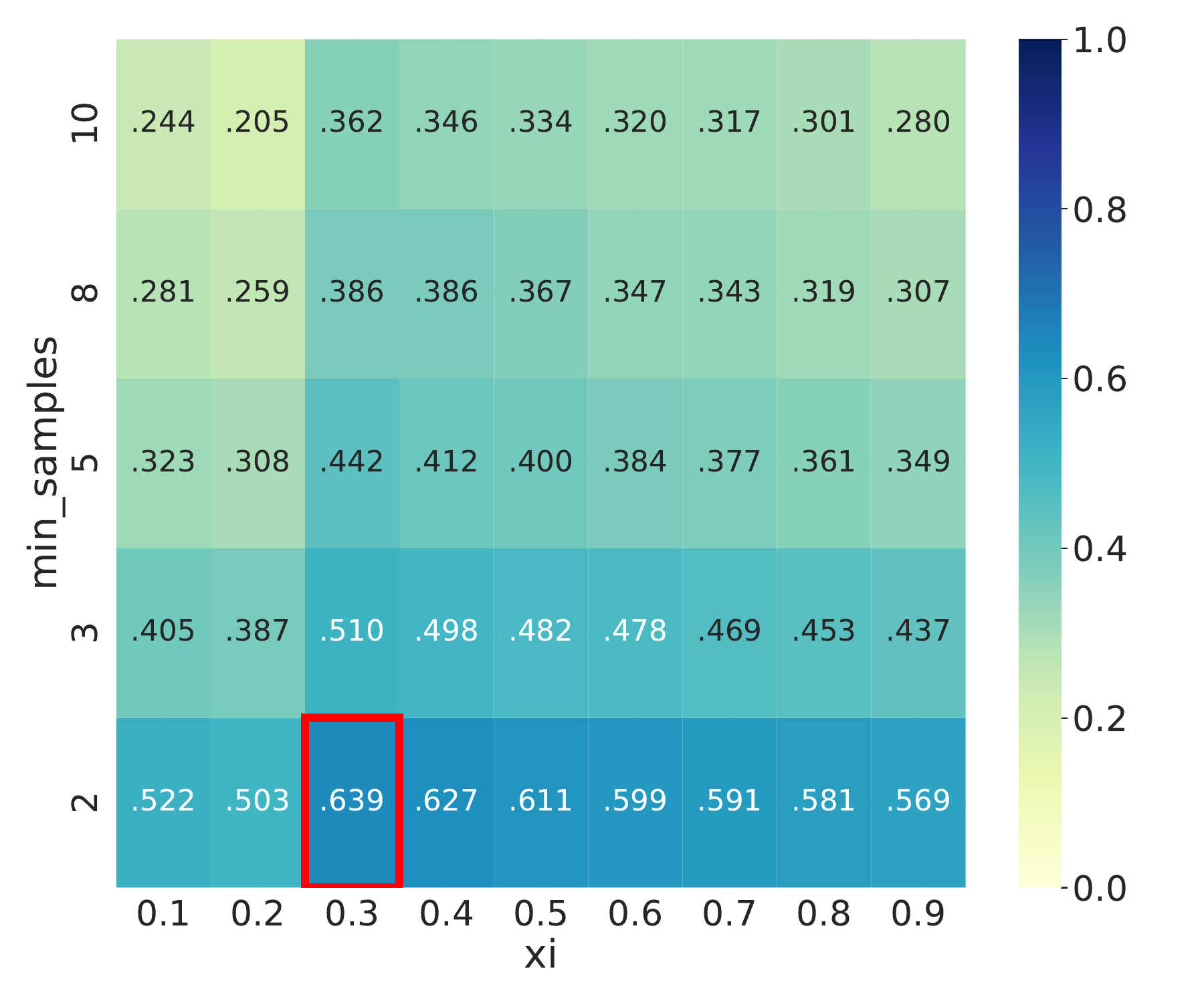}
        \caption{Level 8.}
        \label{fig:o_lvl8_bis}
    \end{subfigure}
    \begin{subfigure}[b]{0.24\textwidth}
        \centering
        \includegraphics[width=\textwidth]{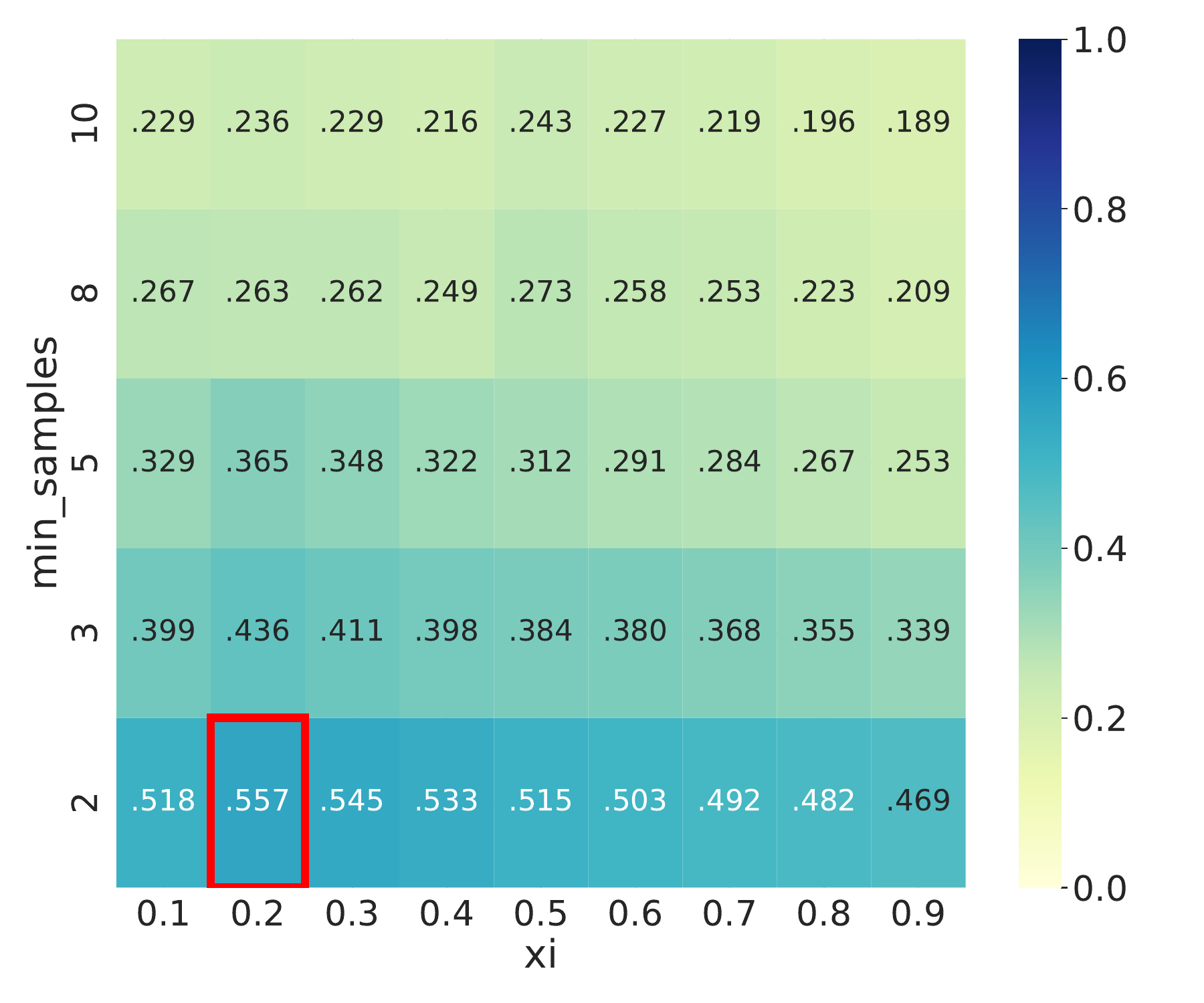}
        \caption{Level 9.}
        \label{fig:o_lvl9_bis}
    \end{subfigure}
    \begin{subfigure}[b]{0.24\textwidth}
        \centering
        \includegraphics[width=\textwidth]{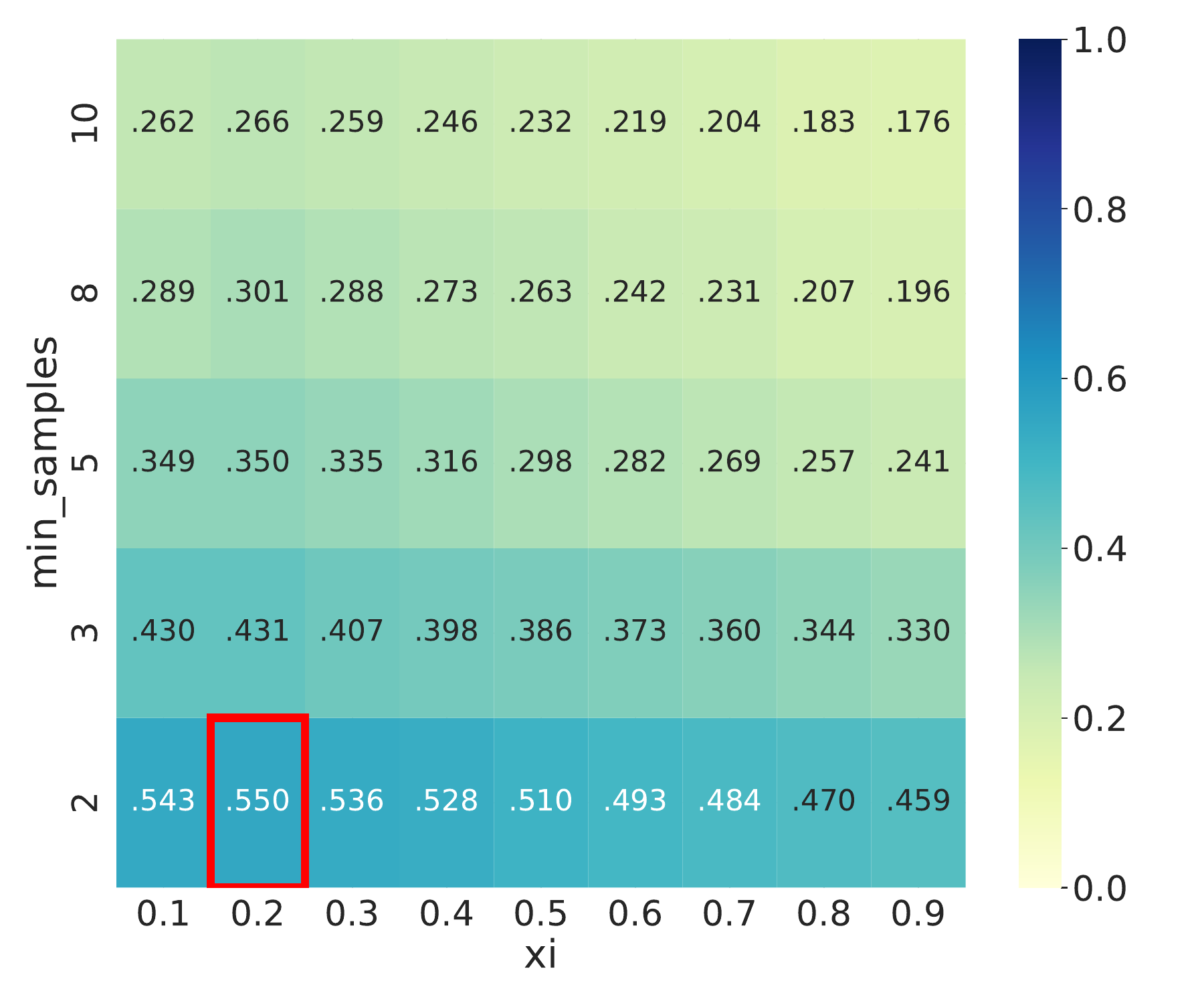}
        \caption{Level 10.}
        \label{fig:o_lvl10_bis}
    \end{subfigure}
    \begin{subfigure}[b]{0.24\textwidth}
        \centering
        \includegraphics[width=\textwidth]{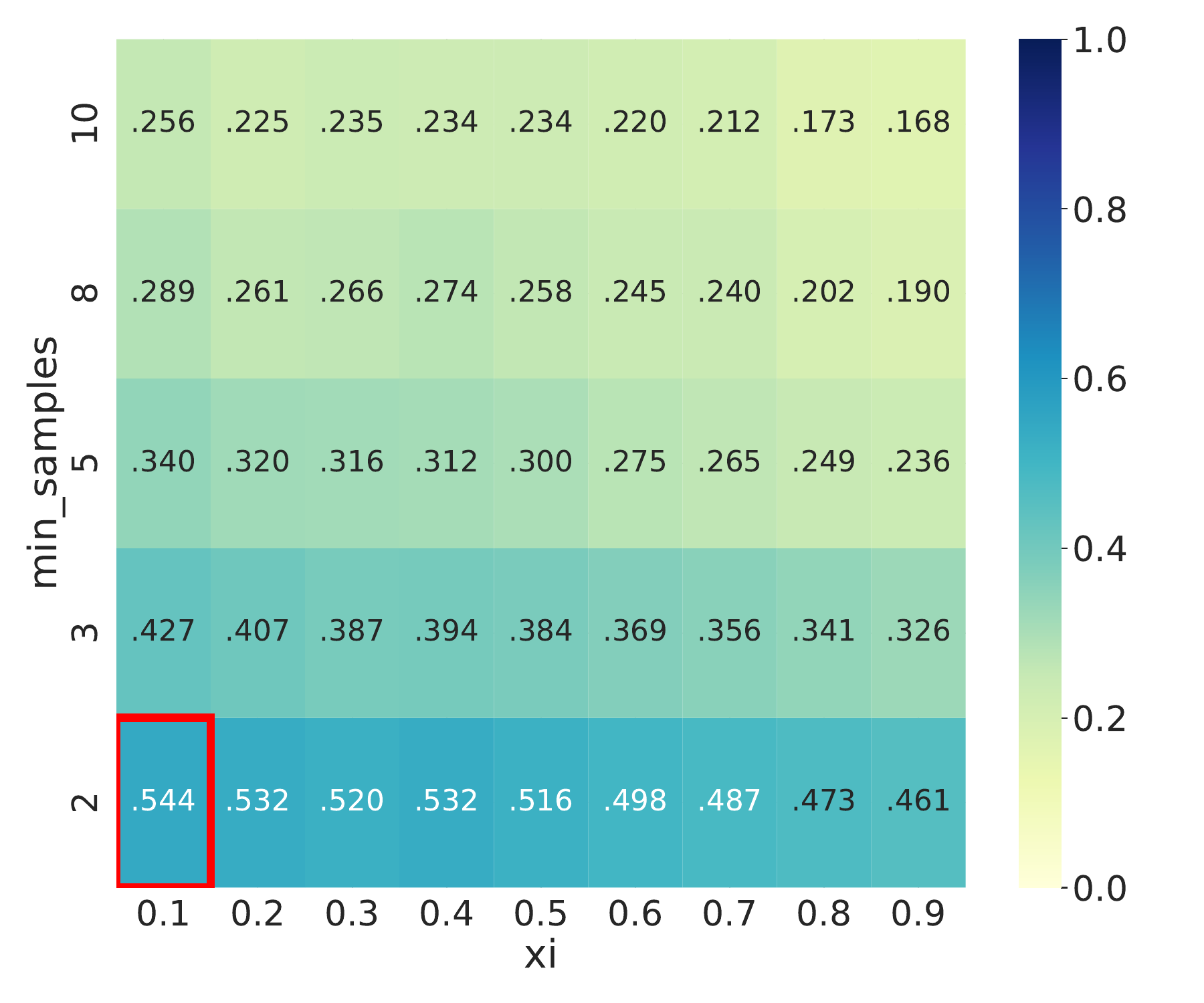}
        \caption{Level 11.}
        \label{fig:o_lvl11_bis}
    \end{subfigure}
    \begin{subfigure}[b]{0.24\textwidth}
        \includegraphics[width=\textwidth]{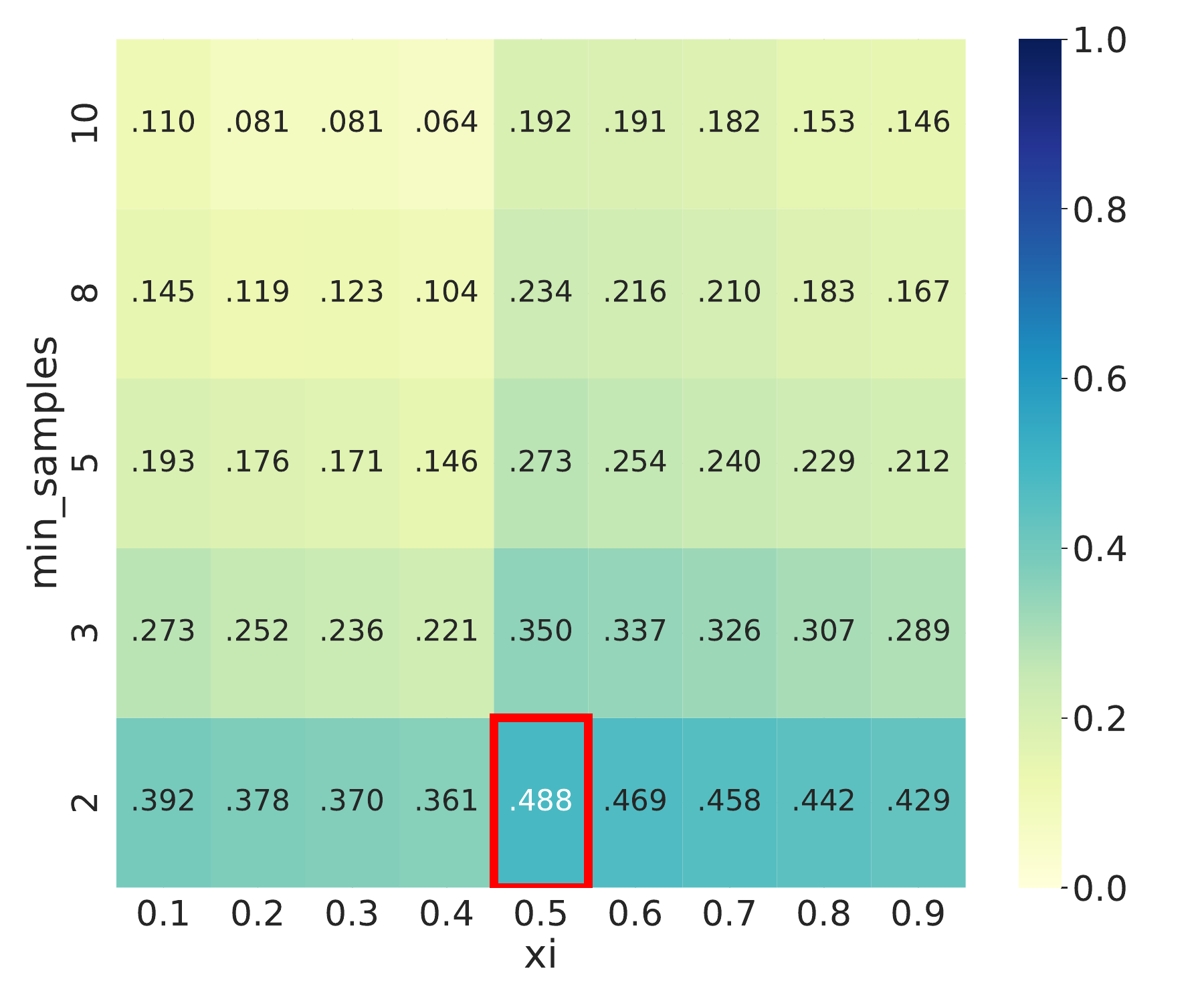}
        \caption{Level 12.}
        \label{fig:o_lvl12_bis}
    \end{subfigure}
    \begin{subfigure}[b]{0.24\textwidth}
        \centering
        \includegraphics[width=\textwidth]{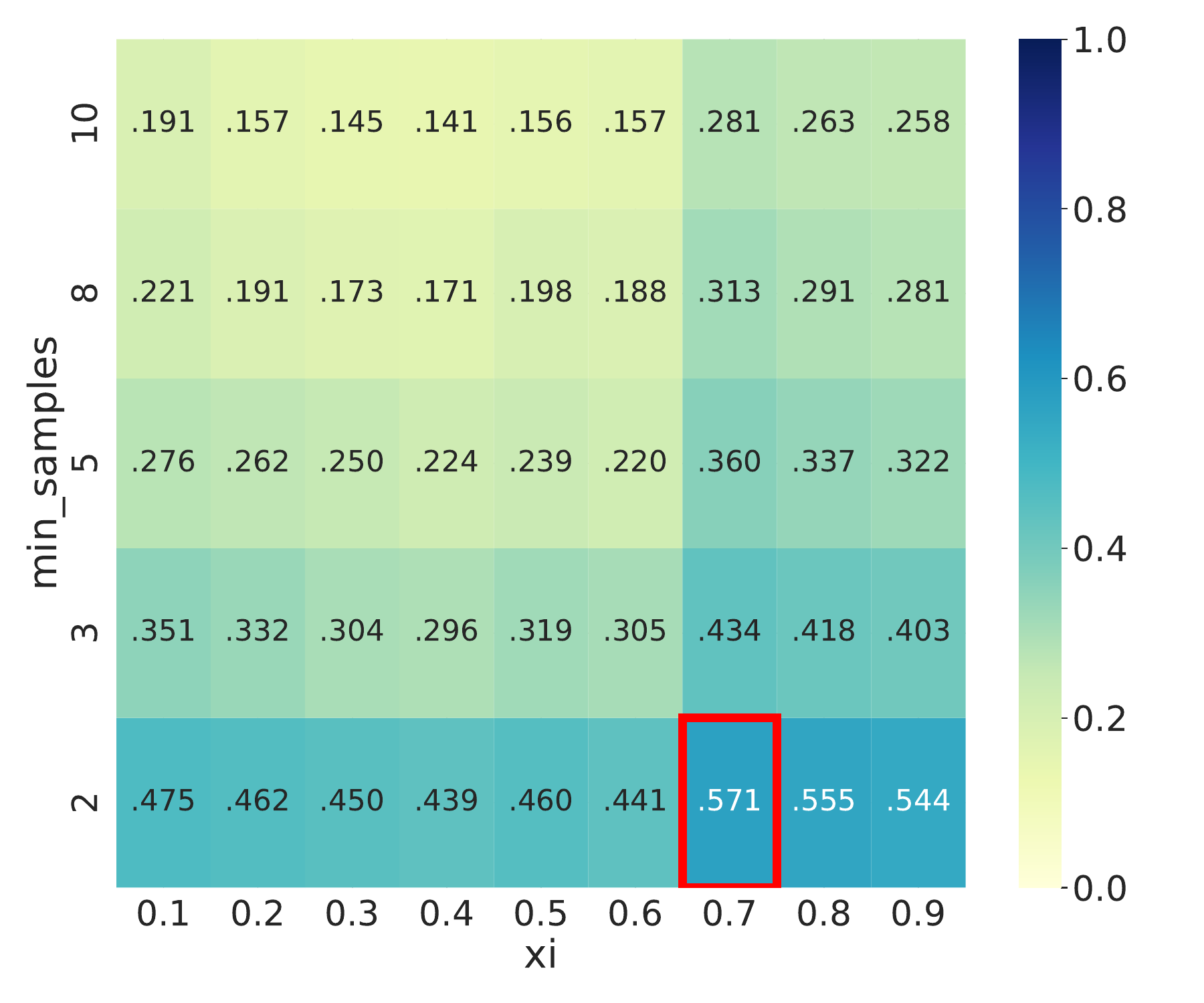}
        \caption{Level 13.}
        \label{fig:o_lvl13_bis}
    \end{subfigure}
    \begin{subfigure}[b]{0.24\textwidth}
        \centering
        \includegraphics[width=\textwidth]{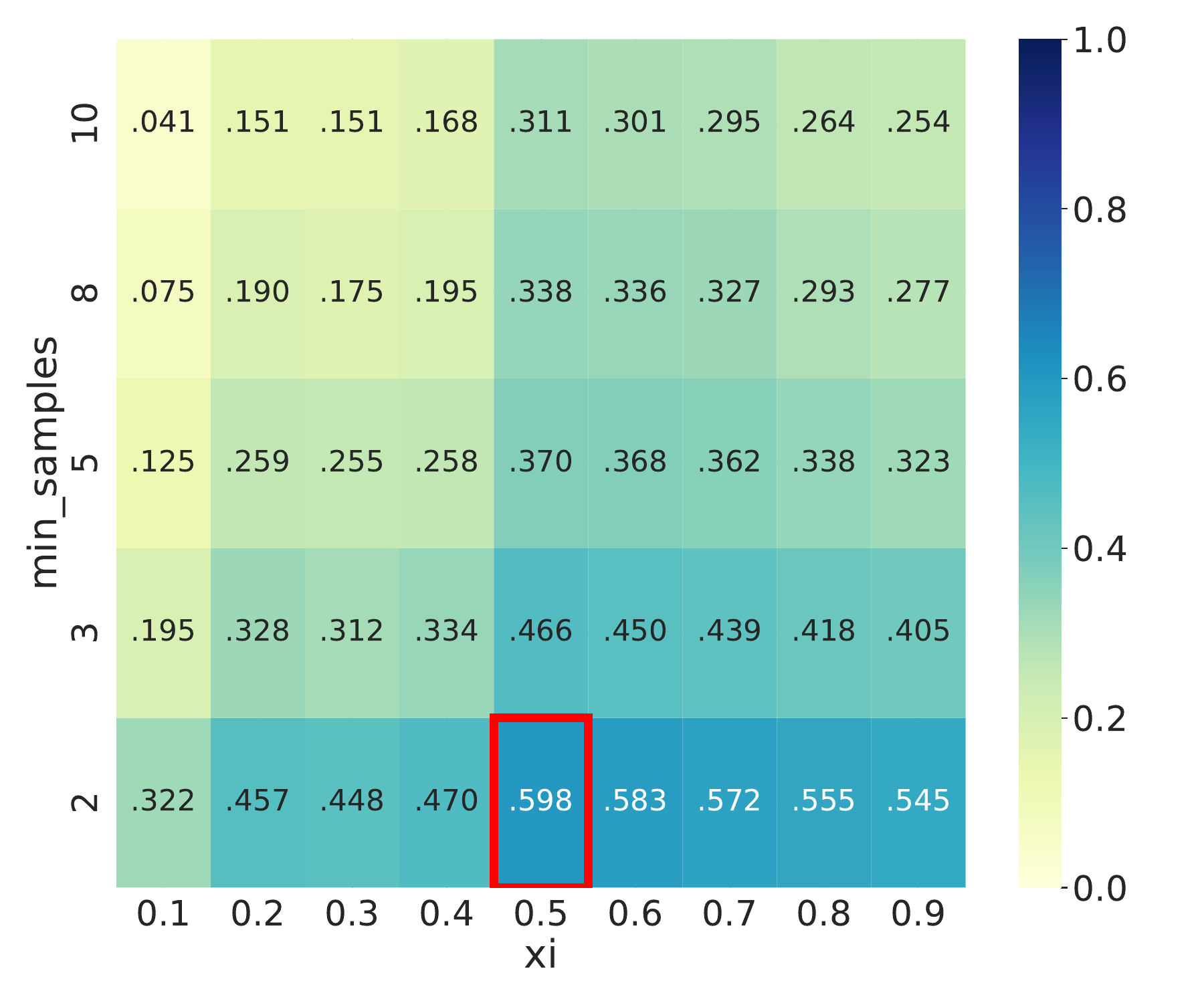}
        \caption{Level 14.}
        \label{fig:o_lvl14_bis}
    \end{subfigure}
    \begin{subfigure}[b]{0.24\textwidth}
        \centering
        \includegraphics[width=\textwidth]{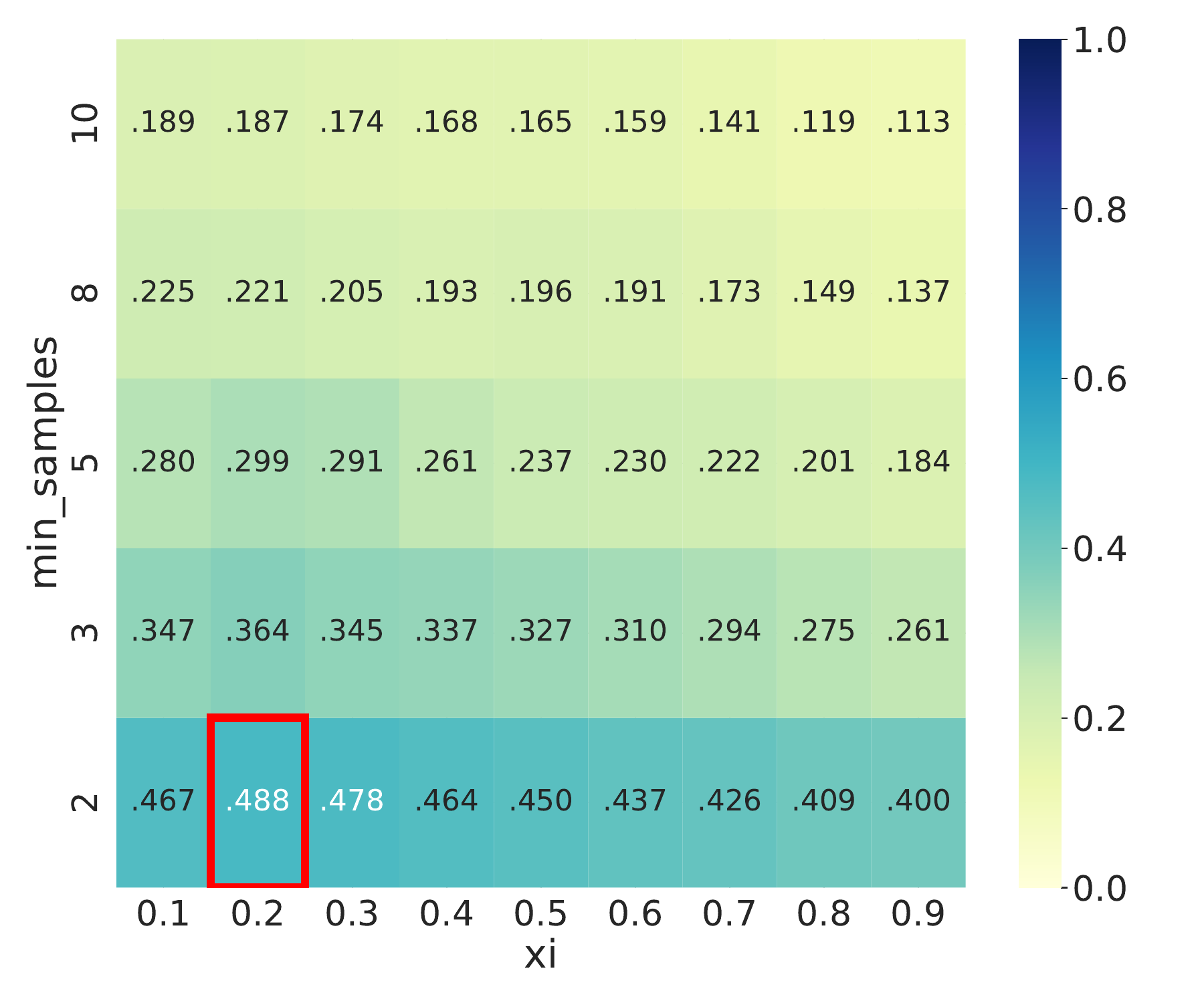}
        \caption{Level 15.}
        \label{fig:o_lvl15_bis}
    \end{subfigure}
    \centering
    \caption{OPTICS $\mathcal{S}_{score}$ during the grid-search using structural and content-based features.}
    \label{fig:optics_gridsearch_v2}
\end{figure*}

\section{Qualitative validation - part II}\label{app:qualitative}

In terms of dispersed clusters, Figure \ref{fig:qualitative_worst} reports the medoids of the clusters with the highest $\mathcal{LJD}(C)$ values. Specifically, Figure \ref{fig:qualitative_worst_1} shows the three selected clusters obtained using only structural features, while Figure \ref{fig:qualitative_worst_2} shows the three selected clusters obtained when structural and content-based features are combined. Compared with the compact clusters reported in Figures \ref{fig:structural-features} and \ref{fig:structural-content_features}, these medoids exhibit more heterogeneous configurations, including trees with reduced depth, simplified structures, or clearly different branching organizations. In some cases, the medoids consist of only three levels; that is, after removing the first two levels, corresponding to \texttt{<html>} and \texttt{<head>}/\texttt{<body>}, the remaining DOM structure is almost completely flattened.
\begin{figure*}[!ht]
    \centering
    \begin{subfigure}{\textwidth}

    \begin{tabular}{ccc}
        \includegraphics[width=0.55\textwidth]{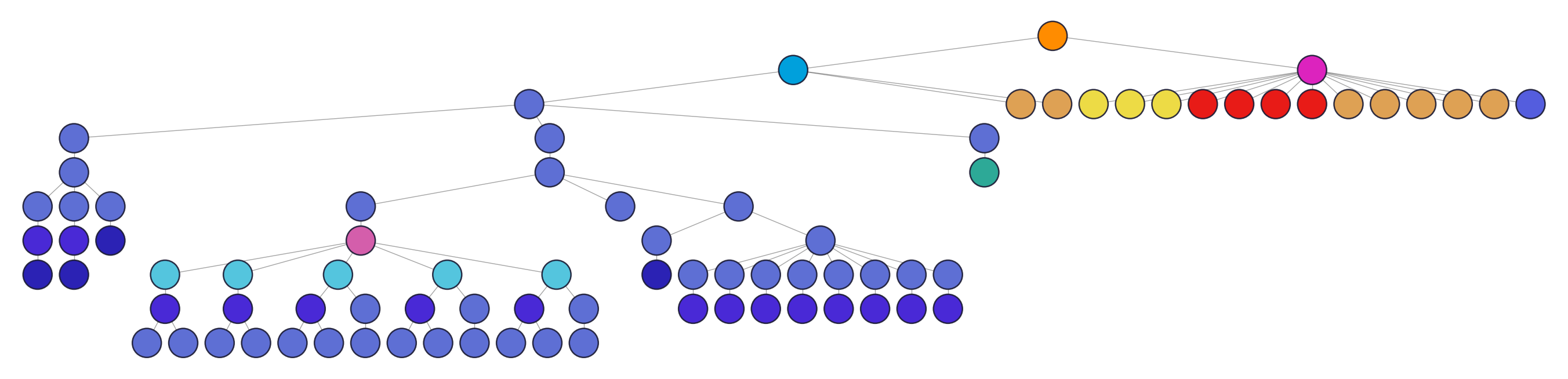} &
        \includegraphics[width=0.25\textwidth]{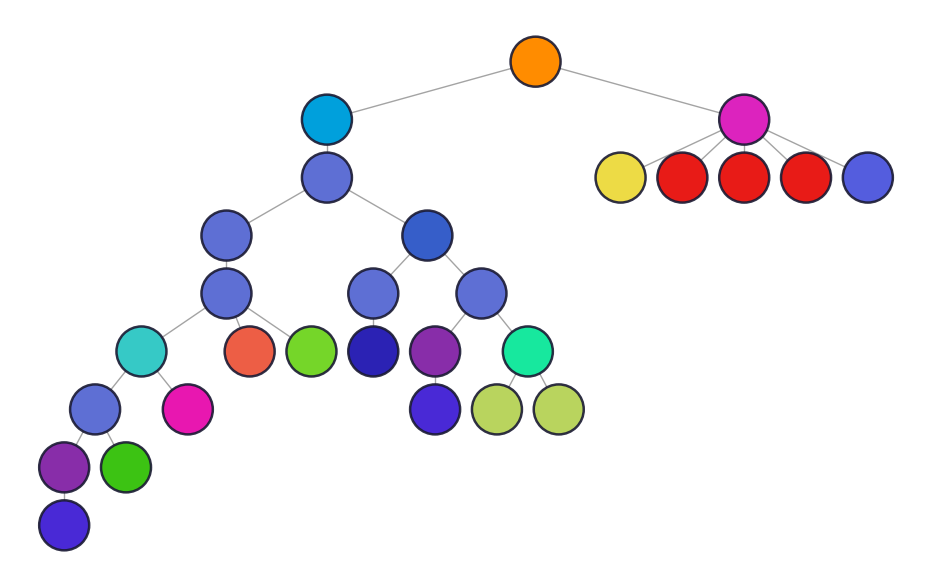} &
        \includegraphics[width=0.15\textwidth]{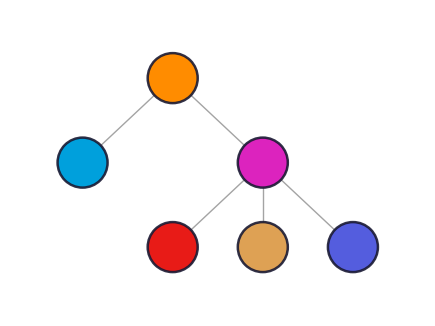} \\
        \textbf{S4} ($\mathcal{LJD}(C) = 0.9916$) &
        \textbf{S5} ($\mathcal{LJD}(C) = 0.9607$) &
        \textbf{S6} ($\mathcal{LJD}(C) = 0.9701$)
    \end{tabular}

    \caption{Structural features}
    \label{fig:qualitative_worst_1}
    \end{subfigure}
    \begin{subfigure}{\textwidth}

    \centering

    \begin{tabular}{ccc}
        \includegraphics[width=0.5\textwidth]{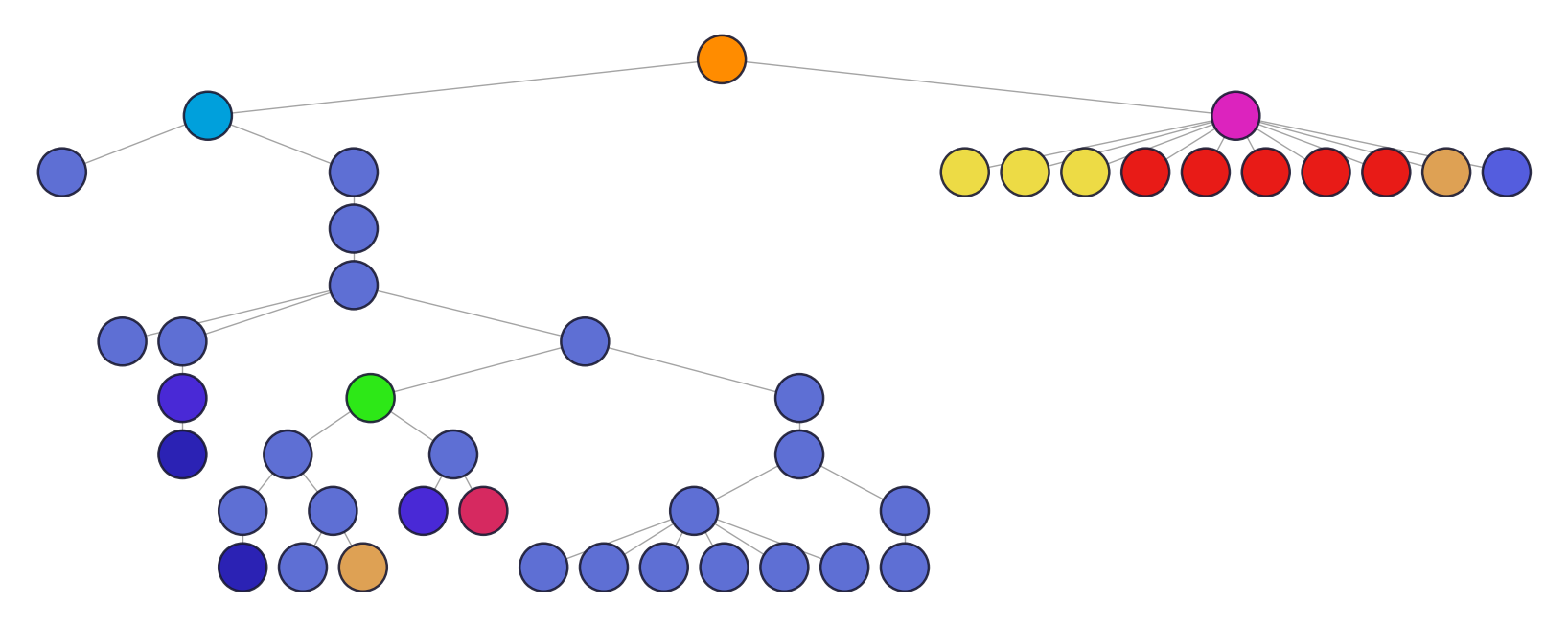} &
        \includegraphics[width=0.15\textwidth]{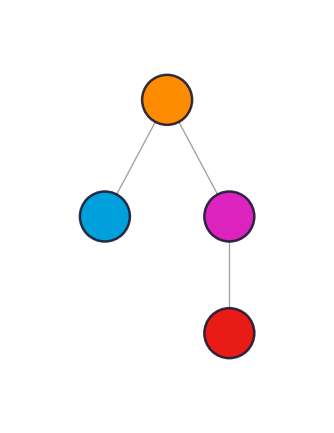} &
        \includegraphics[width=0.2\textwidth]{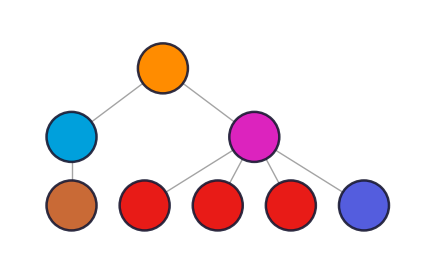} \\
        \textbf{SC4} ($\mathcal{LJD}(C) = 0.9724$) &
        \textbf{SC5} ($\mathcal{LJD}(C) = 0.8626$) &
        \textbf{SC6} ($\mathcal{LJD}(C) = 0.8716$)
    \end{tabular}
    \caption{Structural and content-based features}
    \label{fig:qualitative_worst_2}
    \end{subfigure}

    \caption{Qualitative analysis: - medoids of $N$ dispersed clusters.}
    \label{fig:qualitative_worst}
\end{figure*}

\end{document}